%% file: main.tex
\documentclass[10pt,twocolumn,letterpaper]{article}

 \usepackage{wacv}              
\usepackage{amsmath,amssymb,amsfonts}
\usepackage{algorithmic}
\usepackage{graphicx}
\usepackage{textcomp}
\usepackage{xcolor}
\usepackage{multirow} 
\usepackage{threeparttable}
\usepackage{array}
\usepackage{booktabs}
\usepackage{makecell}

\makeatletter
\@namedef{ver@eso-pic.sty}{2020/01/01}
\makeatother
\usepackage{pdfpages}

\definecolor{darkgreen}{RGB}{0,100,0}  


\definecolor{wacvblue}{rgb}{0.21,0.49,0.74}
\usepackage[pagebackref,breaklinks,colorlinks,allcolors=wacvblue]{hyperref}

\def\wacvPaperID{727} 
\def\confName{WACV}
\def\confYear{2027}

\title{Glass Segmentation with Fusion of Learned and General Visual Features}

\author{Risto Ojala\\
Aalto University\\
Espoo, Finland\\
{\tt\small risto.ojala@aalto.fi}
\and
Tristan Ellison\\
Simon Fraser University\\
Burnaby, Canada\\
{\tt\small tea21@sfu.ca}
\and
Mo Chen\\
Simon Fraser University\\
Burnaby, Canada\\
{\tt\small mochen@sfu.ca}
}

\begin{document}
\maketitle
\begin{abstract}
\input{sections/0_abstract}  
\end{abstract}

\section{Introduction}
\label{sec:intro}
\input{sections/1_intro}

\section{Related Work}
\input{sections/2_relatedwork}

\section{Methods}
\input{sections/3_methods}

\section{Experiments}

\input{sections/4_experiments}

\section{Discussion}
\input{sections/5_discussion}

\section{Conclusion}
\input{sections/6_conclusion}

\section*{Acknowledgements}
This work was partially funded by Nokia Foundation under grant number 20250107, the NSERC Discovery Grants Program, and the Canada CIFAR AI Chairs Program. We gratefully acknowledge the computational resources provided by Aalto Science-IT.

{
    \small
    \bibliographystyle{ieeenat_fullname}
    \bibliography{main}
}

\includepdf[pages=-]{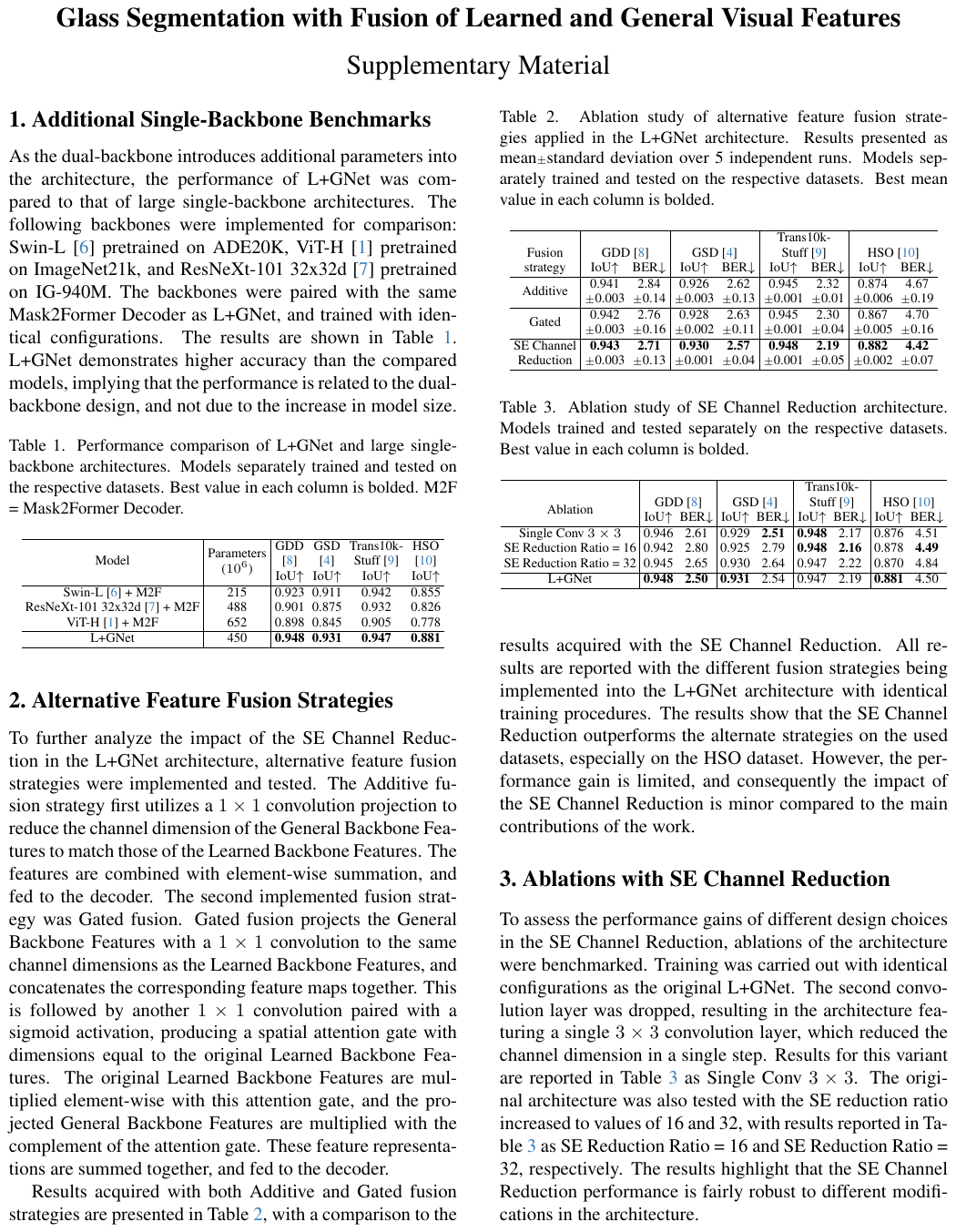}

\end{document}

%% file: sections/0_abstract.tex
Glass surface segmentation from RGB images is a challenging task, with a number of applications in robotics and scene understanding.
As glass lacks coherent visual characteristics, rich context and semantic information is crucial for accurate segmentation.
Consequently, prior works on the task have explored utilization of foundation models and separate semantic backbones.
This paper presents a novel dual-backbone architecture for glass segmentation, applying a frozen foundation model backbone in parallel with a learned backbone trained on task-specific segmentation data.
The learned backbone enables the network to specialize to glass related visual clues, while preserving the general visual feature representations of the foundation model. 
The hierarchical multi-scale features acquired from the dual-backbone are compressed and decoded into segmentation masks.
Benchmarking of the architecture was carried out on four commonly used glass segmentation datasets, achieving state-of-the-art results.
Ablation studies highlight the performance gains of the dual-backbone design and demonstrate the generalizability of the architecture with different backbone choices. 
The model also has a competitive inference speed compared to the previous state-of-the-art method, and surpasses it when using a lighter backbone variant.
The implementation source code and model weights are available at: \url{https://github.com/ojalar/lgnet}.



%% file: sections/1_intro.tex
Glass surface segmentation has been a widely studied problem in the image processing community \cite{mei2020don, lin2021rich}, due to its importance in robotic perception and scene understanding applications.
With glass being transparent and reflective, optical sensors such as cameras and LiDAR have difficulty correctly registering these surfaces as solid material.
Nevertheless, identifying these solid surfaces is critical for indoor scene reconstruction and mapping, as well as obstacle avoidance and navigation of mobile robots. 

Many modalities have been experimented with for glass segmentation, such as LiDAR \cite{zhao20243dref}, RGB-D images \cite{lin2025glass}, and infrared images combined with RGB images \cite{huo2023glass}.
However, glass segmentation from RGB images has remained the most widely studied modality due to its broad applicability, as monocular RGB cameras are some of the cheapest and most commonly used sensors.
Since glass surfaces have a complex and at times non-existent visual appearance, successful detection of a glass surface can often be highly dependent on understanding the visual context.
Therefore, recent foundation models offer a prominent direction for expanding the capabilities of glass segmentation.

\begin{figure}
    \centering
    \includegraphics[width=0.95\columnwidth]{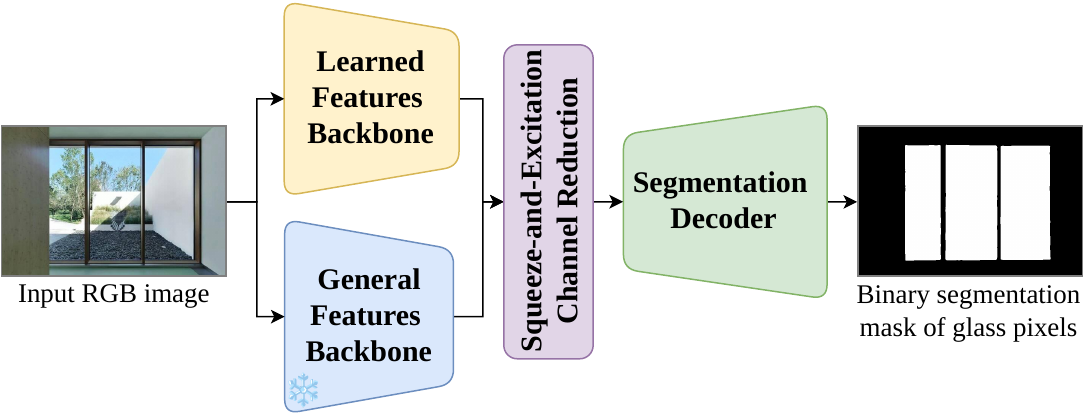}
    \caption{High-level overview of the proposed L+GNet architecture. The dual-backbone utilizes a Learned Features Backbone for generating task-specific features, and a frozen General Features Backbone for context from a vision foundation model.}
    \label{fig:overview}
\end{figure}

This work proposes a novel architecture, L+GNet, for glass segmentation from RGB images, based on a dual-backbone architecture.
An overview of the proposed architecture is presented in Fig. \ref{fig:overview}.
The dual-backbone consists of a Learned Features Backbone and a General Features Backbone.
The Learned Features Backbone produces task-specific representations based on supervised training on glass segmentation data, whereas the General Features Backbone comprises a frozen foundation model that provides general-purpose visual features.
The intuition behind the architecture is to tap into the vast contextual information of a foundation model, while also incorporating directly learned glass related features.

The main contributions of this work can be summarized as follows:
\begin{itemize}
  \item The L+GNet architecture is proposed for glass segmentation, with a novel dual-backbone extracting both general and task-specific learned features.
  \item The fully frozen General Features Backbone produces unmodified foundation model features, enabling efficient integration into perception systems built around a shared foundation model. 
  \item Experimental results show that the L+GNet architecture achieves state-of-the-art results on multiple datasets in terms of accuracy, while maintaining a reasonable computational load.
\end{itemize}


%% file: sections/2_relatedwork.tex
\subsection{Glass Segmentation}

The first deep learning glass segmentation method was proposed in \cite{mei2020don}. Since then, many works have contributed by creating new datasets and techniques, with numerous approaches having used primarily convolutional neural network (CNN) architectures \cite{mei2020don, lin2021rich, xie2020segmenting, xie2021segmenting_tran, he2021enhanced, yu2022progressive, mei2022large, qi2024glass}. 
Many networks have incorporated mechanisms focused on detecting inherent properties of glass. GSDNet \cite{lin2021rich} focuses specifically on the visible reflections present on glass surfaces. During training, a pre-trained reflection detection network was used to guide the predictions of the GSDNet framework. VBNet \cite{qi2024glass} applies a multi-scale segmentation architecture where the low-level features are processed applying visual blurriness blocks in the encoder and the decoder. Visual blurriness was implemented as an average pooling operation, run in parallel with other network layers. As glass itself is difficult to visually detect, the TransLab \cite{xie2020segmenting} architecture focuses on specifically identifying the borders of glass surfaces. The architecture includes a dual-decoder architecture, with a specialized head for boundary detection. As an alternative to the more traditional CNNs, transformer models have also been developed for glass segmentation \cite{xie2021segmenting_tran, yu2022progressive}.

More recent works have explored the use of frozen backbones and diffusion models in glass segmentation. GlassWizard \cite{li2025glasswizard} employs a two-step process using Stable Diffusion 2.0 \cite{rombach2022high} to extract semantic and contextual priors by using a trainable prompt and freezing the diffusion model. In the second phase of training, the diffusion model is trained while the prompt is frozen. Foundation models have also been used for glass segmentation, but with limited task-specific data, it becomes challenging to meaningfully fine-tune these large models. Controllable-LPMoE \cite{sun2025controllable} circumvents this by building trainable features around and into a vision foundation model, while also applying different convolutional kernels to extract unique semantic data. As the most recent works, Controllable-LPMoE and GlassWizard hold the latest state-of-the-art accuracies in the glass segmentation task.

Most similar to the architecture proposed in this paper, GlassSemNet \cite{lin2022exploiting} utilizes a dual-backbone architecture with a DeepLabV3-ResNet50 \cite{chen2017rethinking, he2016deep} model trained as a semantic backbone and a SegFormer MiT backbone \cite{xie2021segformer} applied for spatial feature extraction. 
The semantic backbone was trained on their proposed GSD-S dataset.
The backbone performs semantic segmentation on the input images, providing masks and semantic features for pre-defined classes, such as curtains, blinds, and cabinets. 
Semantic information was proven to be valuable when predicting glass masks. 
However, the used GSD-S dataset is limited to 43 classes, which can limit practical applicability in different environments. 

\subsection{Vision Transformers and Foundation Models} 
Transformers have recently been the most successful models in a number of computer vision tasks, including segmentation.
Vision transformers (ViTs) \cite{dosovitskiy2020image} excel at feature extraction by processing image patches with a global self-attention mechanism. 
The Swin Transformer \cite{liu2021swin} extended on ViTs by applying window-level attention, scaling linearly with respect to input image size. 
The network also shifts window boundaries to prevent feature cut-off in the patching process.
To similarly address scaling, the MixTransformer (MiT) model from the SegFormer architecture \cite{xie2021segformer} introduced efficient self-attention.
Prior to computing global self-attention, the model compresses the length of the key-embedding with a linear layer based on a chosen reduction ratio.
These models perform well as backbones in segmentation architectures, feeding features to a decoder module.
Mask2Former \cite{cheng2022masked} has recently been widely used as segmentation decoder, showing top performance in a large number of applications.

Vision foundation models have emerged based on the vast learning capabilities of the generalizable ViT architecture.
These models utilize extensive self-supervised training processes to learn meaningful, generalizable feature representations. 
DINOv3 \cite{simeoni2025dinov3} is a recent vision foundation model, trained via self-distillation on large-scale unlabelled image data.
Natural language text is also commonly used as an additional modality in the training recipe of foundation models, with SigLIP 2 \cite{tschannen2025siglip} being a high-performance example.
In addition to the image-text modalities, Perception Encoder (PE-Core) \cite{bolya2026perception} incorporated video data in its training recipe.
For input images, these models provide detailed and general-purpose feature maps that can be used for different tasks, including segmentation. 

\subsection{Research Gap}
We study the research gap of advancing glass segmentation performance by more effectively leveraging foundation model features.
Simultaneously, we leave the original feature representations intact, maintaining usability for other perception applications.
We build on previous works \cite{lin2022exploiting, sun2025controllable} by proposing a novel foundation model-based dual-backbone architecture for glass segmentation.
Dual-backbones have been previously explored for glass segmentation \cite{lin2022exploiting}, yet this previous approach used a frozen semantic backbone trained on a fixed number of classes. 
We apply a foundation model in the dual-backbone, producing context via general visual features instead of semantic clues tied to a specific list of trained classes.
Additionally, we incorporate a more efficient fusion strategy for combining the backbone features for the decoder.
Prior work \cite{sun2025controllable} has applied foundation models to glass segmentation by inserting trainable adapters alongside a frozen backbone, injecting task-specific features into the intermediate layers of the backbone.
We demonstrate using a foundation model in parallel with a fully separate task-specific learned backbone.
This contributes to system-level efficiency in online applications, as the foundation model features can also be used for other perception tasks.
Our novel architecture is shown to produce state-of-the-art results on the commonly used datasets, also outperforming the recent diffusion-based glass segmentation approach \cite{li2025glasswizard}.

%% file: sections/3_methods.tex
\begin{figure*}
    \centering
    \includegraphics[width=\textwidth]{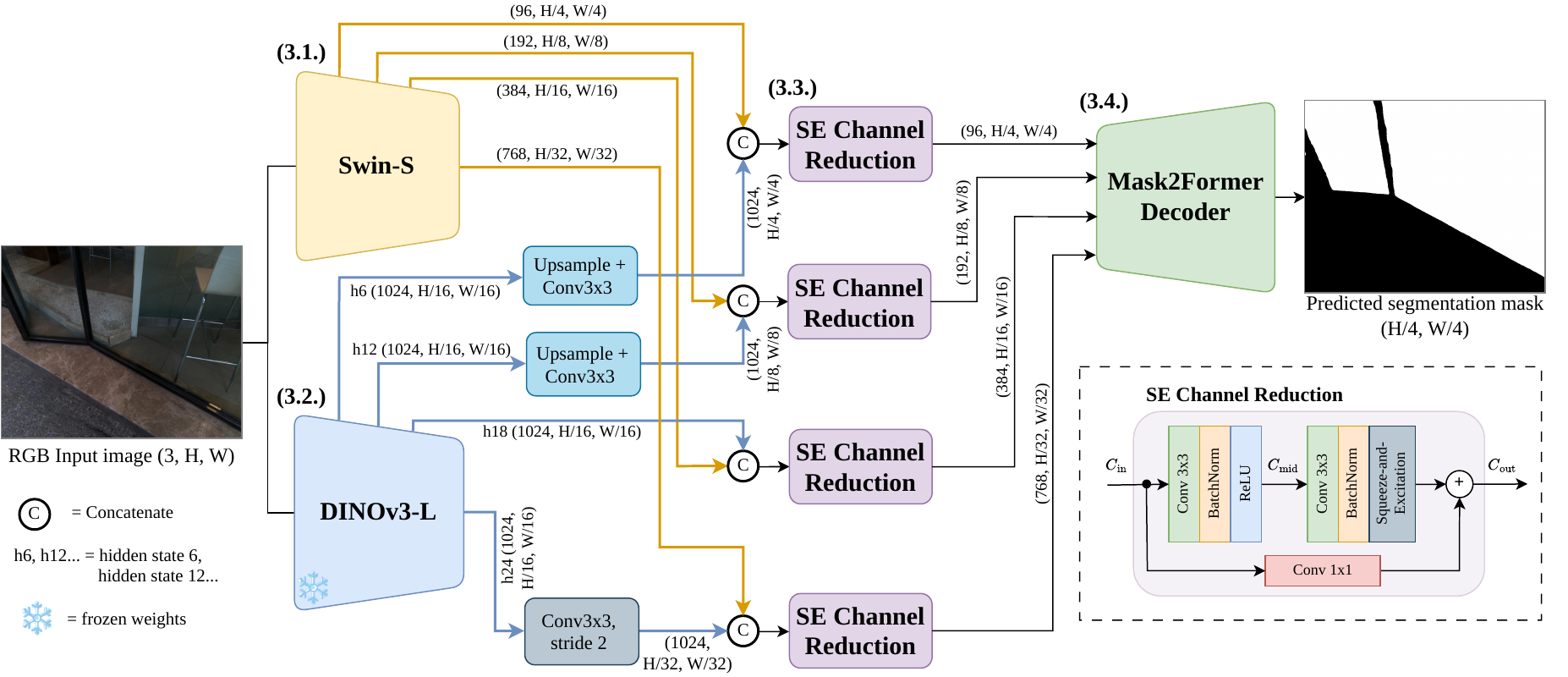}
    \caption{The proposed L+GNet architecture. Bottom right corner shows the detailed contents of a SE Channel Reduction block. Labels (3.1.)-(3.4.) refer to corresponding subsections, which contain detailed descriptions of the specific blocks. }
    \label{fig:architecture}
\end{figure*}

Given an input RGB image, the analyzed task is to generate a binary segmentation mask correctly classifying each pixel in the image.
Each pixel in an image is classified to contain either ``glass'' or ``background''.
We propose a novel deep learning architecture, coined L+GNet, for the glass segmentation task.
The detailed architecture of L+GNet is visualized in Fig. \ref{fig:architecture}, with the building blocks described below.

\subsection{Learned Features Backbone}
The Learned Features Backbone is incorporated into the architecture to learn relevant, task-specific feature representations from available training data for glass segmentation.
A Swin-S transformer model was chosen as the Learned Features Backbone of the L+GNet architecture.
As a transformer-based model, Swin-S not only captures local features, but also long-range spatial dependencies through shifted window-based self-attention.

For an input image, Swin-S produces hierarchical multi-scale feature representations at sizes of 1/4, 1/8, 1/16 and 1/32 of the input image dimensions.
Higher resolution features are useful for producing high resolution segmentation masks.
The higher resolution features are a valuable contribution of the Learned Features Backbone, as the General Features Backbone applied here produces feature maps with a patch size of 16x16 pixels.
Large patch sizes are common for foundation models, as the models are typically based on the ViT-architecture and apply global self-attention mechanisms.

\subsection{General Features Backbone}
\begin{figure}
\centering
\renewcommand{\arraystretch}{0.3}
\setlength{\tabcolsep}{1pt}
\begin{tabular}{*{4}{>{\centering\arraybackslash}m{0.24\columnwidth}}}
\includegraphics[width=\linewidth, height=\linewidth]{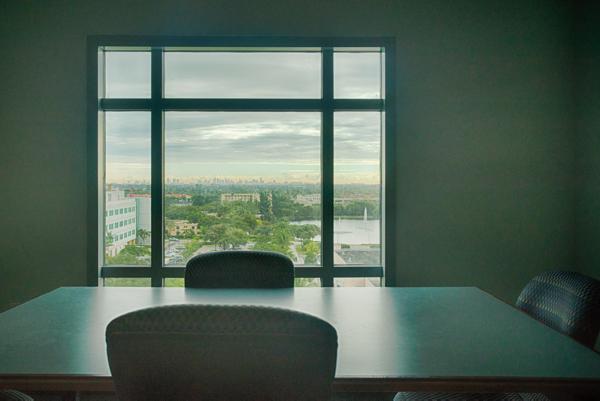} &
\includegraphics[width=\linewidth, height=\linewidth]{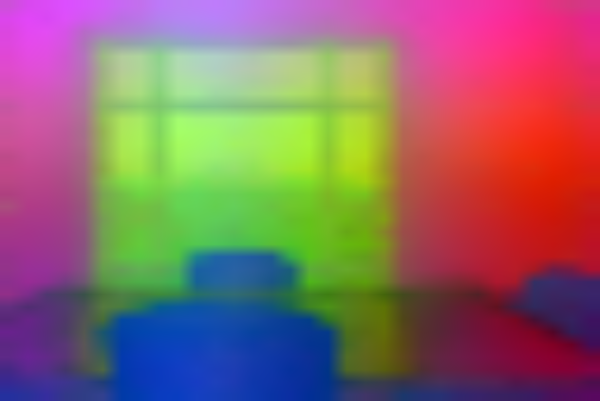}  &
\includegraphics[width=\linewidth, height=\linewidth]{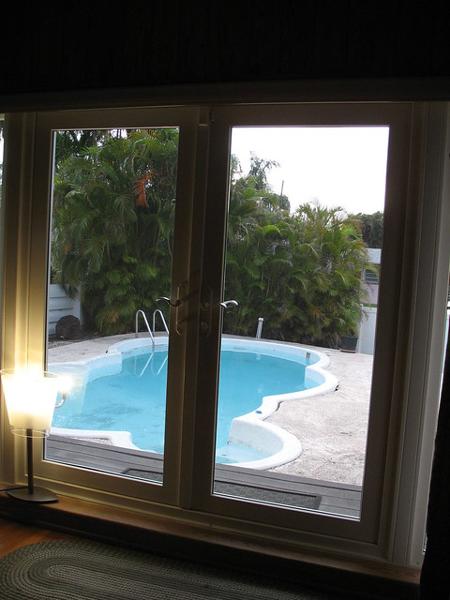} &
\includegraphics[width=\linewidth, height=\linewidth]{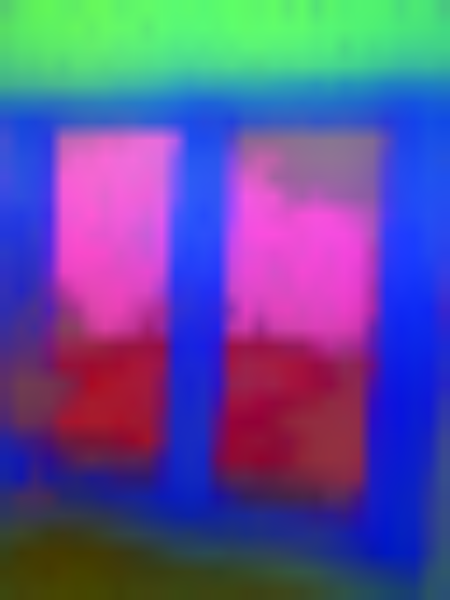} \\
\includegraphics[width=\linewidth, height=\linewidth]{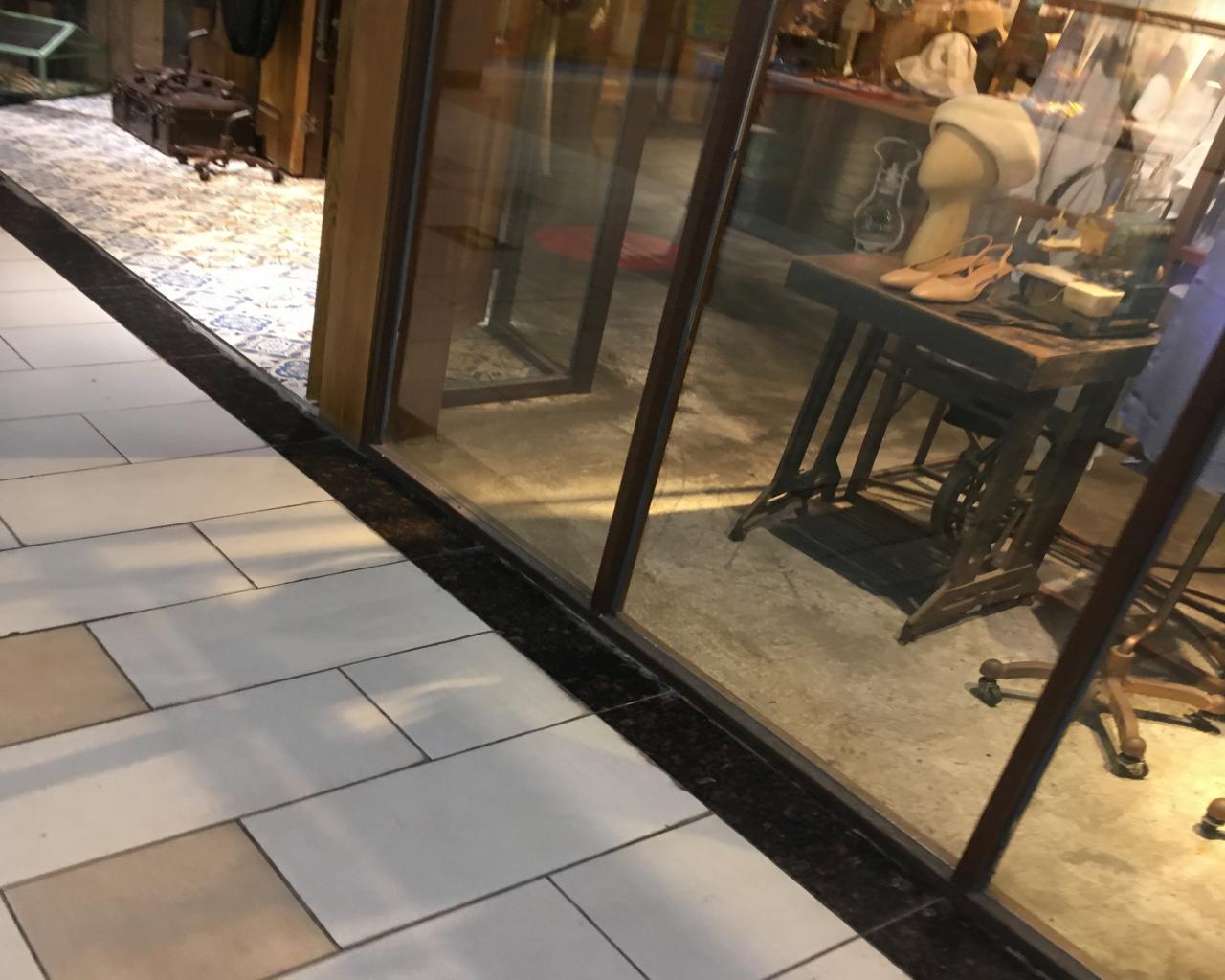} &
\includegraphics[width=\linewidth, height=\linewidth]{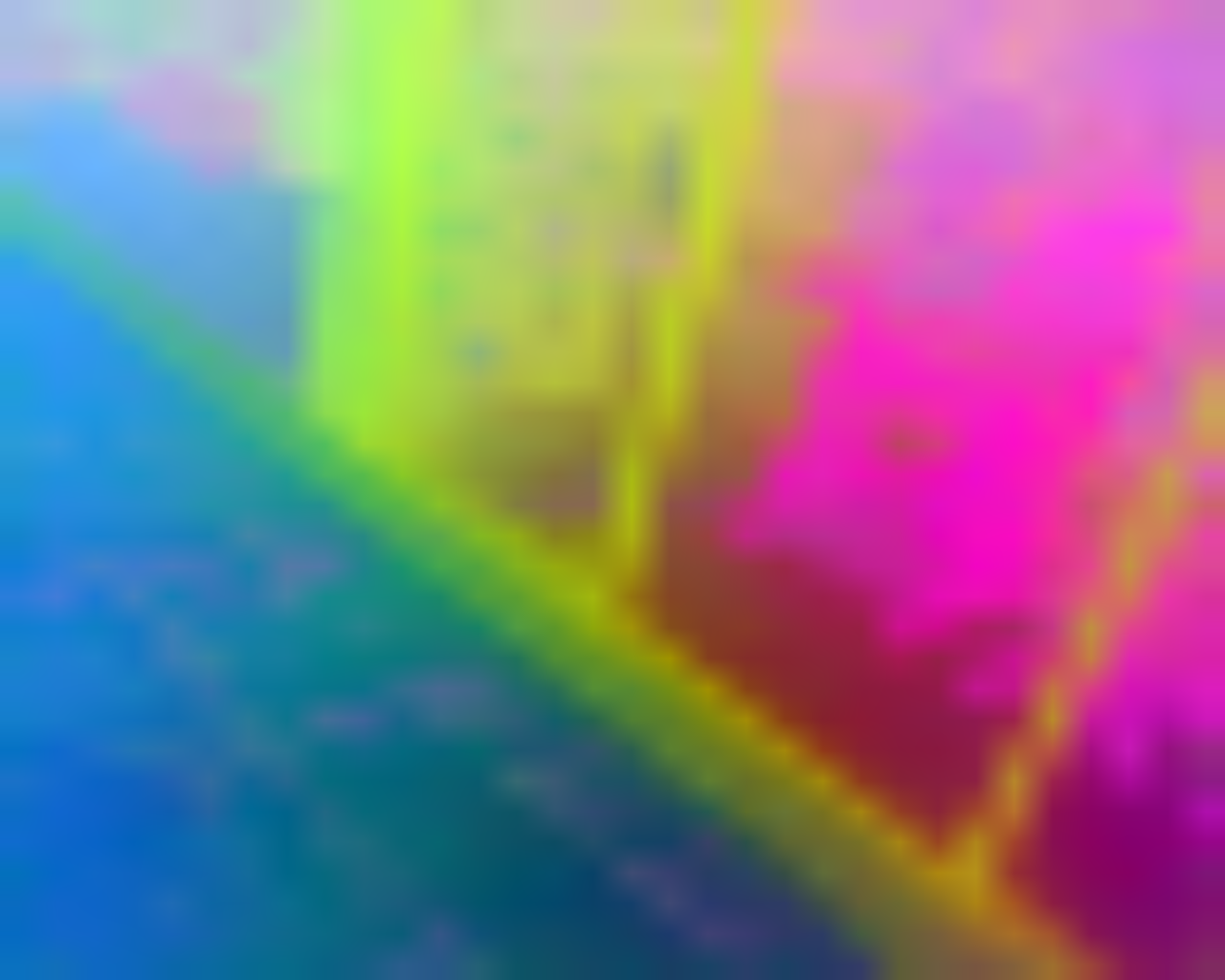} &
\includegraphics[width=\linewidth, height=\linewidth]{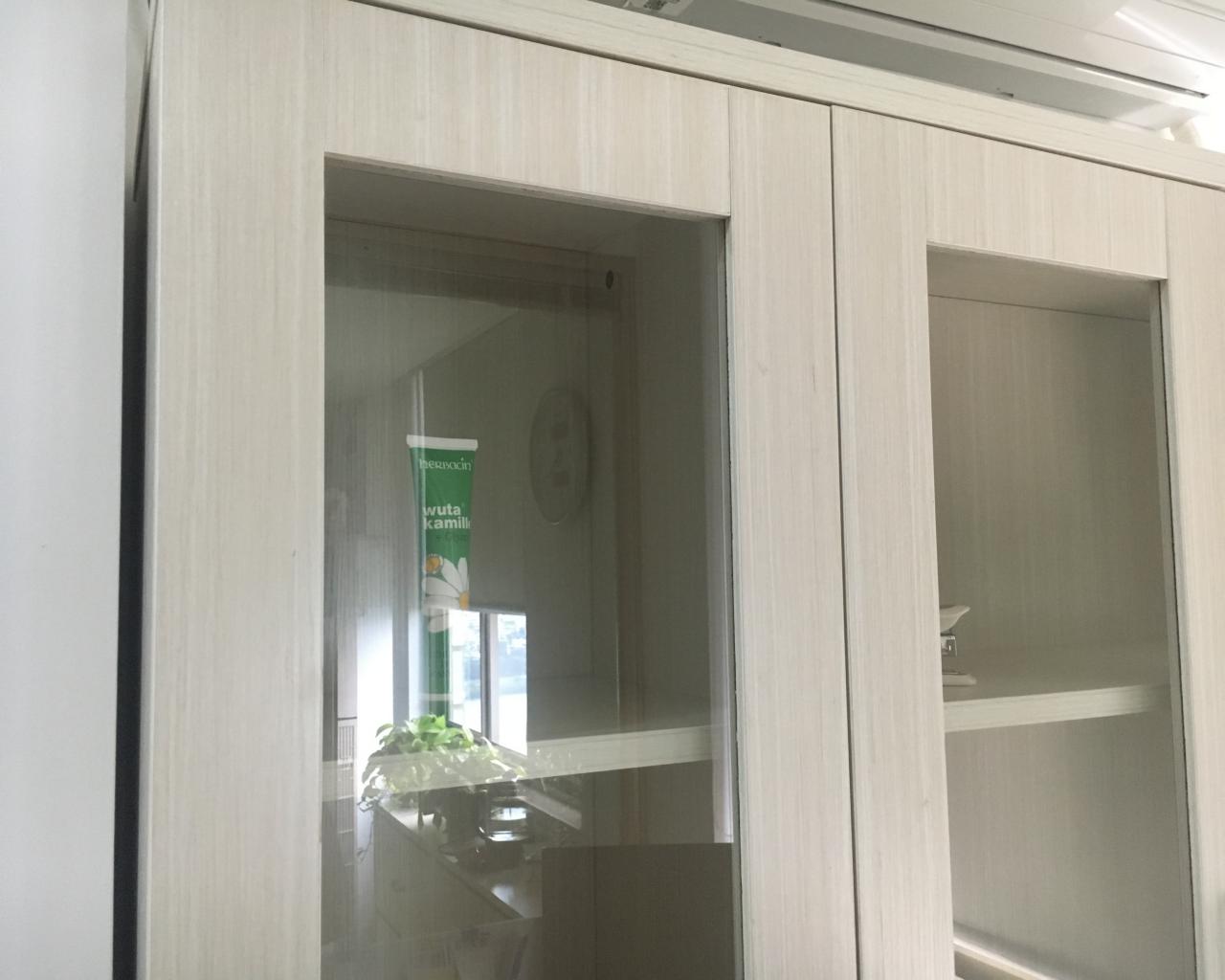} &
\includegraphics[width=\linewidth, height=\linewidth]{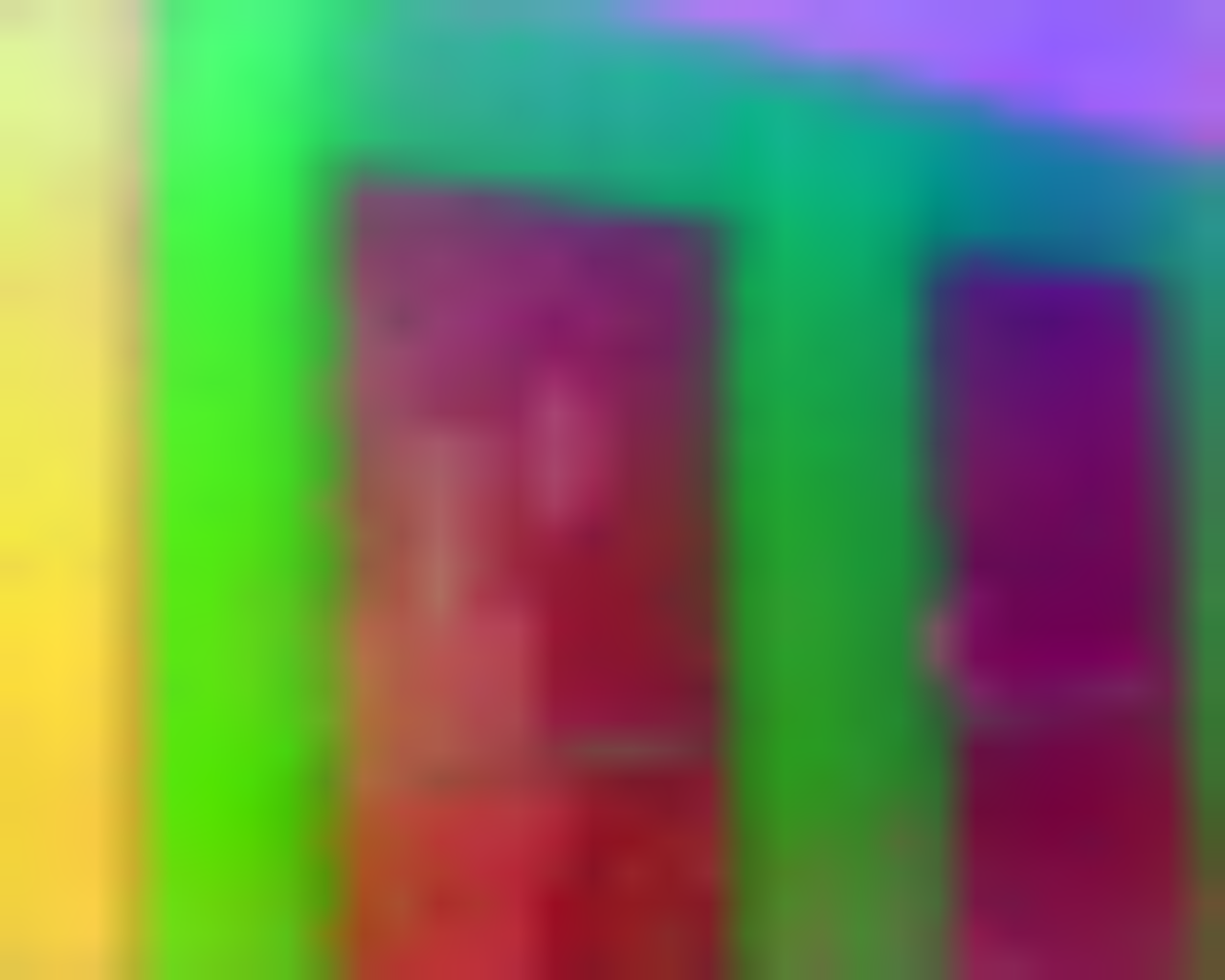} 
\end{tabular}
\caption{The rich semantic information of DINOv3-L final layer patch embeddings visualized with per image principal component analysis into RGB. Images from the GDD \cite{mei2020don} and GSD \cite{lin2021rich} datasets.} 
\label{fig:dino_features}
\end{figure}

The General Features Backbone of the L+GNet architecture applies a foundation model to extract general visual features from the input image.
Due to the limited visual information found directly in transparent glass surfaces, the general features offer important global context for the segmentation task.
The weights of the General Features Backbone are frozen in the L+GNet architecture, not being optimized during training.
DINOv3-L was chosen as the foundation model for the General Features Backbone.
The strong semantic nature of DINOv3-L features is visualized in Figure \ref{fig:dino_features}, highlighting the context provided by the backbone.

DINOv3-L is used in the General Features Backbone by extracting different hidden states from the model, acquiring hierarchical feature representations from different stages of processing.
The model outputs are the hidden states of the patch embeddings after transformer blocks (6, 12, 18, 24).
When experimenting with smaller variants of DINOv3, hidden states from layers (3, 6, 9, 12) were used as backbone outputs.
The layer outputs are upsampled or downsampled with convolutional resizing adapters to match the dimensions of the corresponding multi-scale Learned Features Backbone outputs.
Corresponding outputs from the two backbones are concatenated along the feature channel dimension.

\subsection{Squeeze-and-Excitation Channel Reduction}
Concatenation of the outputs from the two backbones results in a high number of extracted features.
To reduce the number of feature channels for the Segmentation Decoder, the SE Channel Reduction block was implemented.
The detailed contents of the block are visualized in Fig. \ref{fig:architecture}.
The block follows a standard residual SE block architecture \cite{hu2018squeeze}, with the exception that the feature channel is reduced in a stepwise fashion to avoid drastic reduction of the channel dimension in a single layer.
The first convolutional layer in the block outputs a channel dimension of $C_\text{mid}$, which is defined as
\begin{equation}
C_\text{mid} = \max\left(\left\lfloor \frac{C_\text{in}}{2} \right\rfloor,\ C_\text{out}\right),
\end{equation}
where $C_\text{in}$ denotes the number of channels in the input, and $C_\text{out}$ denotes the number of channels in the output.
The SE operation provides the channel reduction with an attention-mechanism which amplifies or dampens specific feature channels.
This is relevant for the architecture, as the general-purpose feature representations from the General Features Backbone can have varying levels of importance for the segmentation task.


\subsection{Segmentation Decoder}
A standard Mask2Former Decoder \cite{cheng2022masked} was adopted to generate segmentation masks from the feature representations output by the SE Channel Reduction.
The standard decoder design choice results in most of the representational capacity of the architecture being in the dual-backbone, keeping the decoder lightweight.
Other similar decoders, such as DPT \cite{ranftl2021vision} or SegFormer decoder \cite{xie2021segformer}, can be conveniently swapped into the architecture.
The Mask2Former Decoder consists of a pixel decoder and a transformer decoder. 
The pixel decoder employs a deformable attention transformer \cite{zhu2020deformable} to fuse multi-scale features and produce dense, high-resolution pixel embeddings. 
The transformer decoder leverages global context through masked attention mechanisms to generate mask predictions based on the pixel decoder outputs, in a DETR-style query formulation \cite{carion2020end}. 
Given the input feature maps, the decoder produces class queries $\mathbf{Z} \in \mathbb{R}^{Q \times (K+1)}$ and mask queries $\mathbf{M} \in \mathbb{R}^{Q \times H_l \times W_l}$, where $Q$ refers to the number of queries, and $K$ refers to the number of semantic classes (here ``glass'' and ``background'', $K=2$).
The implementation incorporates a ``no object'' class, corresponding to the added extra column in $\mathbf{Z}$.
$H_l$ and $W_l$ refer to the height and width of mask logit maps, respectively, corresponding to downsampled dimensions of the original image.
Class probabilities $\textbf{C} \in [0,1]^{Q\times K}$ are acquired applying softmax over all $K+1$ classes, retaining only the actual $K$ semantic classes
\begin{equation}
C_{q,k} = \frac{\exp(Z_{q,k})}{\displaystyle\sum_{k'=1}^{K+1}\exp(Z_{q,k'})}, \qquad k = 1,\dots,K.
\end{equation}
Mask logits $\mathbf{M}$ are converted into mask probabilities $\mathbf{P} \in [0,1]^{Q \times H_l \times W_l}$ with an element-wise sigmoid operation.
Pixel-wise semantic segmentation logits $\mathbf{S} \in \mathbb{R}^{K \times H_l \times W_l}$ are obtained by
\begin{equation}
S_k(h_l,w_l) = \sum_{q=1}^{Q} C_{q,k}\,P_q(h_l,w_l).
\end{equation}
These logits are bilinearly interpolated to original image dimensions, and binary pixel-wise predictions are acquired by selecting the class with the maximal logit value.

%% file: sections/4_experiments.tex
\subsection{Datasets}
Latest versions of four datasets were used to evaluate the proposed L+GNet: GDD \cite{mei2020don}, GSD \cite{lin2021rich}, Trans10k-Stuff \cite{xie2020segmenting}, and HSO \cite{yu2022progressive}.
The datasets contain 2980/936, 3285/813, 2455/1771, and 3070/1782 train/test images, respectively.
Validation splits are not present in the datasets.
The official training and testing splits were used for training and testing, respectively.
These datasets were selected based on wide representation in previous literature \cite{yu2022progressive, qi2024glass, li2025glasswizard}.
L+GNet and its variants were trained on the respective training splits and evaluated on the corresponding testing splits.
Additionally, an experiment was conducted where the model was trained on all of the available training splits combined, and evaluated on the testing splits.

\subsection{Implementation details}
L+GNet was implemented with PyTorch and Hugging Face, with pretrained models acquired from Hugging Face repositories.
For Mask2Former Decoder and Swin-S, ADE20k pretrained weights were used for initialization.
Hyperparameters for the Mask2Former Decoder and Swin-S were taken directly from the Hugging Face implementations.
Both the feature size and hidden dimension of the Mask2Former model were set to the default value of 256.
A reduction ratio of 8 was used in the SE blocks of the L+GNet architecture.
The loss function from the original Mask2Former paper \cite{cheng2022masked} was used to train all models.
For ablating the backbone choices, ADE20k pretrained MiT-B3 and Swin-B were implemented from Hugging Face, as well as SigLIP 2-L and PE-Core-L were sourced from Hugging Face and timm, respectively.

Following recent literature \cite{li2025glasswizard,qi2024glass}, intersection-over-union (IoU) and balanced error rate (BER) were used for evaluating model performance.
In all experiments, L+GNet and its variants were trained for 30 epochs with a batch size of 16, using the AdamW optimizer, with a weight decay of $10^{-4}$.
The final checkpoint after the last epoch was always saved for testing.
A linearly decreasing learning rate scheduler was used, with an initial learning rate of $10^{-4}$, and a warm-up of 500 steps.
PyTorch mixed precision was used during training to reduce the memory footprint.
During training, random horizontal and vertical flips were used for data augmentation.
During training and testing, images were resized to $512\times 512$ pixels prior to being fed to the models.
All accuracy testing was carried out with FP32 precision.

\subsection{Accuracy Benchmark}
Results for the accuracy benchmark on the four datasets are shown in Table \ref{tab:results}, with comparison to scores achieved by other methods in previous literature.
Table \ref{tab:results_all} compares the performance of L+GNet with the previous state of the art when trained on the combined training data of all four datasets.
Additional accuracy benchmarks are presented in the Supplementary Material of the paper.
Overall, the results show that L+GNet performs exceptionally well on a number of datasets, improving state-of-the-art results by a notable margin.
Samples of segmentation masks produced by L+GNet on the testing sets are visualized in Fig. \ref{fig:samples_segmentation}.
Failure cases of L+GNet segmentation results are shown in Fig. \ref{fig:failures}.
Also, a visual comparison to previous state of the art is shown in Fig. \ref{fig:comparison}, highlighting challenging scenarios where previous state of the art fails, yet L+GNet manages to perform segmentation well.

\begin{figure*}[tb]
  \centering
  \includegraphics[trim=0.3cm 0.3cm 0.3cm 0.3cm, clip, width=0.24\linewidth]{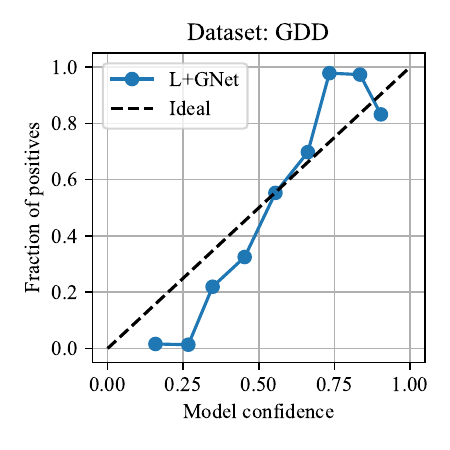}
  \includegraphics[trim=0.3cm 0.3cm 0.3cm 0.3cm, clip, width=0.24\linewidth]{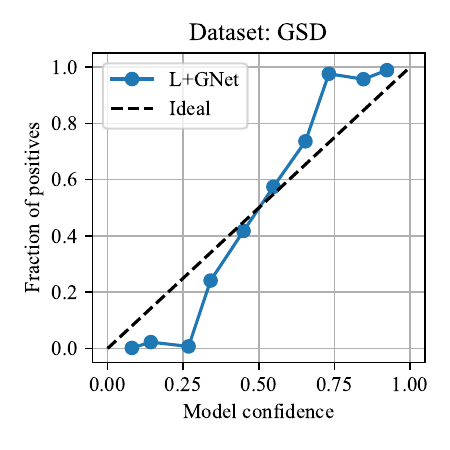}
  \includegraphics[trim=0.3cm 0.3cm 0.3cm 0.3cm, clip, width=0.24\linewidth]{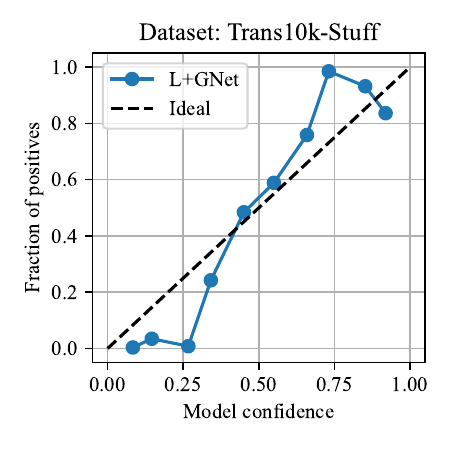}
  \includegraphics[trim=0.3cm 0.3cm 0.3cm 0.3cm, clip, width=0.24\linewidth]{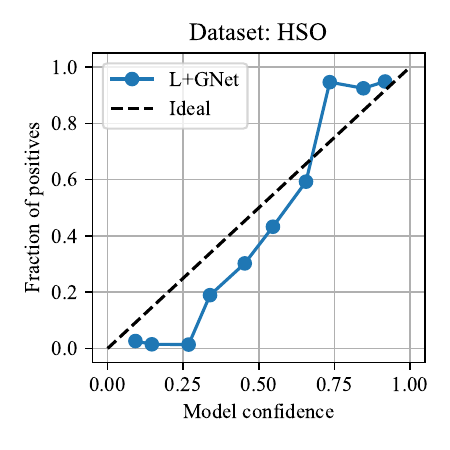}
  \caption{Calibration curves for the L+GNet model on the datasets used for testing. Model trained on the respective training splits.}
  \label{fig:calibration}
\end{figure*}

Although L+GNet excels in binary segmentation, the model had difficulties providing calibrated prediction confidences.
Calibration curves (i.e. reliability diagrams) for each dataset are presented in Fig. \ref{fig:calibration}, with confidence values acquired applying softmax on the pixel-wise segmentation logits.
The model was rarely reporting confidence values at either extreme of the scale, with few predictions assigned confidence values outside of an interval of approximately 0.3 to 0.7.
This is likely due to the DETR-style query-based segmentation strategy of the Mask2Former Decoder, fusing several mask predictions into the final prediction.

\begin{table}
\centering
\begin{threeparttable}
\caption{Comparison of different models on multiple datasets. Results for SAM \cite{han2023segment}, GlassSemNet \cite{lin2022exploiting} and C-LPMoE \cite{sun2025controllable} taken from the respective papers. Results for VBNet, EBLNet and GDNet-B taken from \cite{qi2024glass}. Results for GlassWizard taken from the supplementary material of the paper \cite{li2025glasswizard}. Results for GDNet \cite{mei2020don} and GSDNet \cite{lin2021rich} taken from respective papers as well as from \cite{yu2022progressive}. Other benchmark results taken from \cite{yu2022progressive}. Bolded values indicate the best single-shot result in each column. Colored percentage values indicate the relative difference of L+GNet results compared to best results in the previous literature.}
\label{tab:results}
\scriptsize
\setlength{\tabcolsep}{0.8pt}
\begin{tabular}{l|cc|cc|cc|cc}
\hline
\multirow{2}{*}{Model} & \multicolumn{2}{c|}{GDD \cite{mei2020don}} & \multicolumn{2}{c|}{GSD \cite{lin2021rich}} & \multicolumn{2}{c|}{\begin{tabular}[b]{@{}c@{}}Trans10k-\\Stuff \cite{xie2020segmenting}\end{tabular}} & \multicolumn{2}{c}{HSO \cite{yu2022progressive}} \\
& IoU$\uparrow$ & BER$\downarrow$
& IoU$\uparrow$ & BER$\downarrow$
& IoU$\uparrow$ & BER$\downarrow$
& IoU$\uparrow$ & BER$\downarrow$ \\
\hline
SAM \cite{han2023segment} & 0.485 & 26.08 & 0.506 & 23.91 & - & - & - & - \\
TransLab \cite{xie2020segmenting} & 0.816 & 9.70 & 0.781 & 9.19 & 0.871 & 5.44 & 0.743 & 12.00 \\
Trans2Seg \cite{xie2021segmenting_tran} & 0.844 & 7.36 & 0.797 & 8.21 & 0.750 & 10.73 & 0.780 & 9.65 \\
GDNet \cite{mei2020don} & 0.876 & 5.62 & 0.825 & 6.41 & 0.887 & 4.72 & 0.787 & 9.32 \\
GSDNet \cite{lin2021rich} & 0.875 & 5.71 & 0.836 & 6.12 & 0.897 & 4.52 & 0.789 & 9.79 \\
EBLNet \cite{he2021enhanced} & 0.882 & 5.38 & 0.817 & 6.75 & - & - & - & - \\
PGSNet \cite{yu2022progressive} & 0.878 & 5.56 & 0.836 & 6.25 & 0.898 & 4.39 & 0.801 & 9.08 \\
GDNet-B \cite{mei2022large} & 0.878 & 5.52 & - & - & - & - & - & - \\
GlassSemNet \cite{lin2022exploiting} & 0.908 & 4.34 & 0.856 & 5.60 & - & - & - & - \\
VBNet \cite{qi2024glass} & 0.907 & 4.70 & 0.861 & 5.51 & 0.916 & 3.41 & 0.831 & 7.65 \\
C-LPMoE-u \cite{sun2025controllable} & 0.923 & - & - & - & - & - & - & - \\
GlassWizard \cite{li2025glasswizard} & 0.921 & 3.86 & 0.891 & 4.14 & 0.930 & 2.91 & 0.867 & 6.06 \\
\hline
L+GNet\tnote{a} & \textbf{0.948} & \textbf{2.50} & \textbf{0.931} & \textbf{2.54} & \textbf{0.947} & \textbf{2.24} & \textbf{0.881} & \textbf{4.50} \\
& \textcolor{darkgreen}{+2.7\%} & \textcolor{darkgreen}{-35\%} & \textcolor{darkgreen}{+4.5\%} & \textcolor{darkgreen}{-39\%} & \textcolor{darkgreen}{+1.8\%} & \textcolor{darkgreen}{-23\%} & \textcolor{darkgreen}{+1.6\%} & \textcolor{darkgreen}{-26\%} \\
\hline
L+GNet\tnote{b} &
\begin{tabular}[t]{@{}c@{}}0.943\\{\tiny$\pm$}0.003\end{tabular} &
\begin{tabular}[t]{@{}c@{}}2.71\\{\tiny$\pm$}0.13\end{tabular} &
\begin{tabular}[t]{@{}c@{}}0.930\\{\tiny$\pm$}0.001\end{tabular} &
\begin{tabular}[t]{@{}c@{}}2.57\\{\tiny$\pm$}0.04\end{tabular} &
\begin{tabular}[t]{@{}c@{}}0.948\\{\tiny$\pm$}0.001\end{tabular} &
\begin{tabular}[t]{@{}c@{}}2.19\\{\tiny$\pm$}0.05\end{tabular} &
\begin{tabular}[t]{@{}c@{}}0.882\\{\tiny$\pm$}0.002\end{tabular} &
\begin{tabular}[t]{@{}c@{}}4.42\\{\tiny$\pm$}0.07\end{tabular} \\
\hline
\end{tabular}
\begin{tablenotes}
\scriptsize
\item[a] Result corresponds to the publicly shared weights.
\item[b] Mean {\tiny$\pm$} standard deviation over 5 independent training runs.
\end{tablenotes}
\end{threeparttable}
\end{table}

\begin{table}
\centering
\begin{threeparttable}
\caption{Segmentation accuracy comparison to previous state of the art, with models trained on all of the datasets. Results for GlassWizard taken from the respective paper \cite{li2025glasswizard}. Bolded values indicate the best single-shot result in each column. Colored percentage values indicate the relative difference of L+GNet results compared to previous state of the art.}
\label{tab:results_all}
\setlength{\tabcolsep}{0.8pt}
\scriptsize
\begin{tabular}{p{1.8cm}|cc|cc|cc|cc}
\hline
\multirow{2}{*}{Model} & \multicolumn{2}{c|}{GDD \cite{mei2020don}} & \multicolumn{2}{c|}{GSD \cite{lin2021rich}} & \multicolumn{2}{c|}{\begin{tabular}[b]{@{}c@{}}Trans10k-\\Stuff \cite{xie2020segmenting}\end{tabular}} & \multicolumn{2}{c}{HSO \cite{yu2022progressive}} \\
& IoU$\uparrow$ & BER$\downarrow$
& IoU$\uparrow$ & BER$\downarrow$
& IoU$\uparrow$ & BER$\downarrow$
& IoU$\uparrow$ & BER$\downarrow$ \\
\hline
GlassWizard \cite{li2025glasswizard} & 0.933 & 3.62 & 0.904 & 4.03 & 0.928 & 3.04 & 0.879 & 5.44 \\
\hline
L+GNet\tnote{a} & \textbf{0.951} & \textbf{2.33} & \textbf{0.938} & \textbf{2.28} & \textbf{0.951} & \textbf{2.05} & \textbf{0.893} & \textbf{3.98} \\
& \textcolor{darkgreen}{+1.9\%} & \textcolor{darkgreen}{-36\%} & \textcolor{darkgreen}{+3.8\%} & \textcolor{darkgreen}{-43\%} & \textcolor{darkgreen}{+2.5\%} & \textcolor{darkgreen}{-33\%} & \textcolor{darkgreen}{+1.6\%} & \textcolor{darkgreen}{-27\%} \\
\hline
L+GNet\tnote{b} &
\begin{tabular}[t]{@{}c@{}}0.953\\{\tiny$\pm$}0.002\end{tabular} &
\begin{tabular}[t]{@{}c@{}}2.23\\{\tiny$\pm$}0.08\end{tabular} &
\begin{tabular}[t]{@{}c@{}}0.939\\{\tiny$\pm$}0.001\end{tabular} &
\begin{tabular}[t]{@{}c@{}}2.22\\{\tiny$\pm$}0.04\end{tabular} &
\begin{tabular}[t]{@{}c@{}}0.952\\{\tiny$\pm$}0.001\end{tabular} &
\begin{tabular}[t]{@{}c@{}}2.00\\{\tiny$\pm$}0.03\end{tabular} &
\begin{tabular}[t]{@{}c@{}}0.900\\{\tiny$\pm$}0.004\end{tabular} &
\begin{tabular}[t]{@{}c@{}}3.77\\{\tiny$\pm$}0.13\end{tabular} \\
\hline
\end{tabular}
\begin{tablenotes}
\scriptsize
\item[a] Result corresponds to the publicly shared weights.
\item[b] Mean {\tiny$\pm$} standard deviation over 5 independent runs.
\end{tablenotes}
\end{threeparttable}
\end{table}

\begin{figure*}
\centering
\setlength{\tabcolsep}{3pt}
\scriptsize
\begin{tabular}{>{\centering\arraybackslash}m{0.7cm}|*{7}{>{\centering\arraybackslash}m{0.25\columnwidth}}}

\rotatebox{90}{GDD \cite{mei2020don}} &
\includegraphics[width=\linewidth,height=\linewidth]{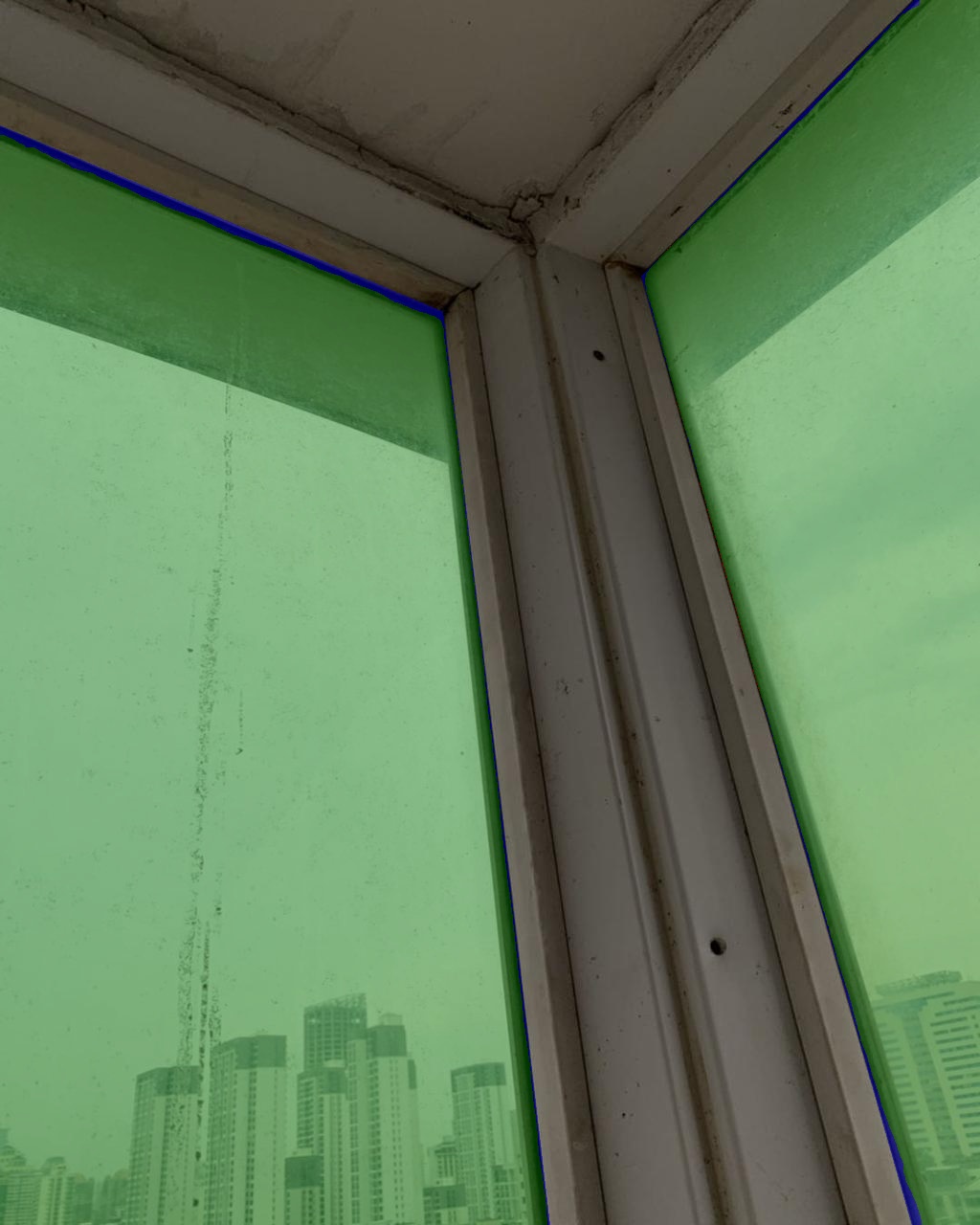} &
\includegraphics[width=\linewidth,height=\linewidth]{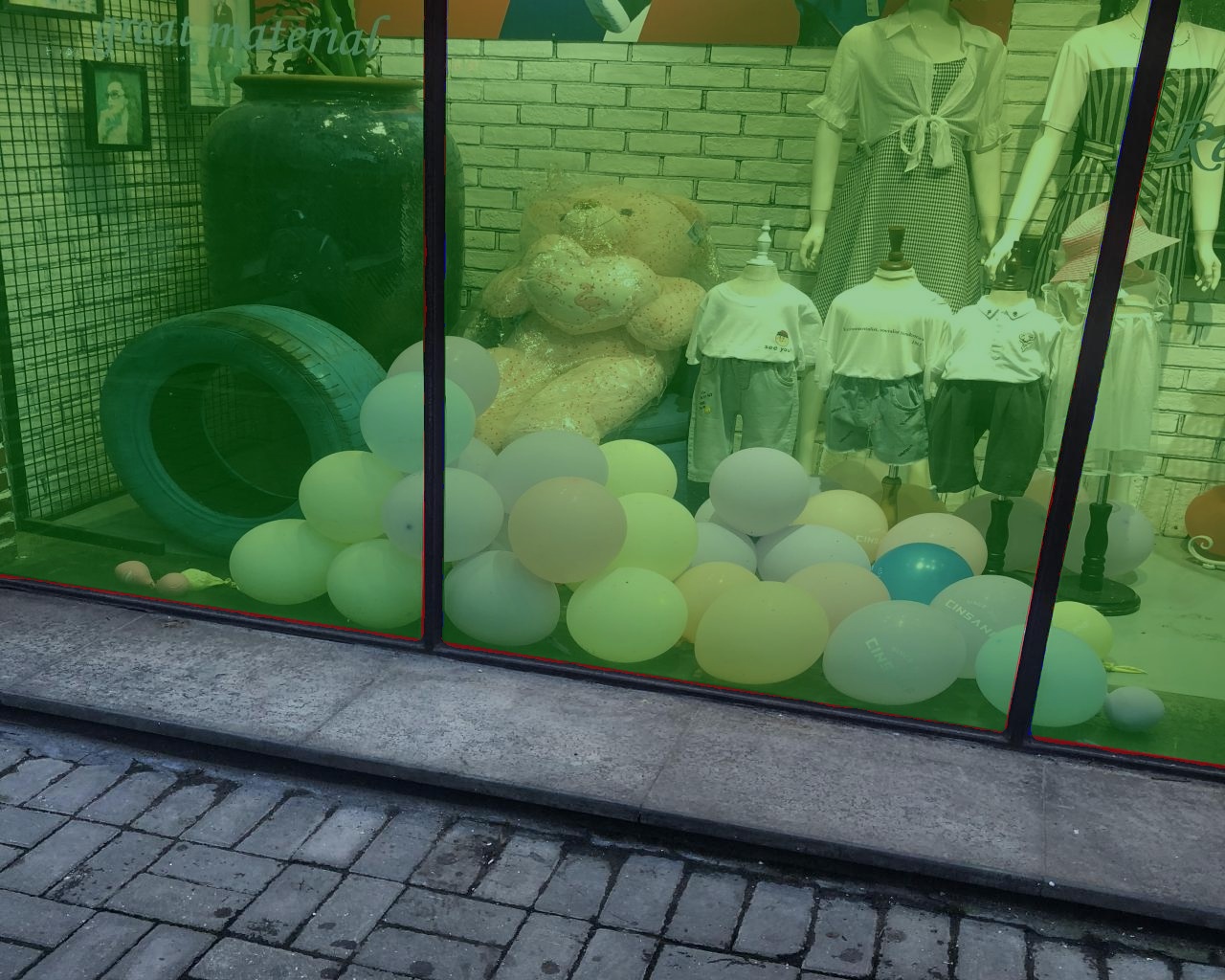} &
\includegraphics[width=\linewidth,height=\linewidth]{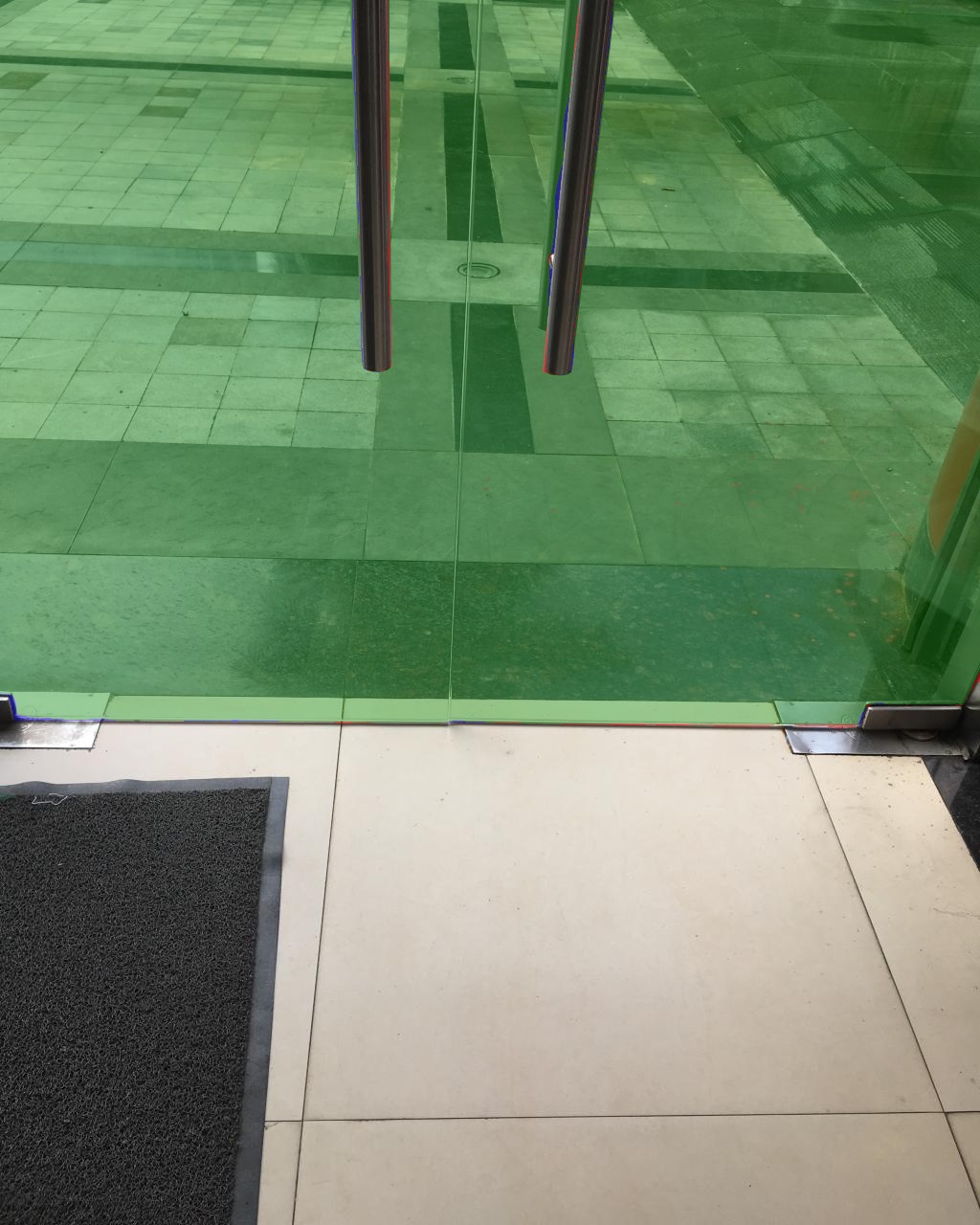} &
\includegraphics[width=\linewidth,height=\linewidth]{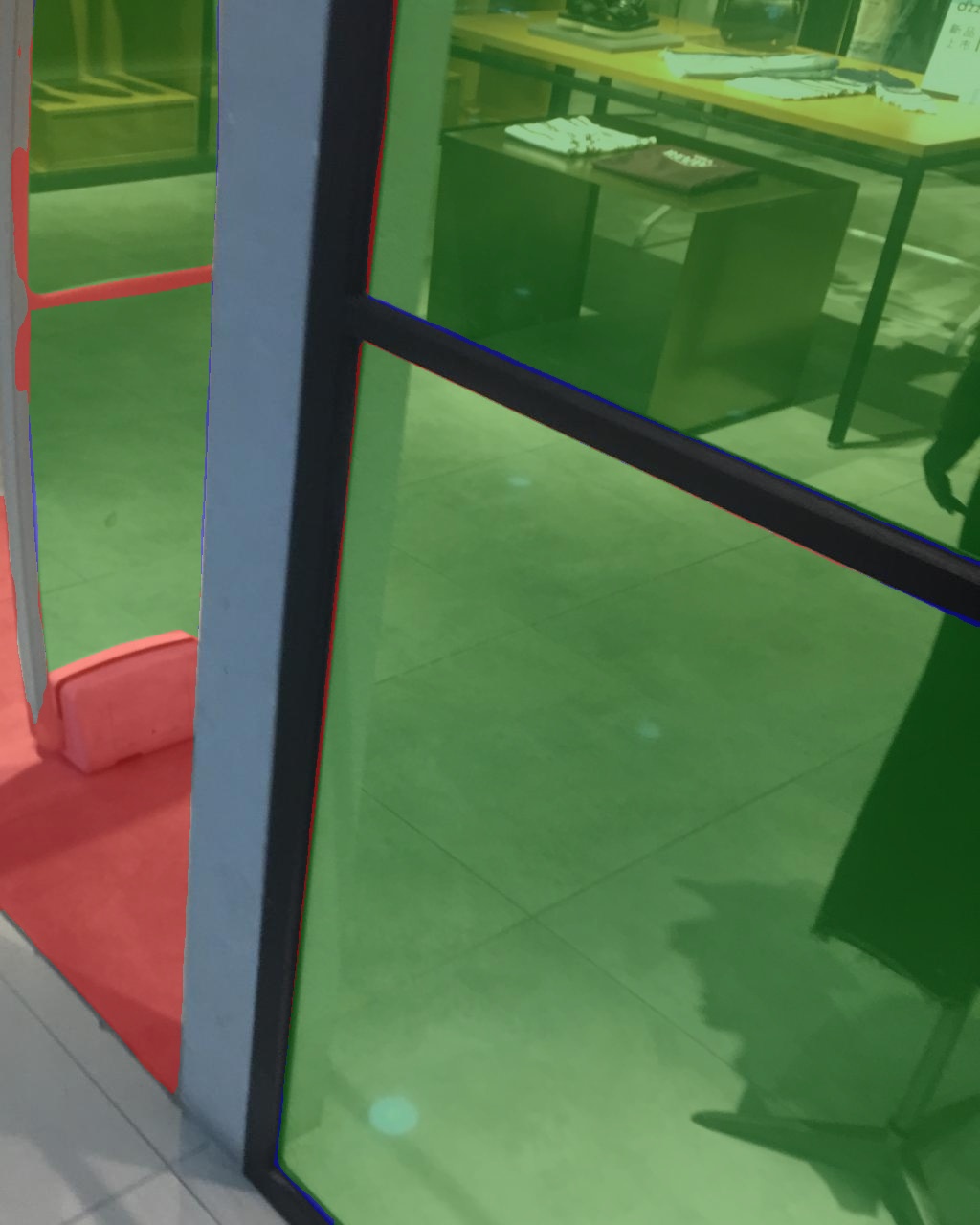} &
\includegraphics[width=\linewidth,height=\linewidth]{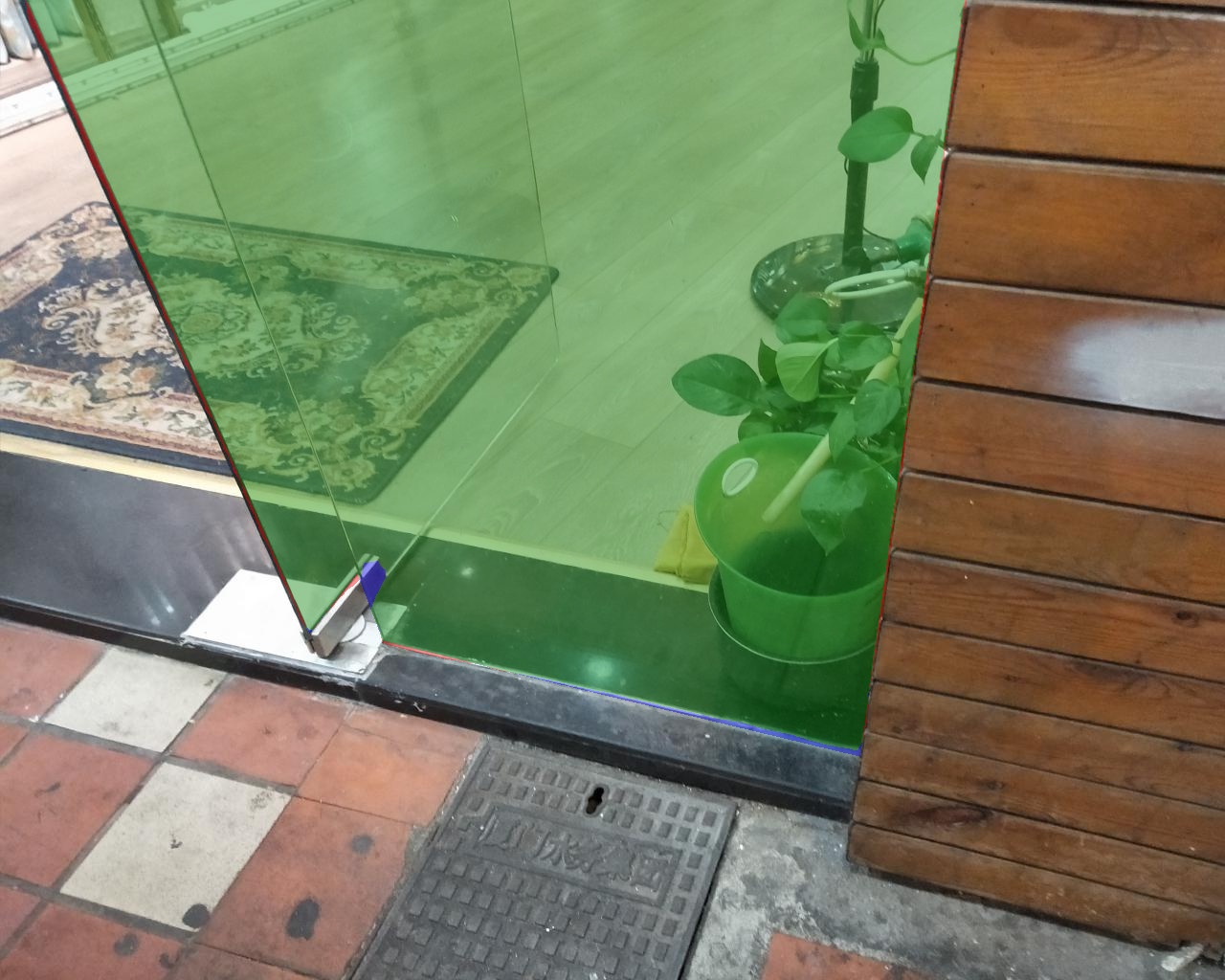} &
\includegraphics[width=\linewidth,height=\linewidth]{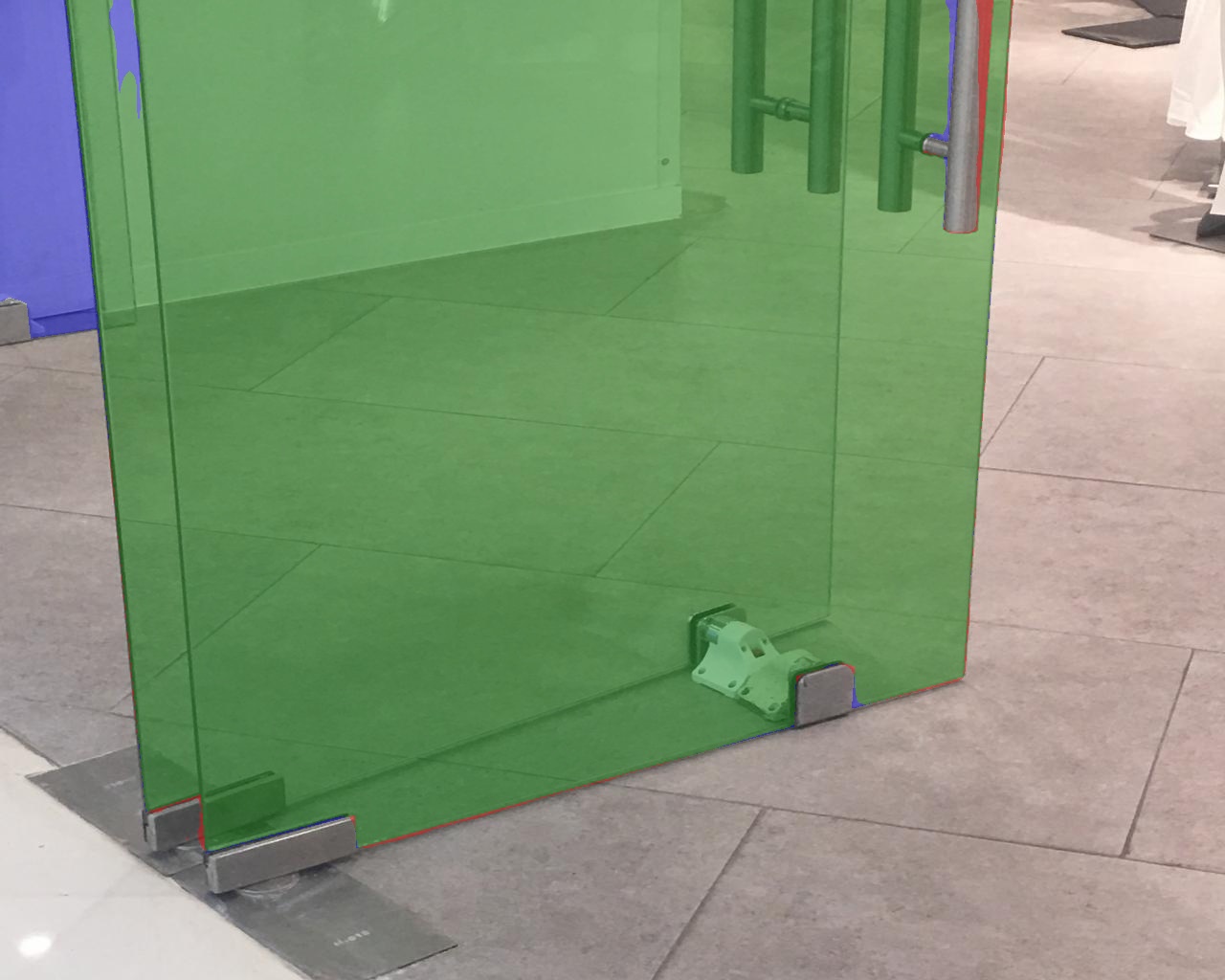} &
\includegraphics[width=\linewidth,height=\linewidth]{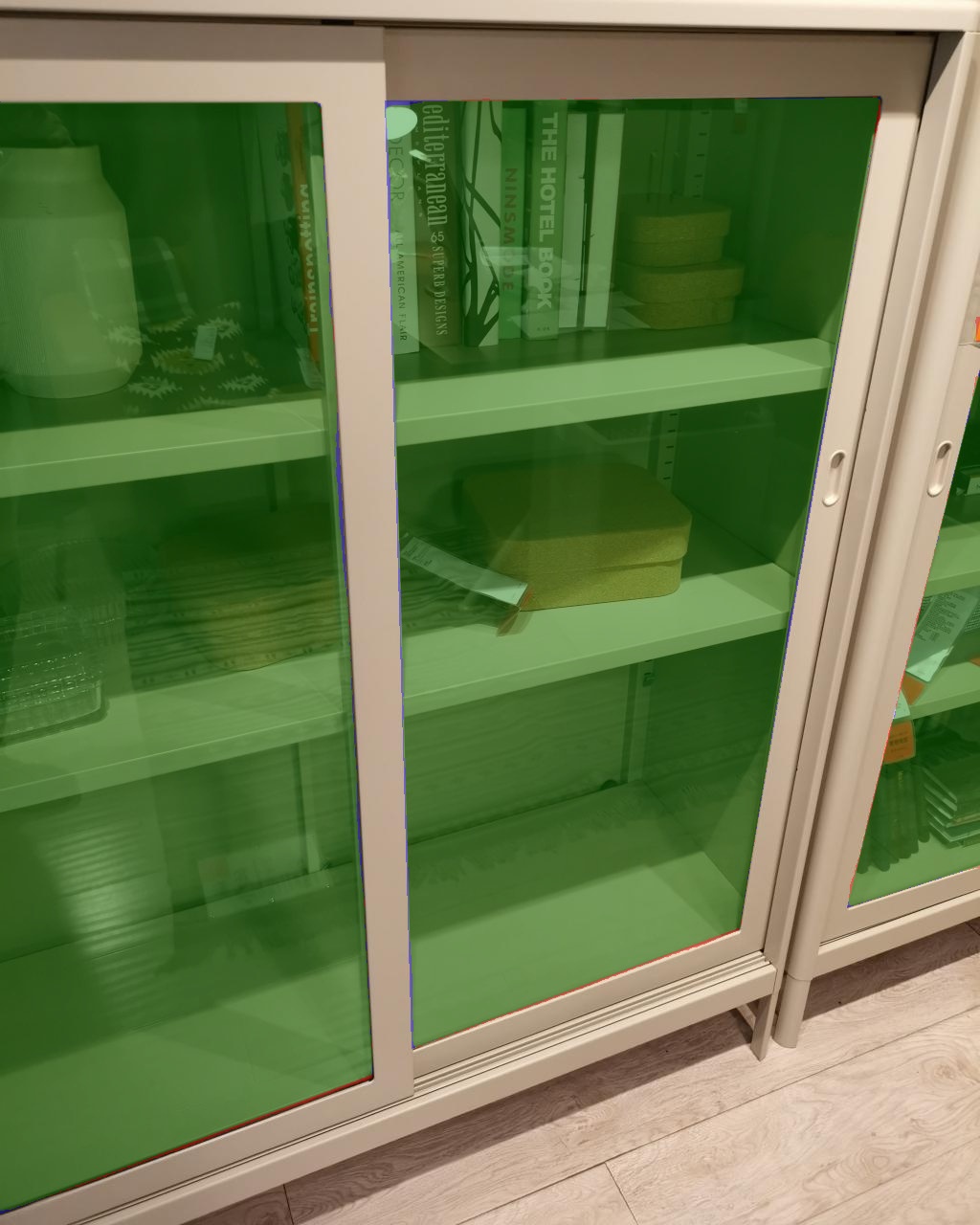} \\

\rotatebox{90}{GSD \cite{lin2021rich}} &
\includegraphics[width=\linewidth,height=\linewidth]{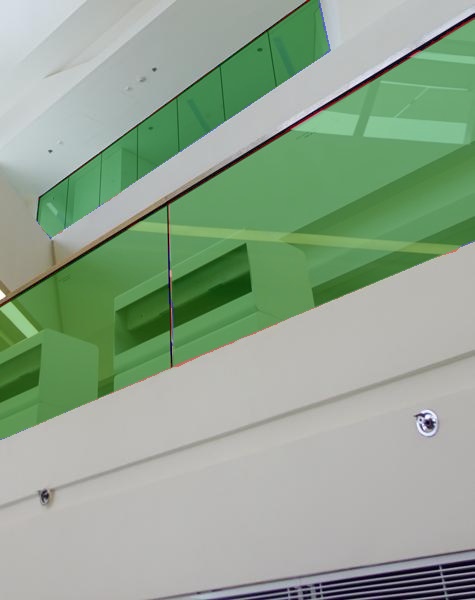} &
\includegraphics[width=\linewidth,height=\linewidth]{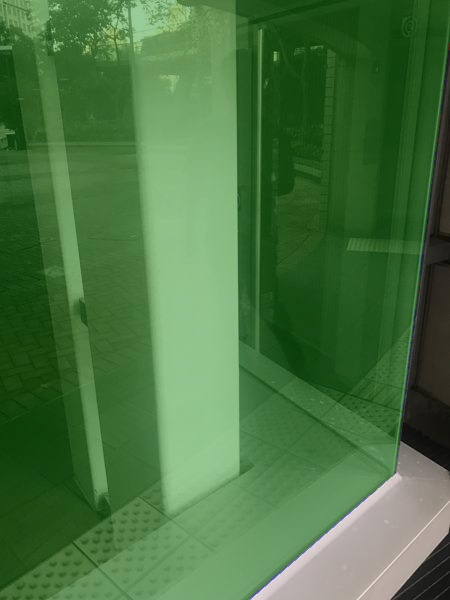} &
\includegraphics[width=\linewidth,height=\linewidth]{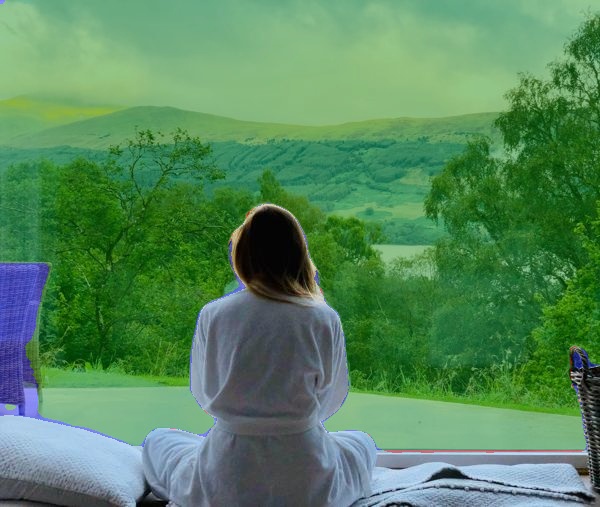} &
\includegraphics[width=\linewidth,height=\linewidth]{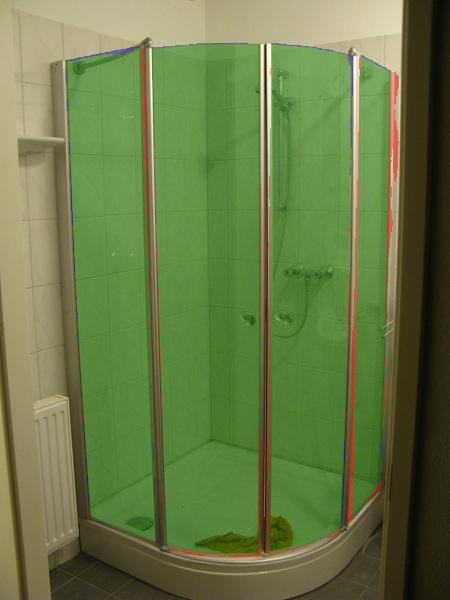} &
\includegraphics[width=\linewidth,height=\linewidth]{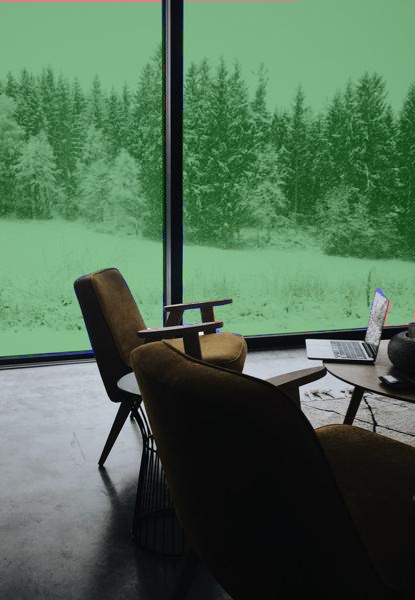} &
\includegraphics[width=\linewidth,height=\linewidth]{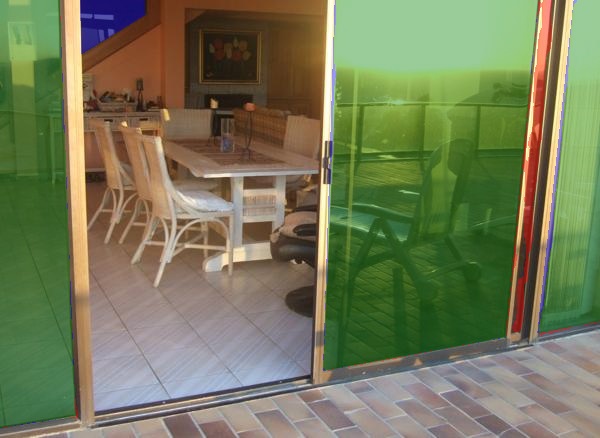} &
\includegraphics[width=\linewidth,height=\linewidth]{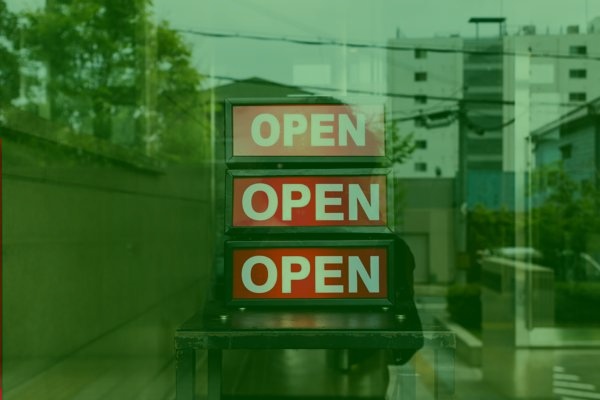} \\

\rotatebox{90}{\makecell{Trans10k-\\Stuff \cite{xie2020segmenting}}} &
\includegraphics[width=\linewidth,height=\linewidth]{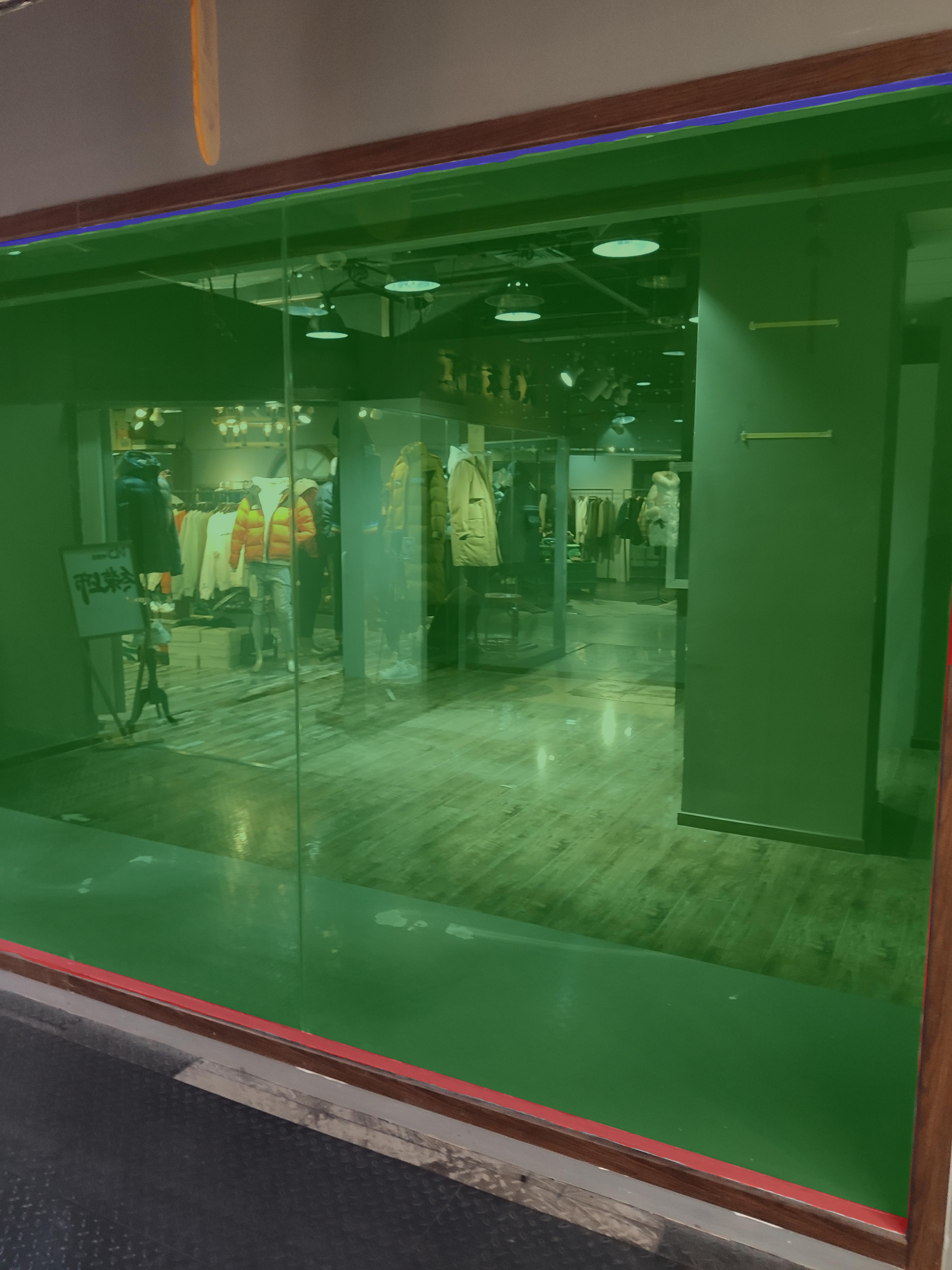} &
\includegraphics[width=\linewidth,height=\linewidth]{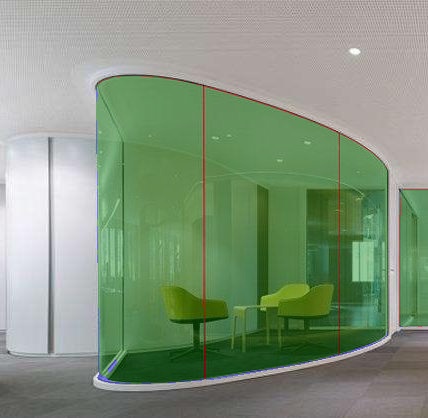} &
\includegraphics[width=\linewidth,height=\linewidth]{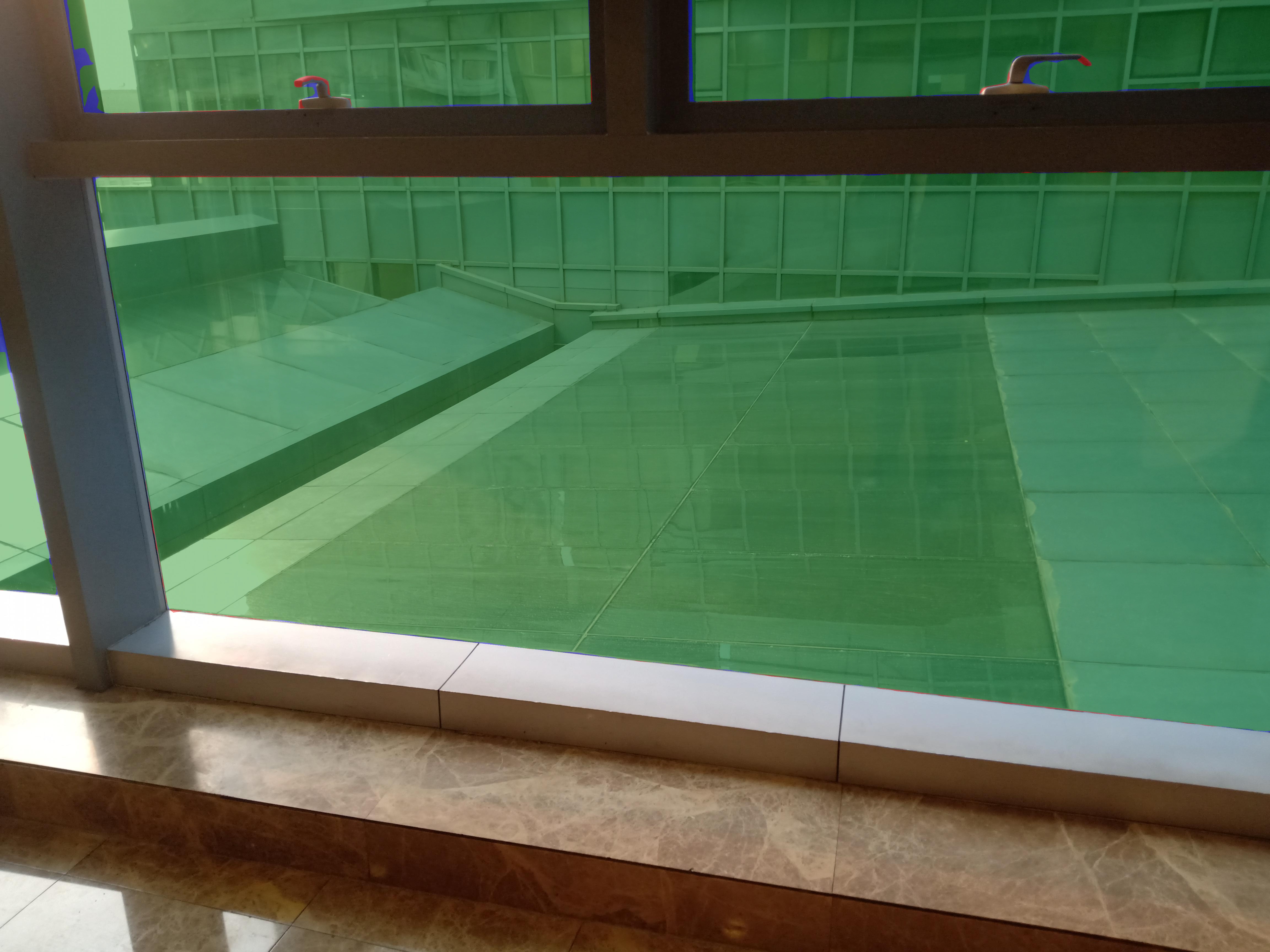} &
\includegraphics[width=\linewidth,height=\linewidth]{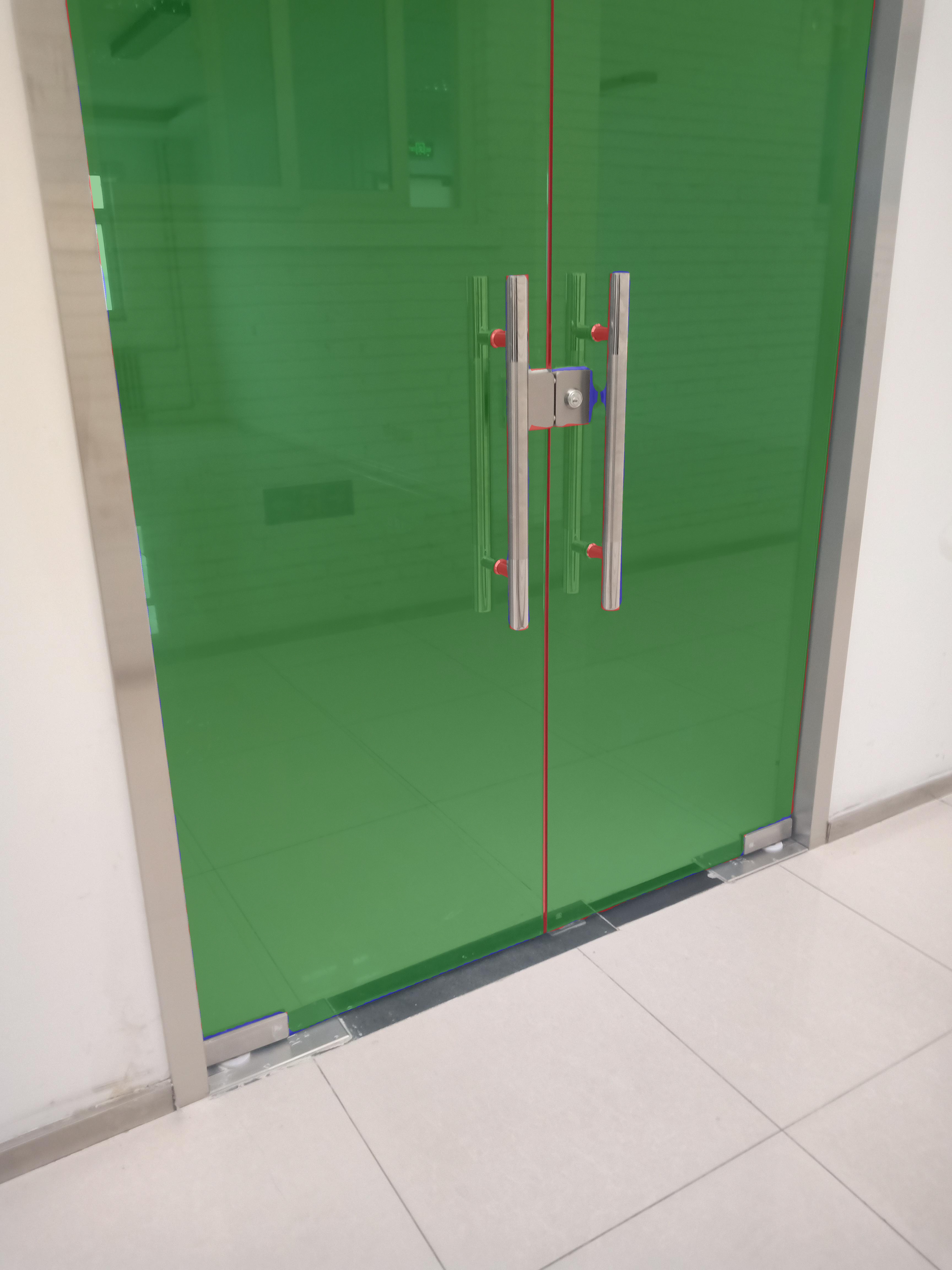} &
\includegraphics[width=\linewidth,height=\linewidth]{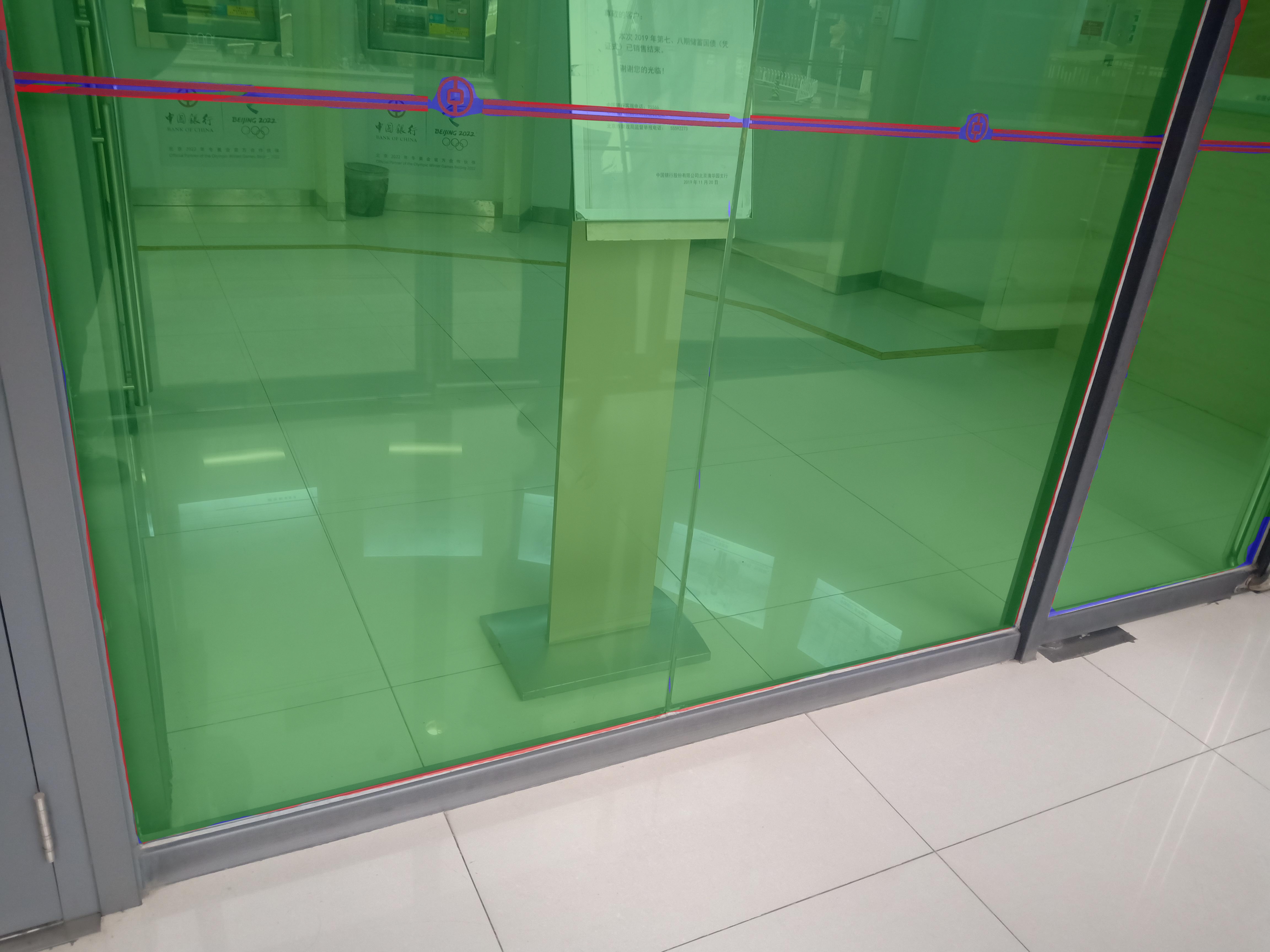} &
\includegraphics[width=\linewidth,height=\linewidth]{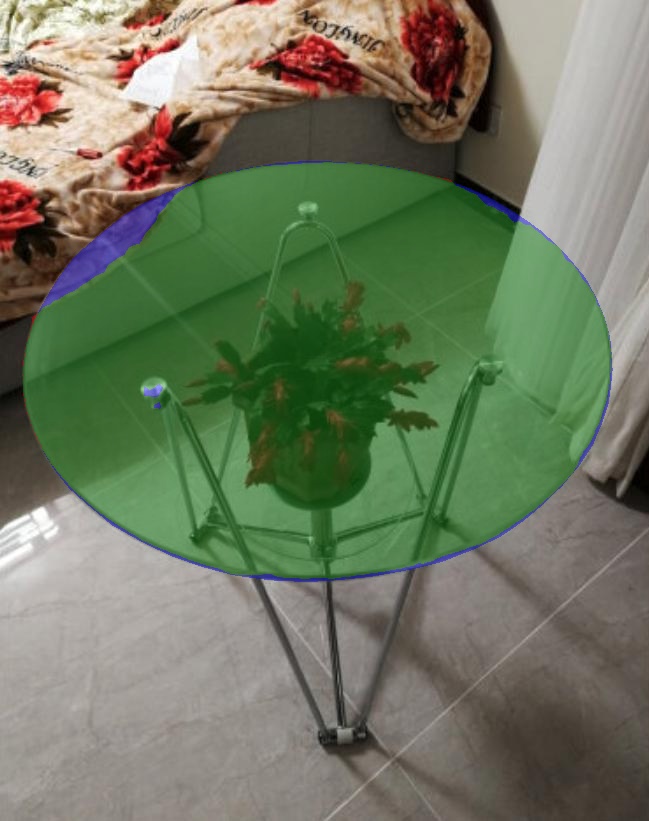} &
\includegraphics[width=\linewidth,height=\linewidth]{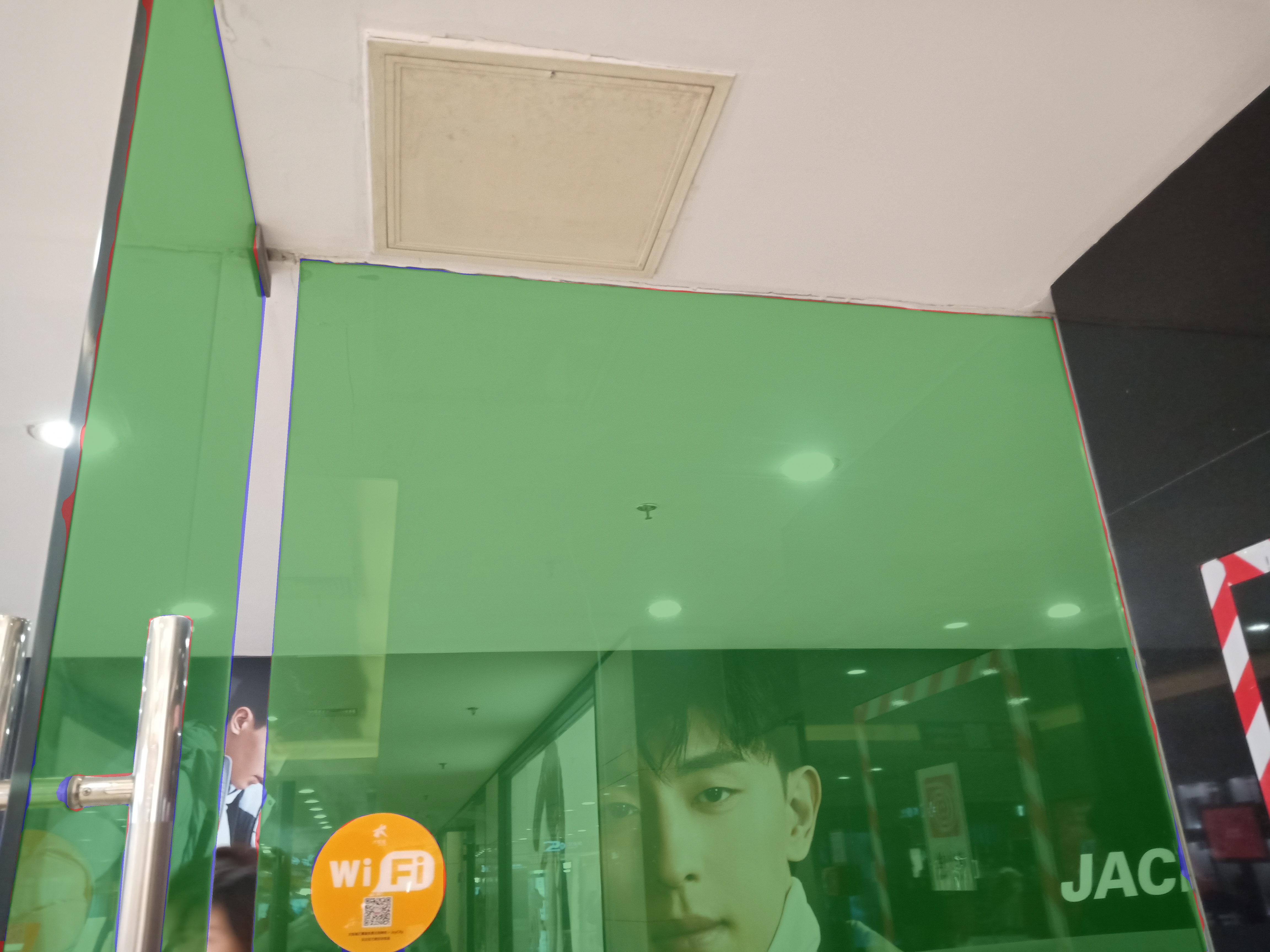} \\

\rotatebox{90}{HSO \cite{yu2022progressive}} &
\includegraphics[width=\linewidth,height=\linewidth]{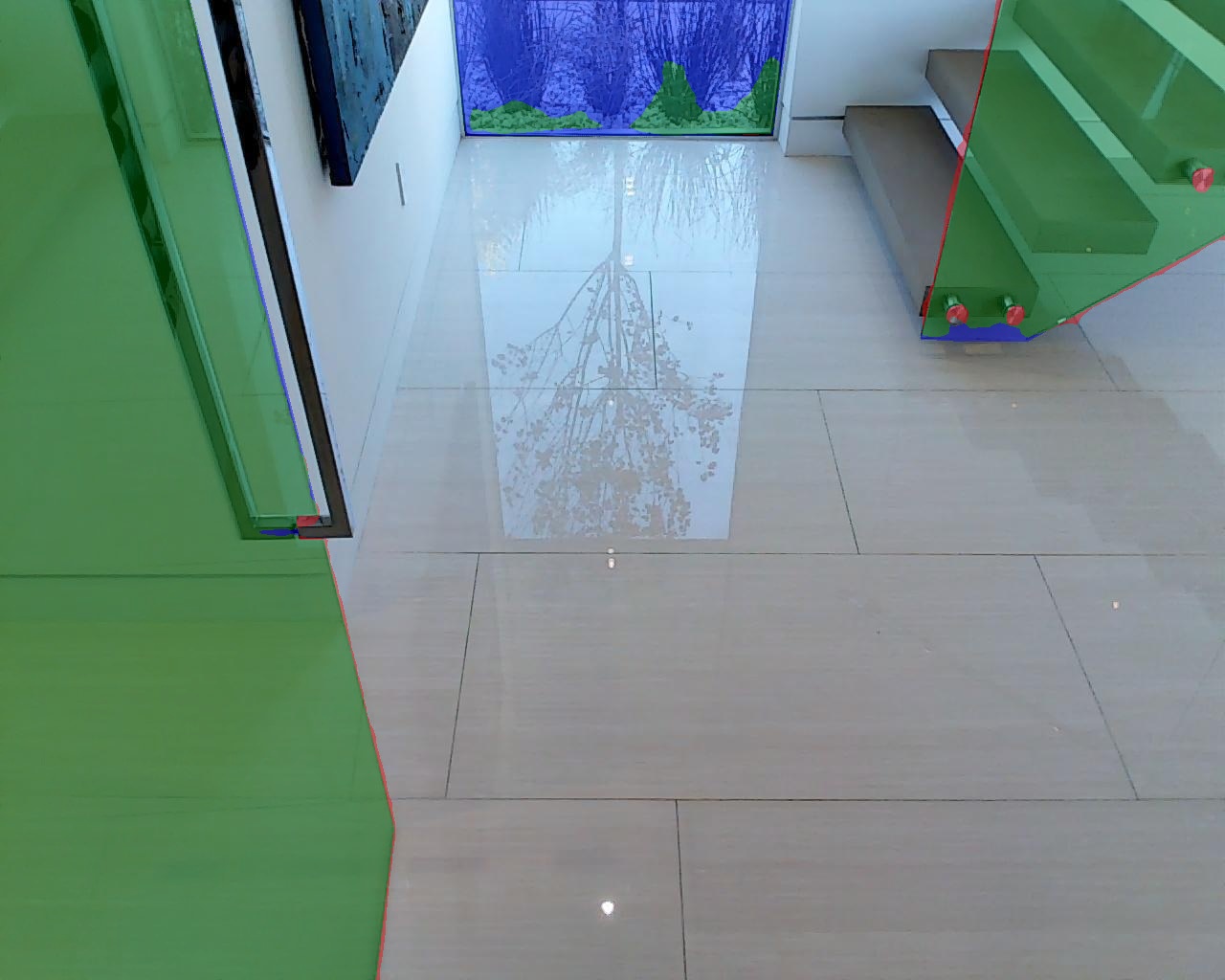} &
\includegraphics[width=\linewidth,height=\linewidth]{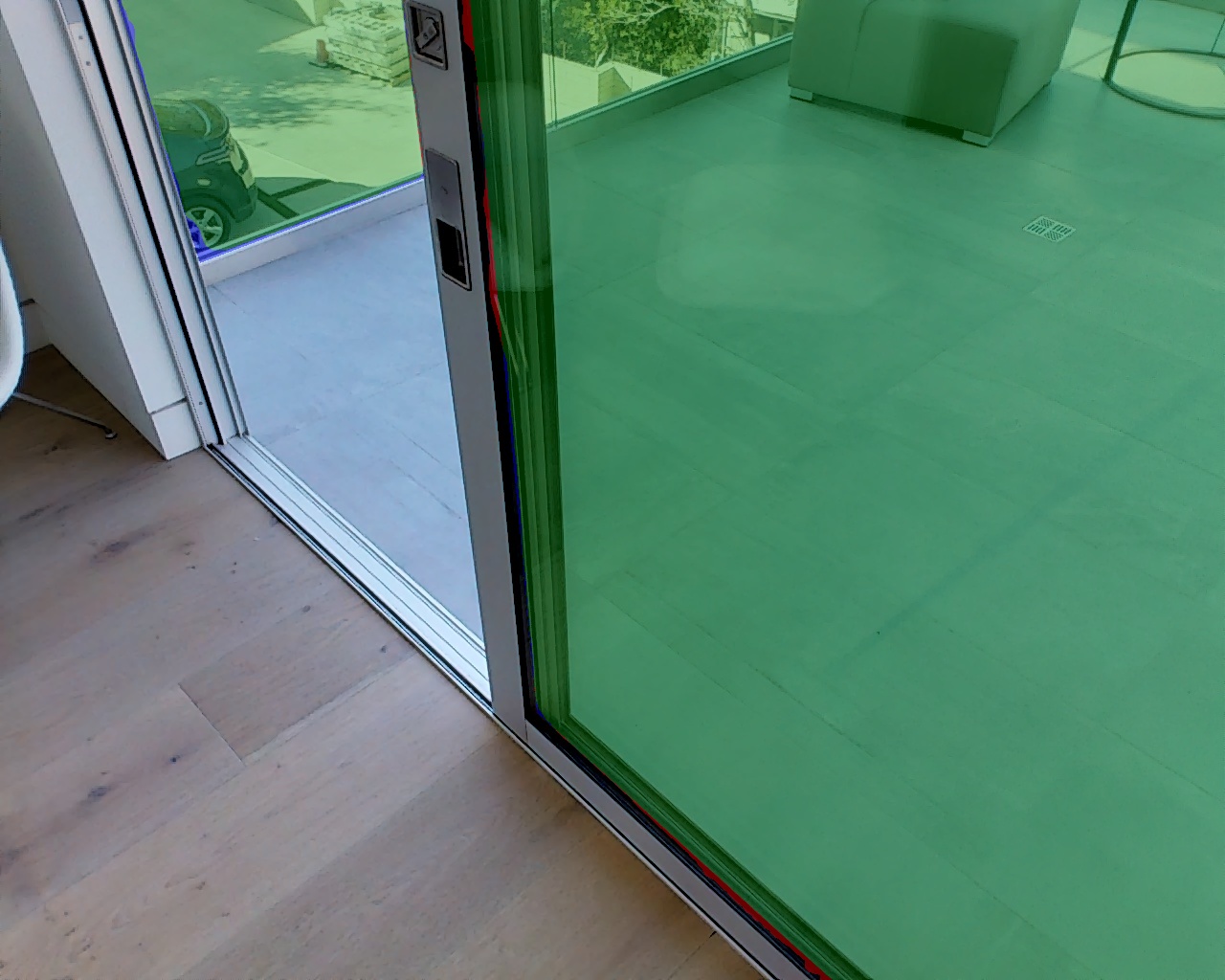} &
\includegraphics[width=\linewidth,height=\linewidth]{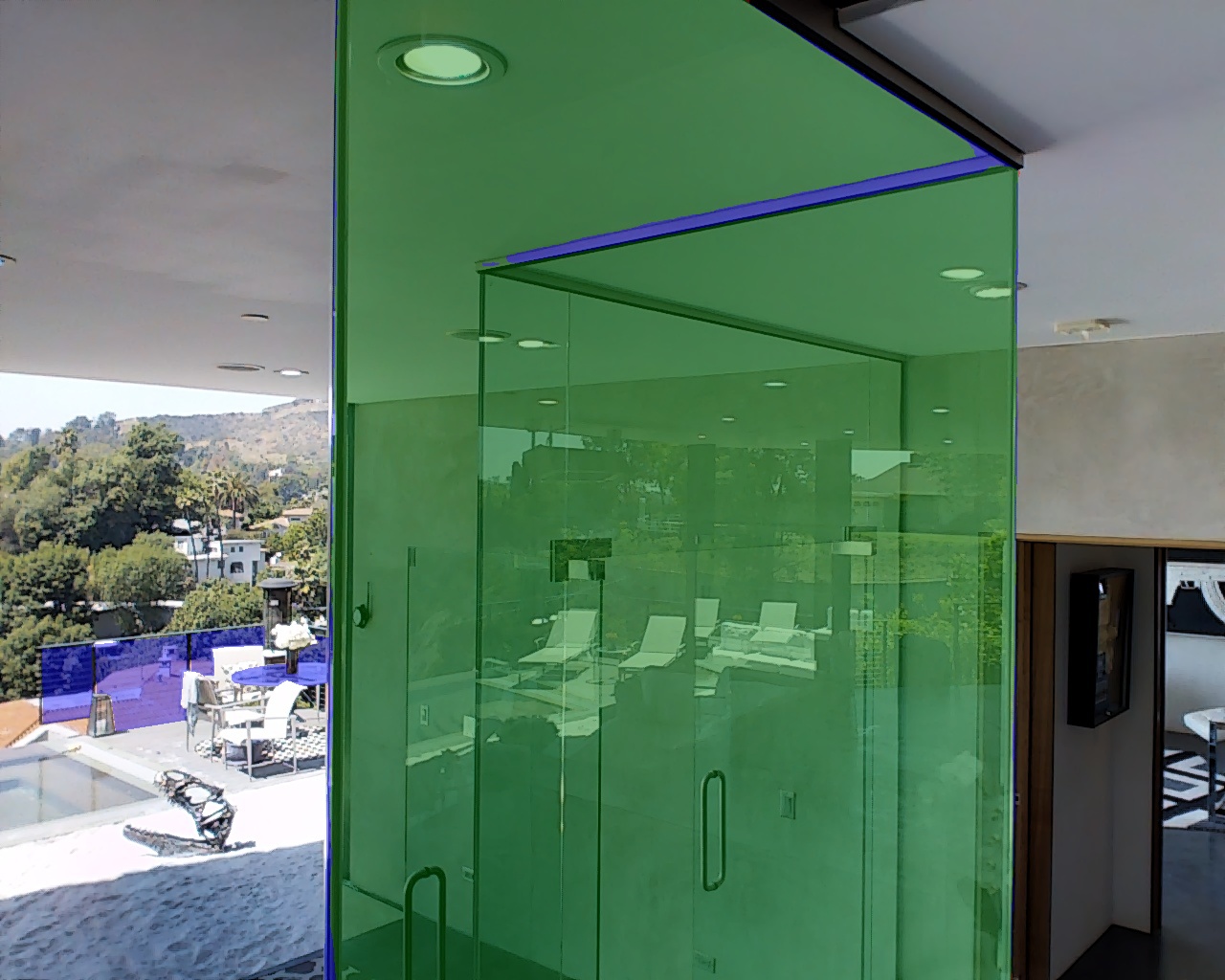} &
\includegraphics[width=\linewidth,height=\linewidth]{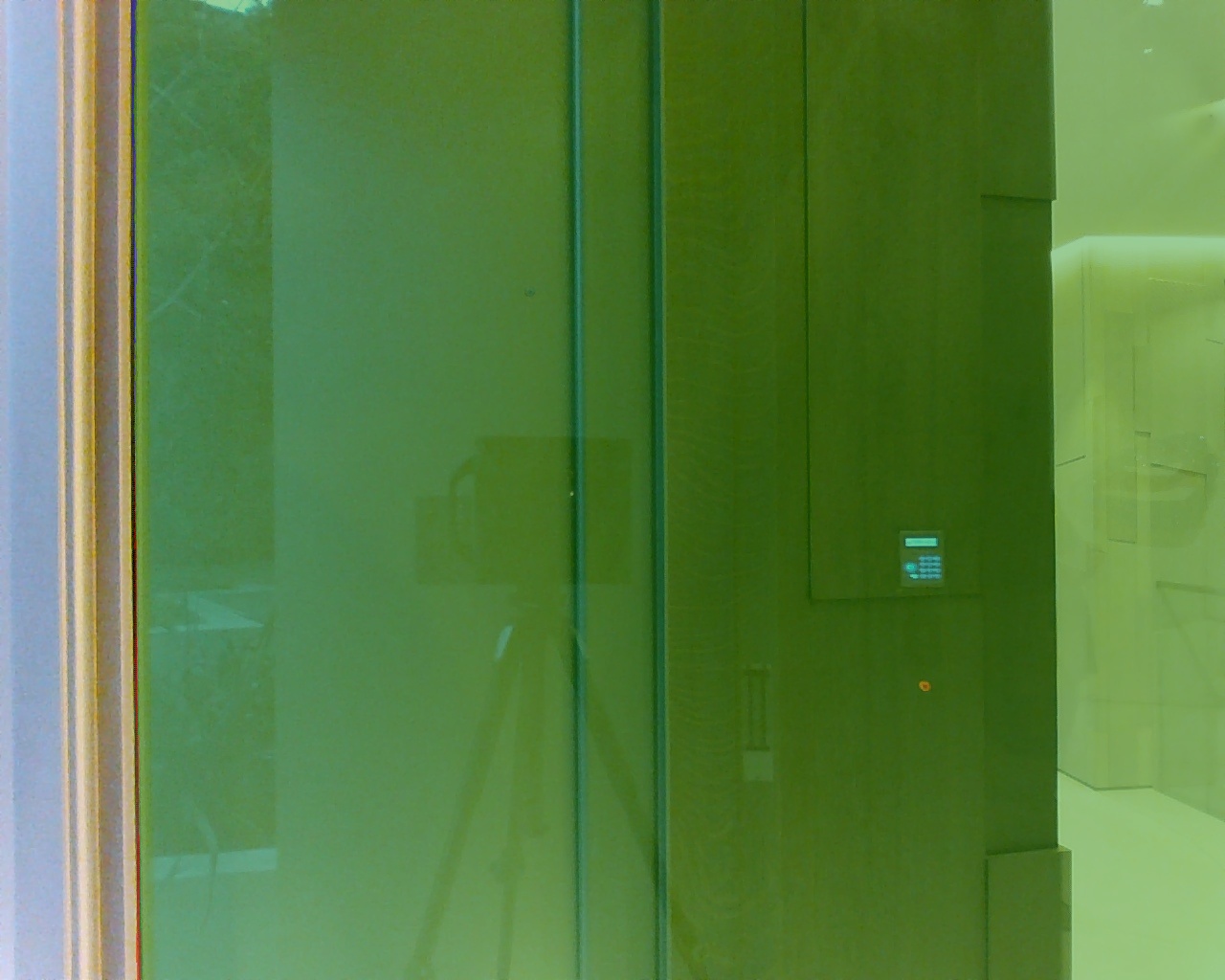} &
\includegraphics[width=\linewidth,height=\linewidth]{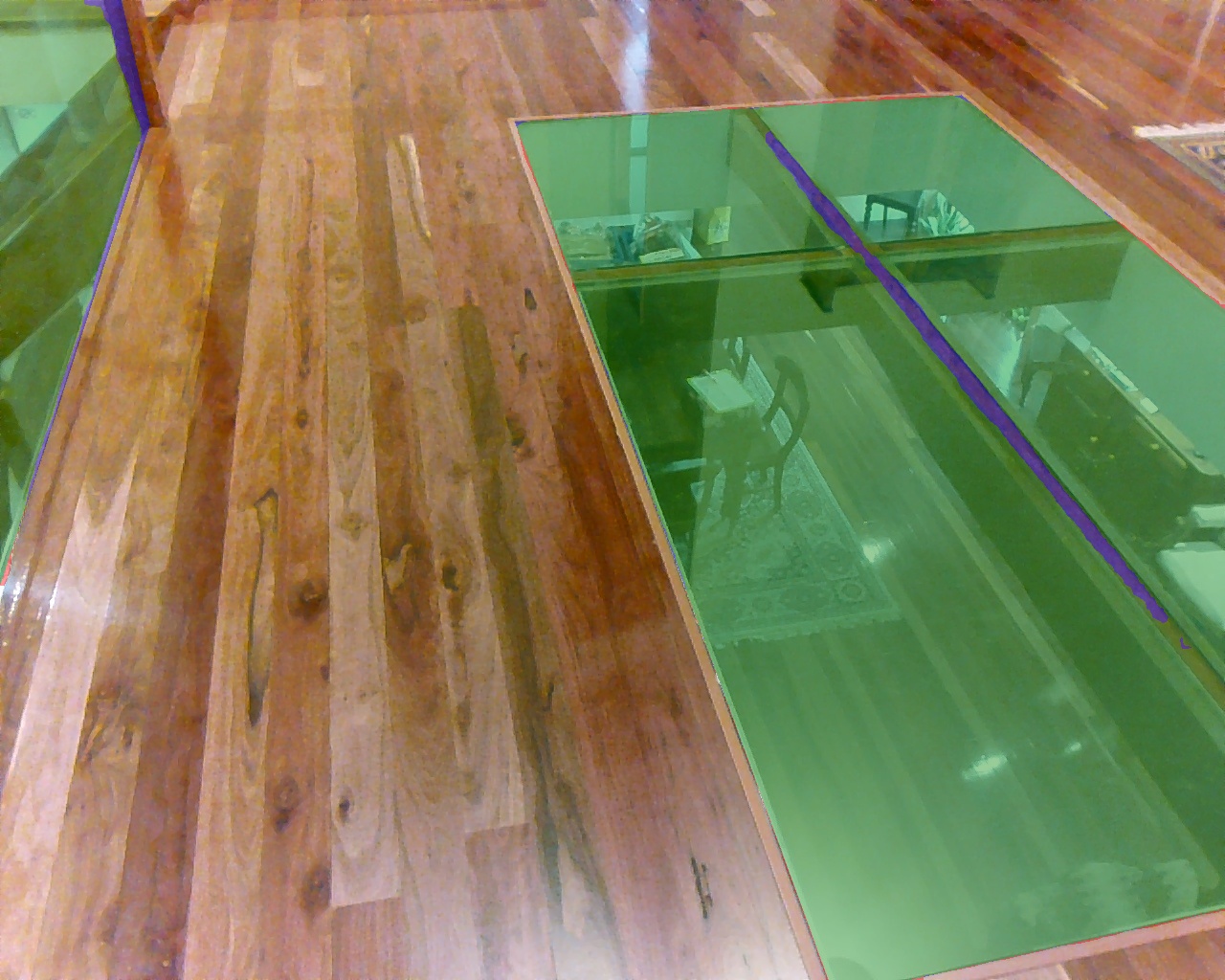} &
\includegraphics[width=\linewidth,height=\linewidth]{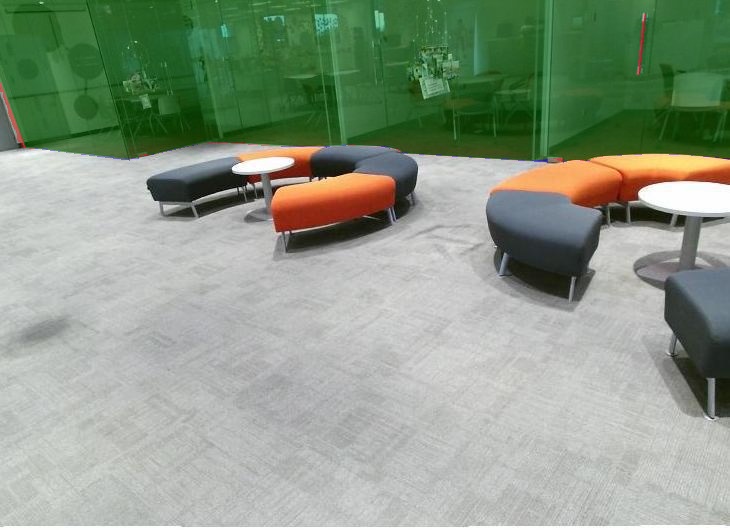} &
\includegraphics[width=\linewidth,height=\linewidth]{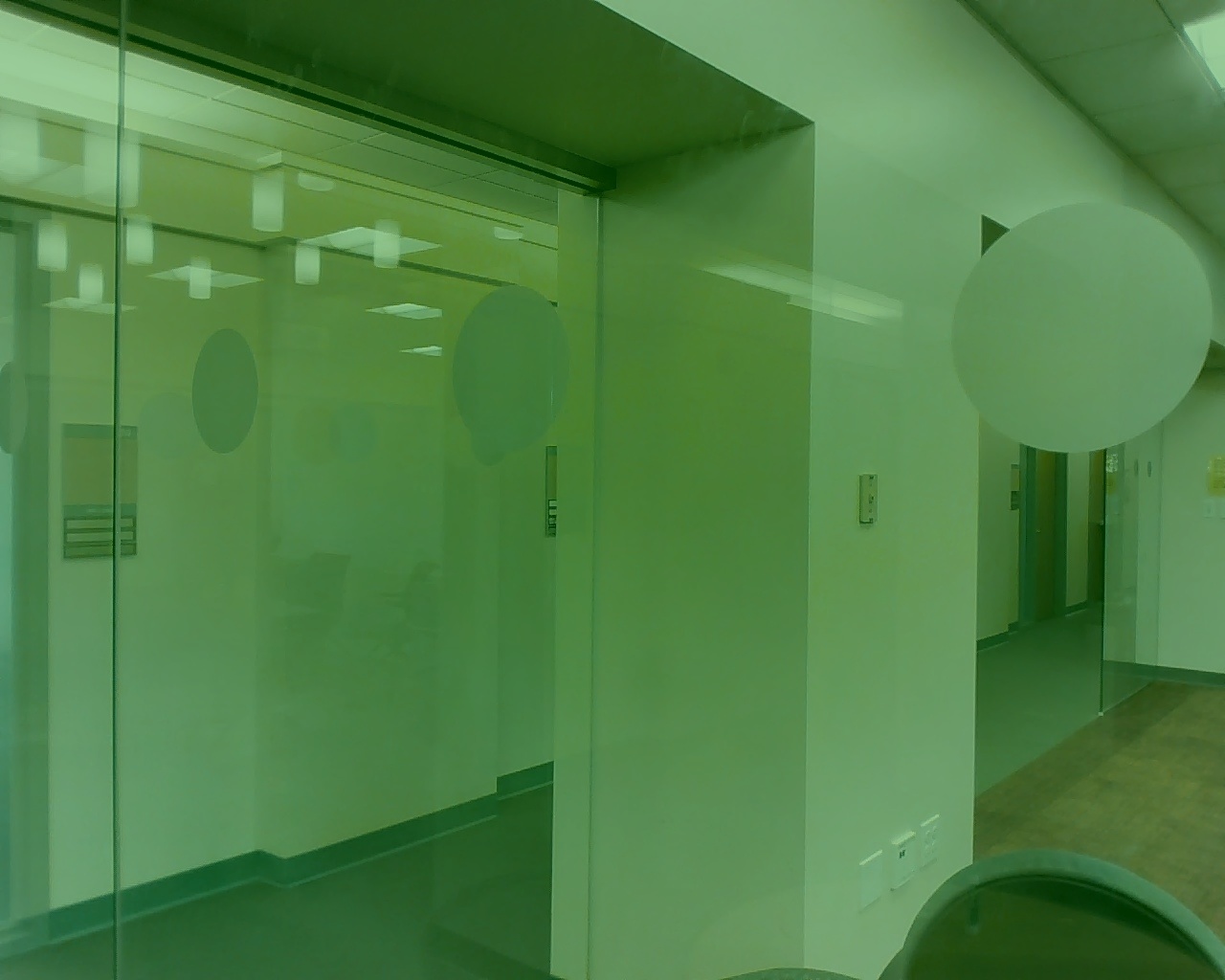} 

\end{tabular}

\caption{Samples of segmentation masks produced by L+GNet on the different testing sets. Model trained with the combined training sets. True positives overlaid in green, false positives overlaid in red, and false negatives overlaid in blue.}
\label{fig:samples_segmentation}
\end{figure*}

\begin{figure*}
\centering
\setlength{\tabcolsep}{3pt}
\scriptsize
\begin{tabular}{>{\centering\arraybackslash}m{0.5cm}|*{7}{>{\centering\arraybackslash}m{0.25\columnwidth}}}

\rotatebox{90}{\makecell{Original\\ image}} &
\includegraphics[width=\linewidth,height=\linewidth]{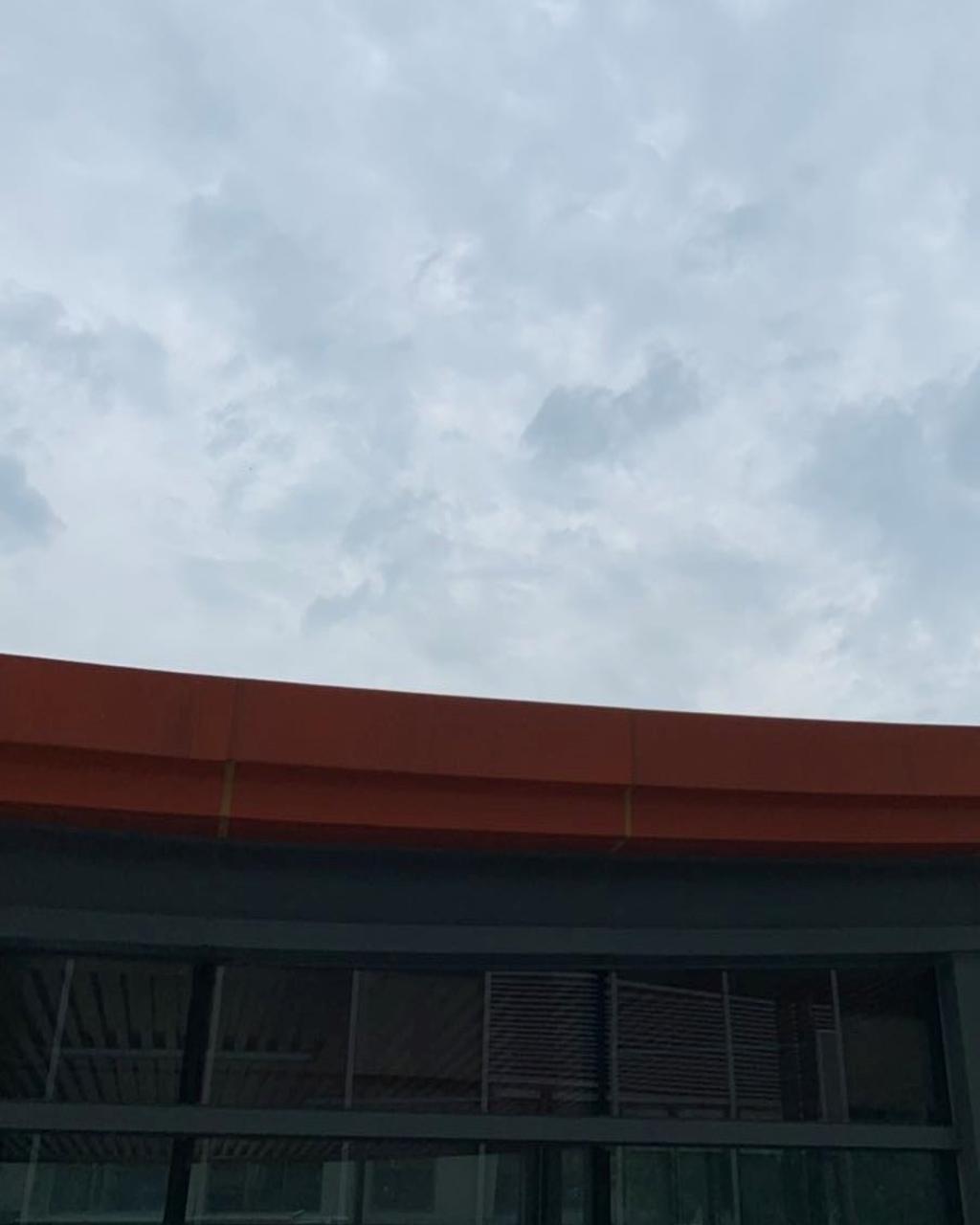} &
\includegraphics[width=\linewidth,height=\linewidth]{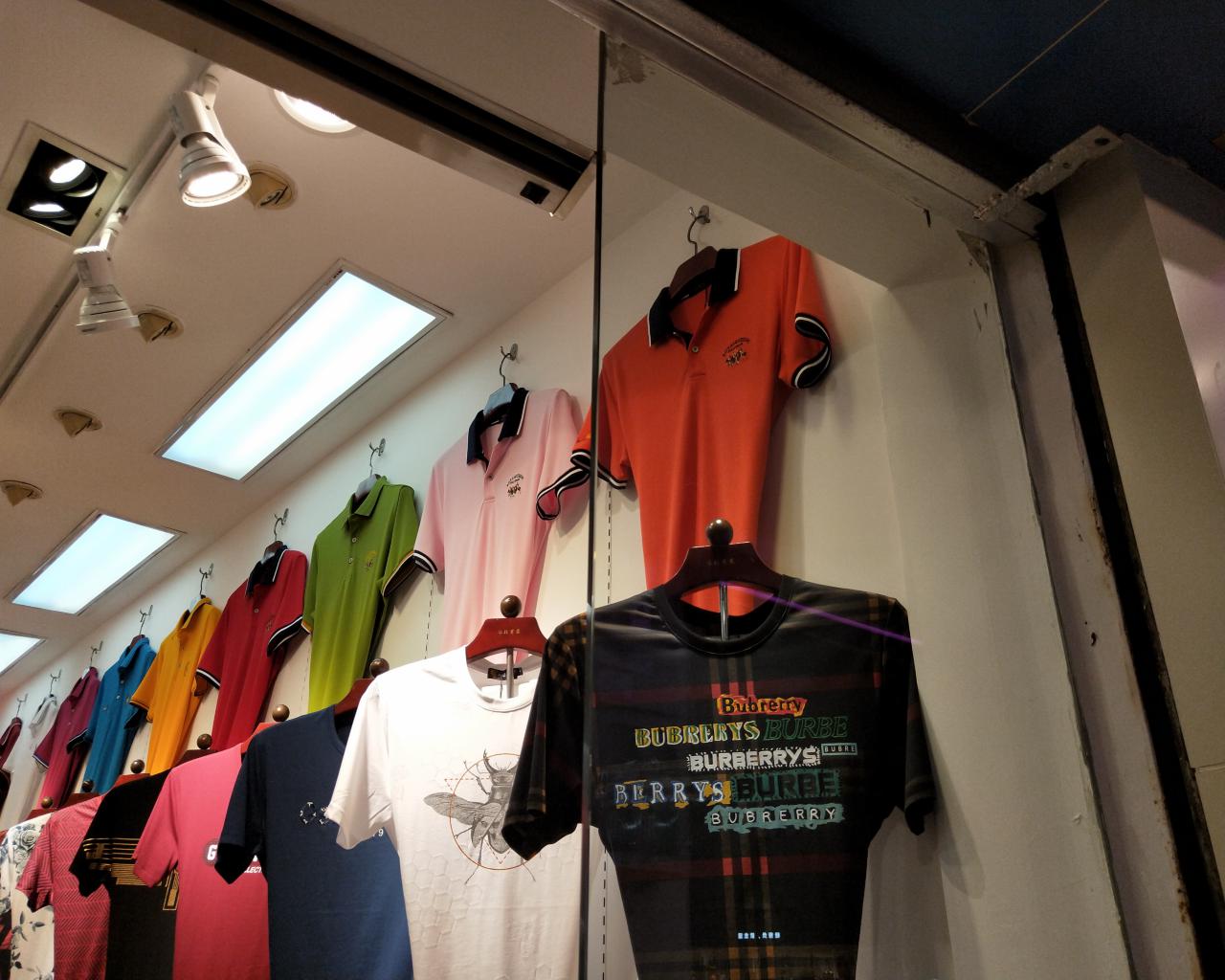} &
\includegraphics[width=\linewidth,height=\linewidth]{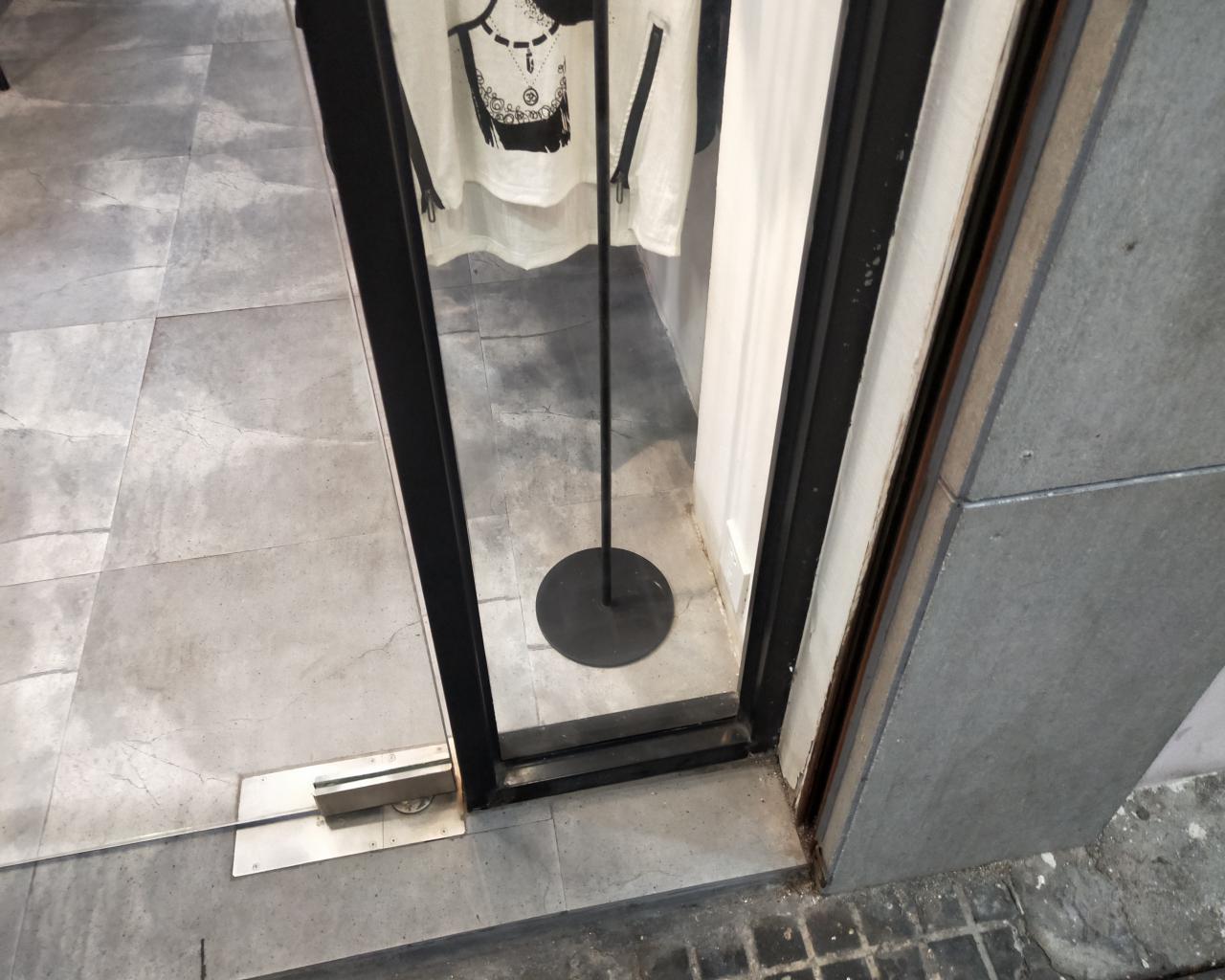} &
\includegraphics[width=\linewidth,height=\linewidth]{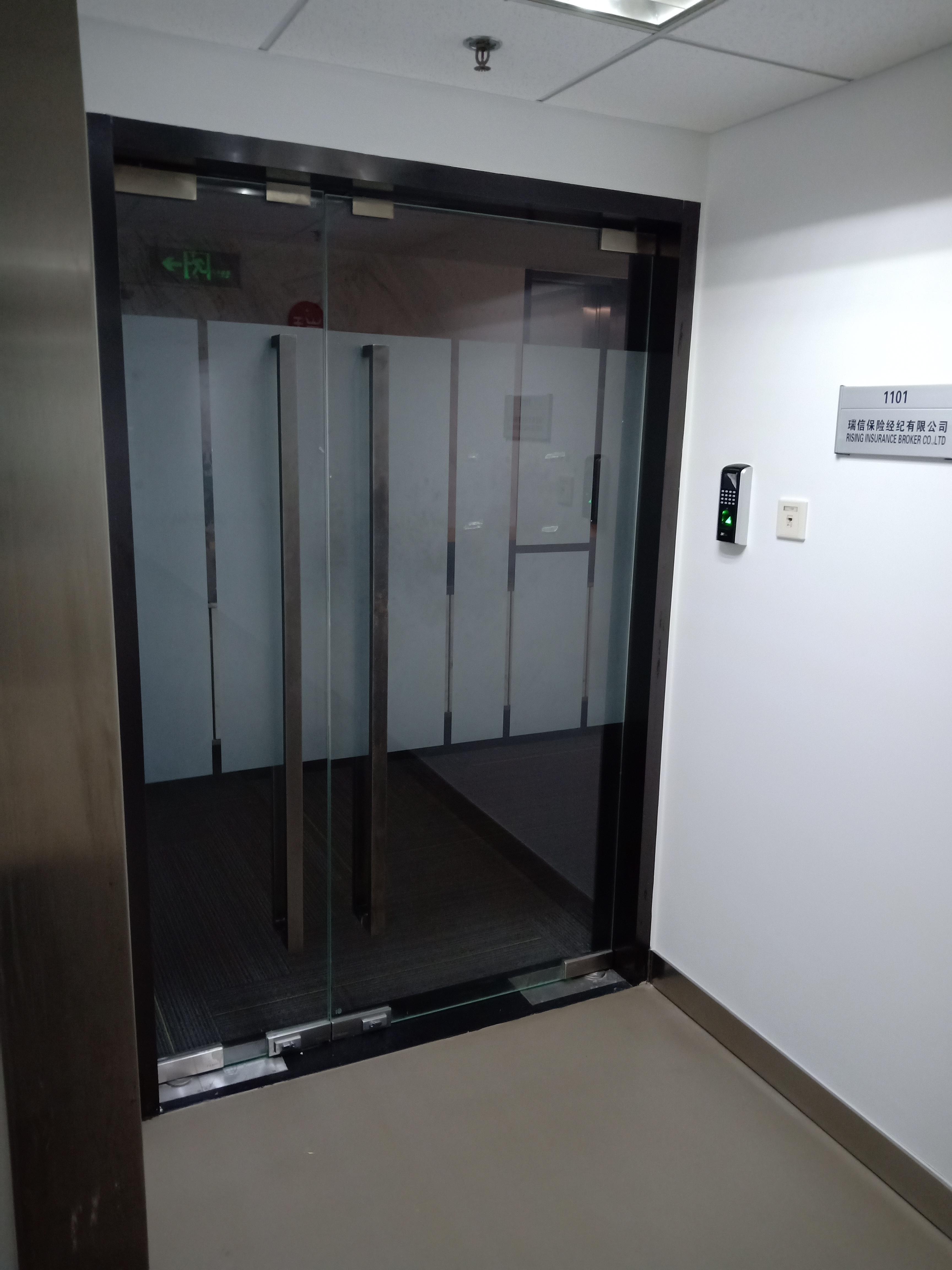} &
\includegraphics[width=\linewidth,height=\linewidth]{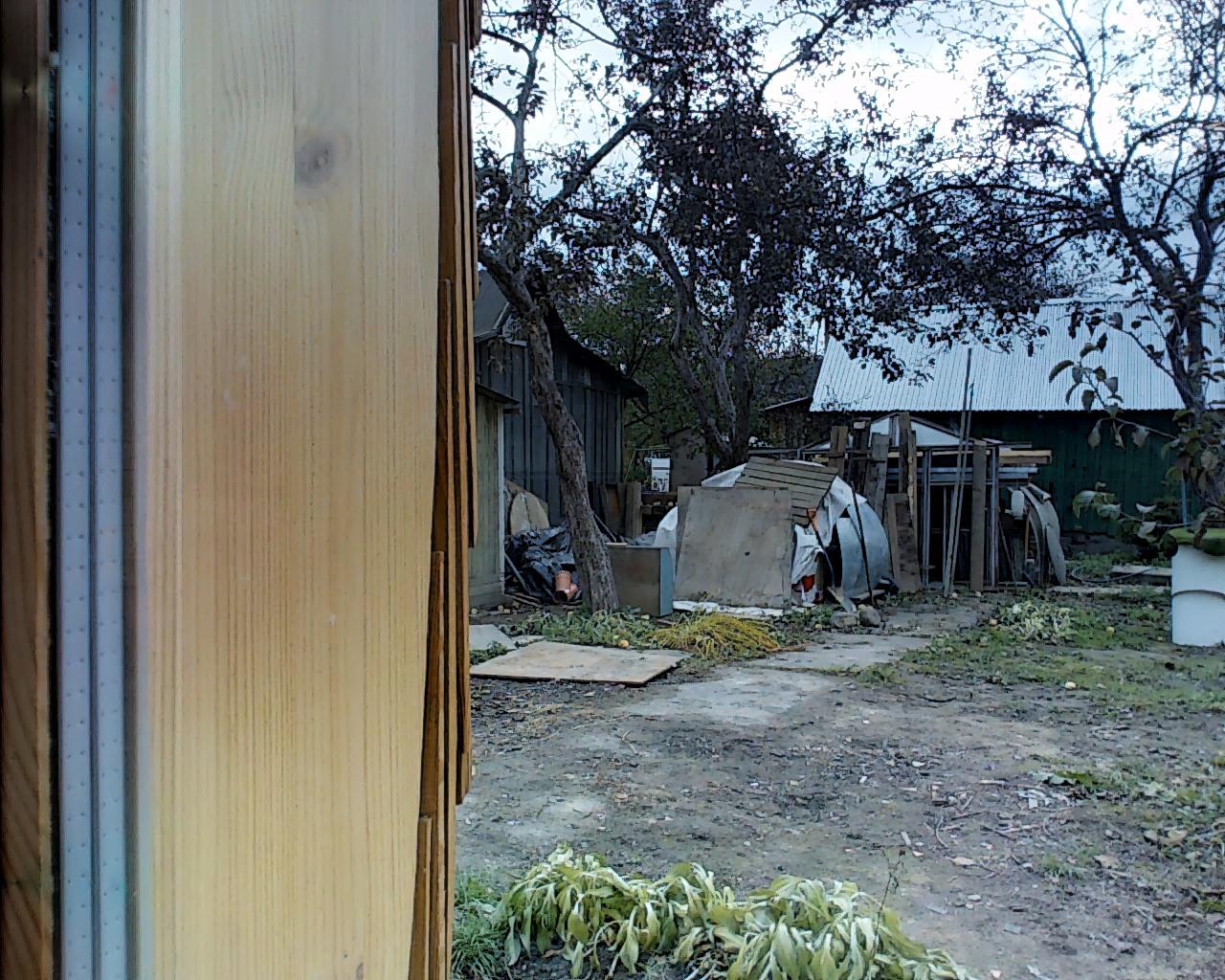} &
\includegraphics[width=\linewidth,height=\linewidth]{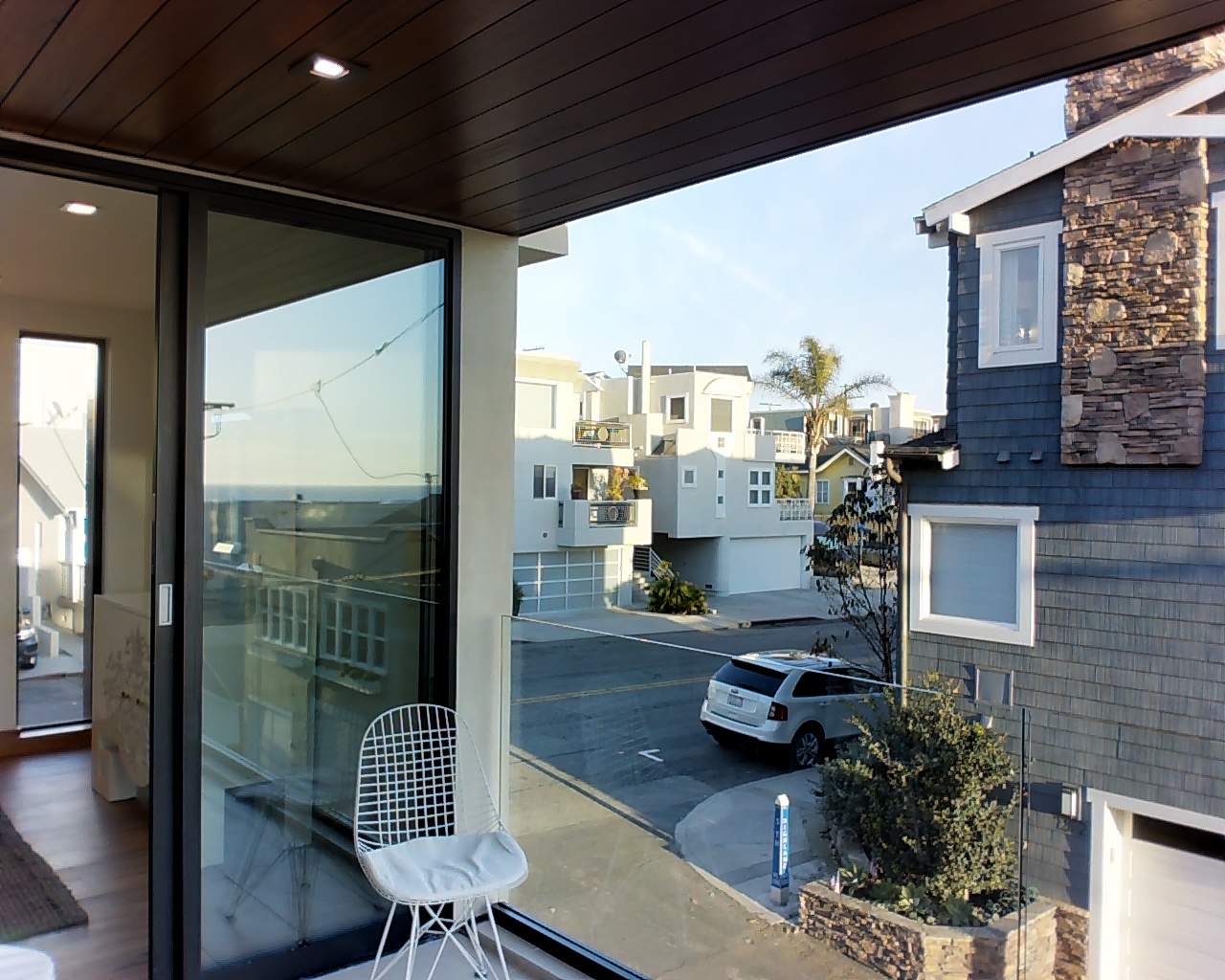} &
\includegraphics[width=\linewidth,height=\linewidth]{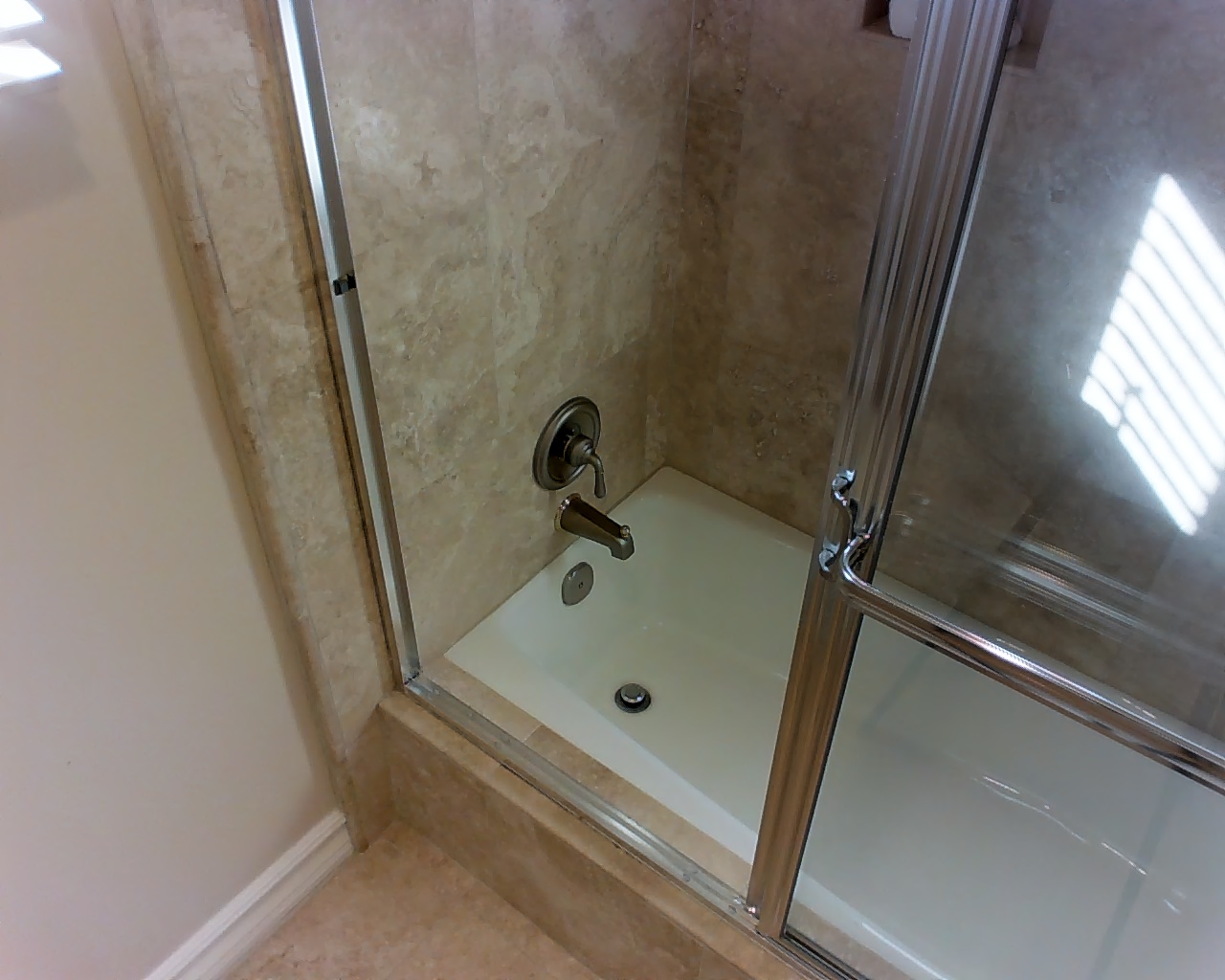} \\

\rotatebox{90}{\makecell{L+GNet}} &
\includegraphics[width=\linewidth,height=\linewidth]{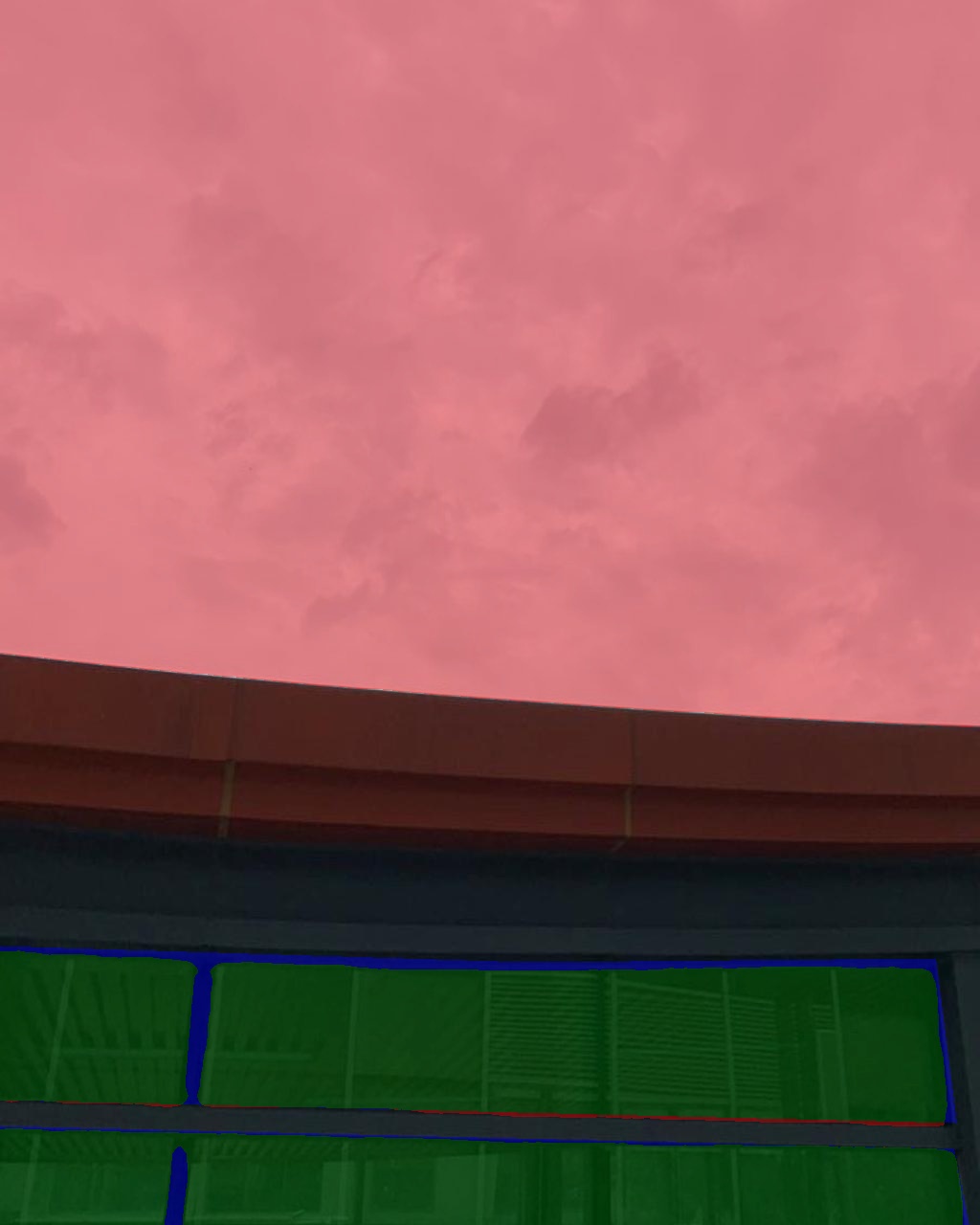} &
\includegraphics[width=\linewidth,height=\linewidth]{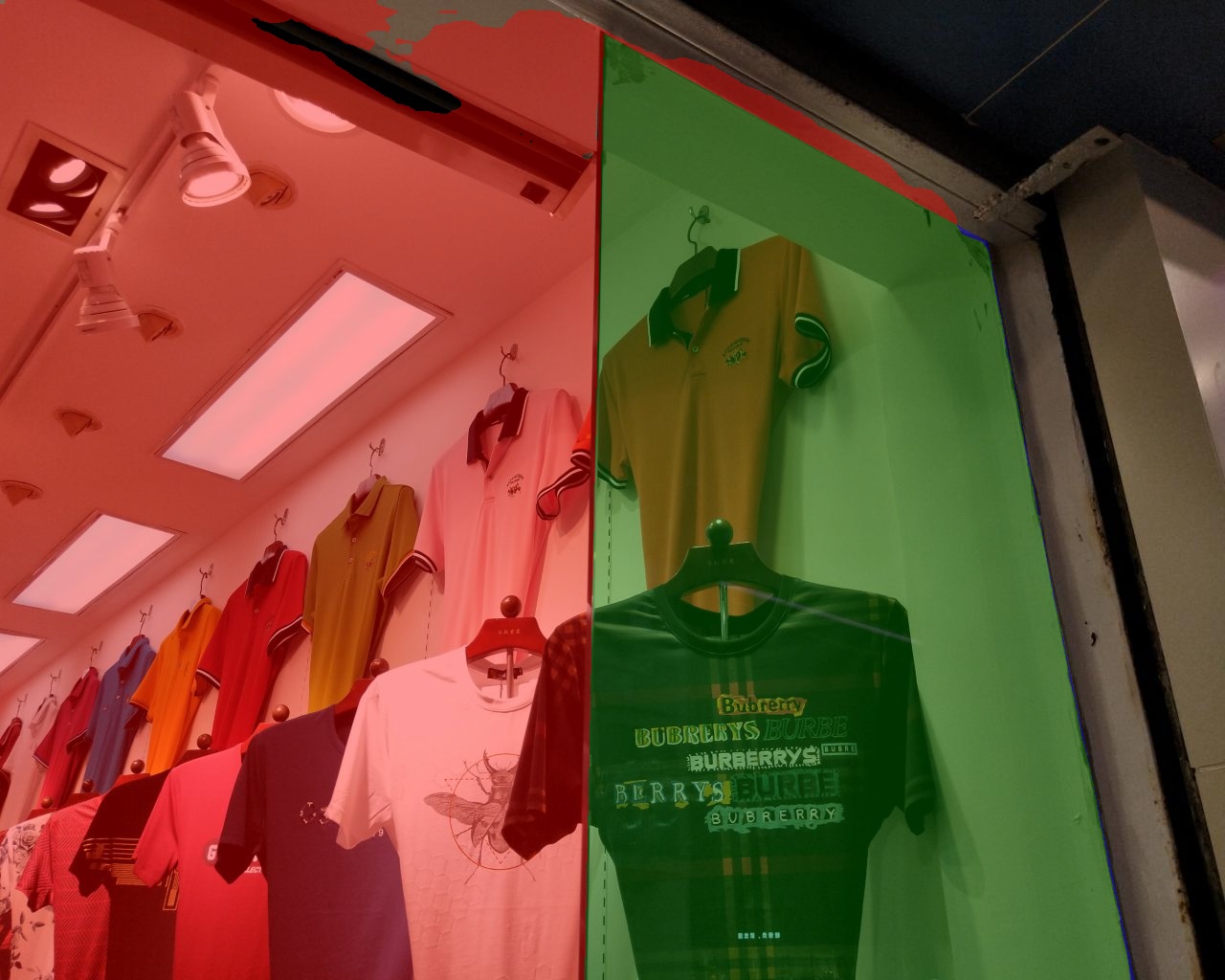} &
\includegraphics[width=\linewidth,height=\linewidth]{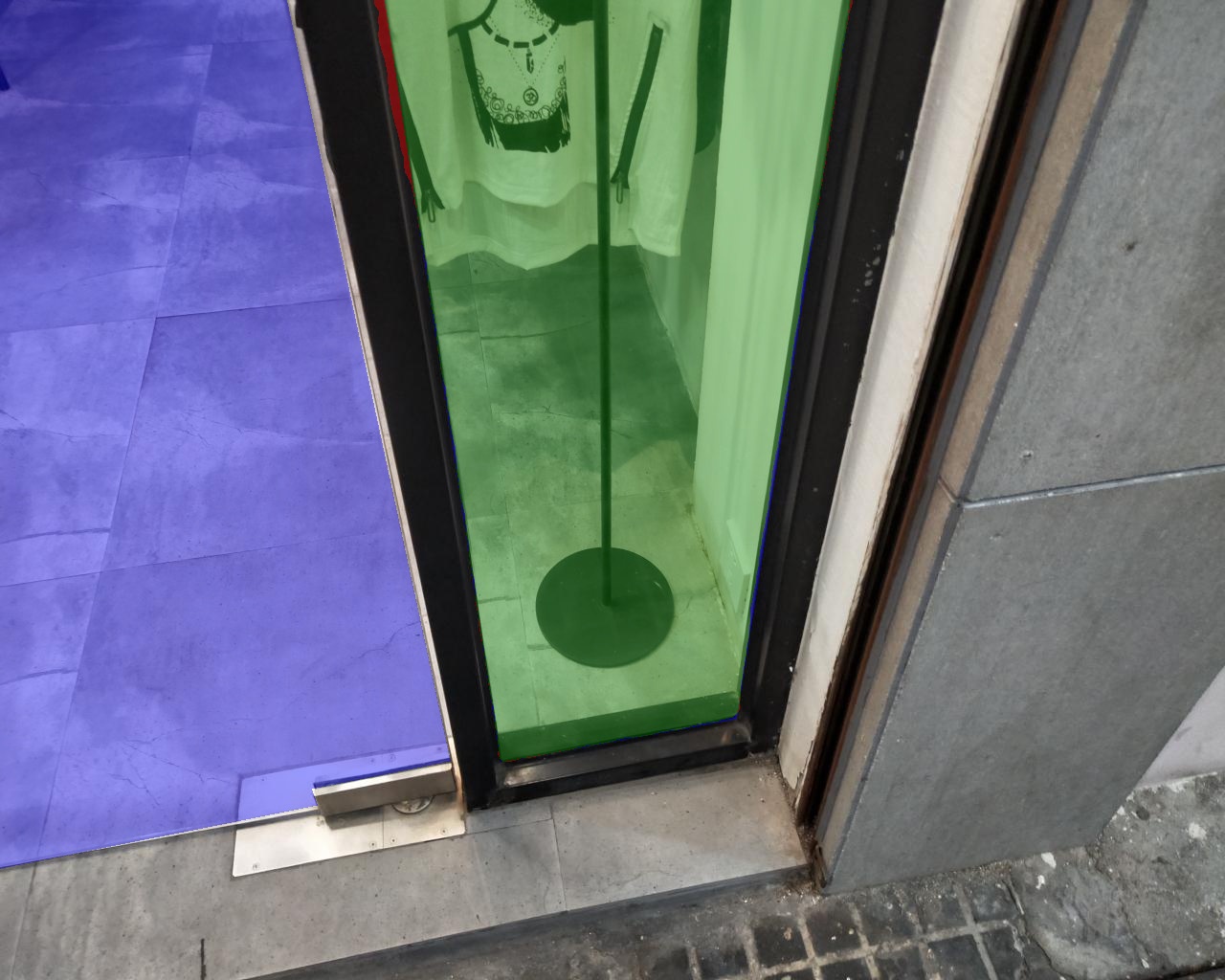} &
\includegraphics[width=\linewidth,height=\linewidth]{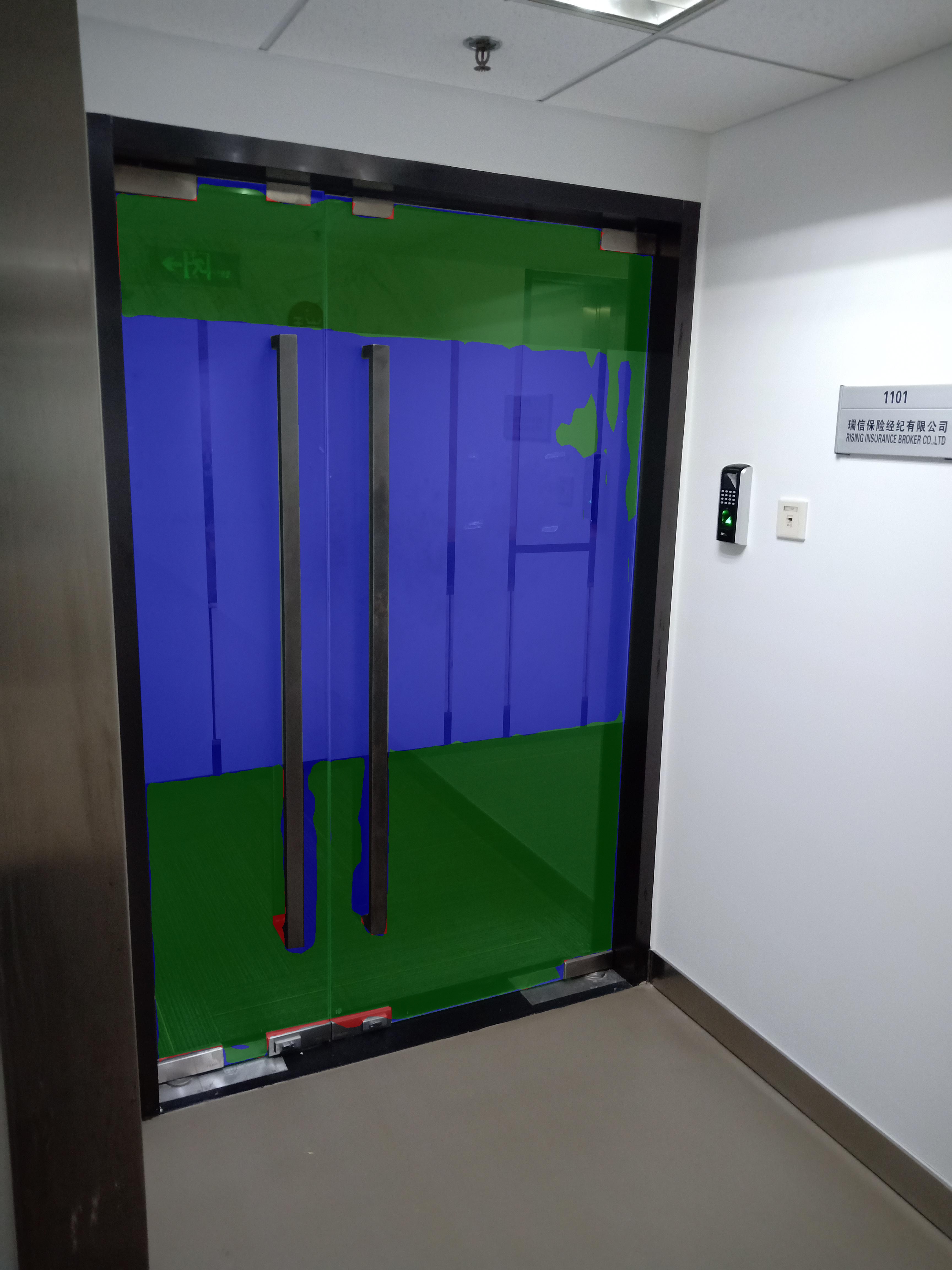} &
\includegraphics[width=\linewidth,height=\linewidth]{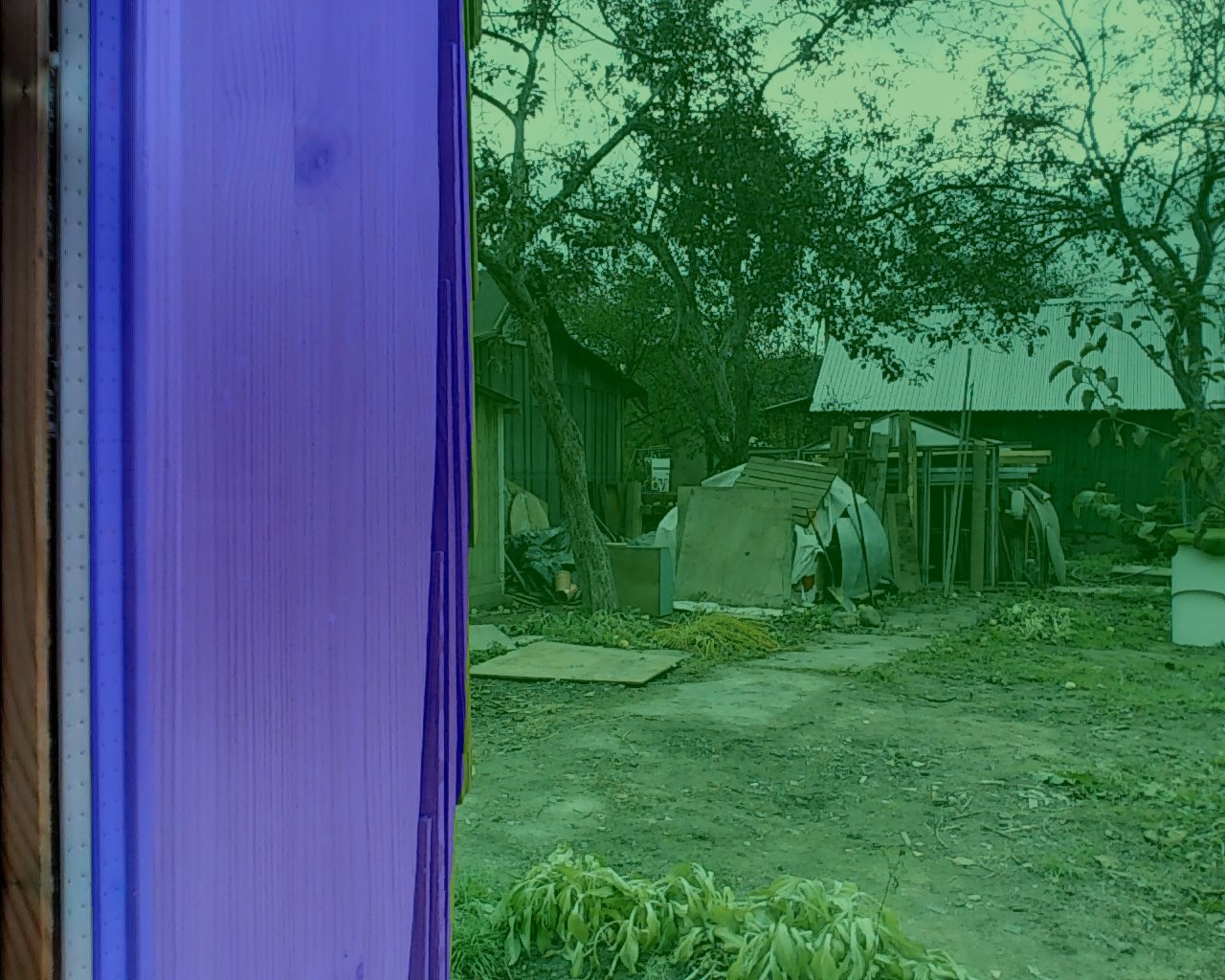} &
\includegraphics[width=\linewidth,height=\linewidth]{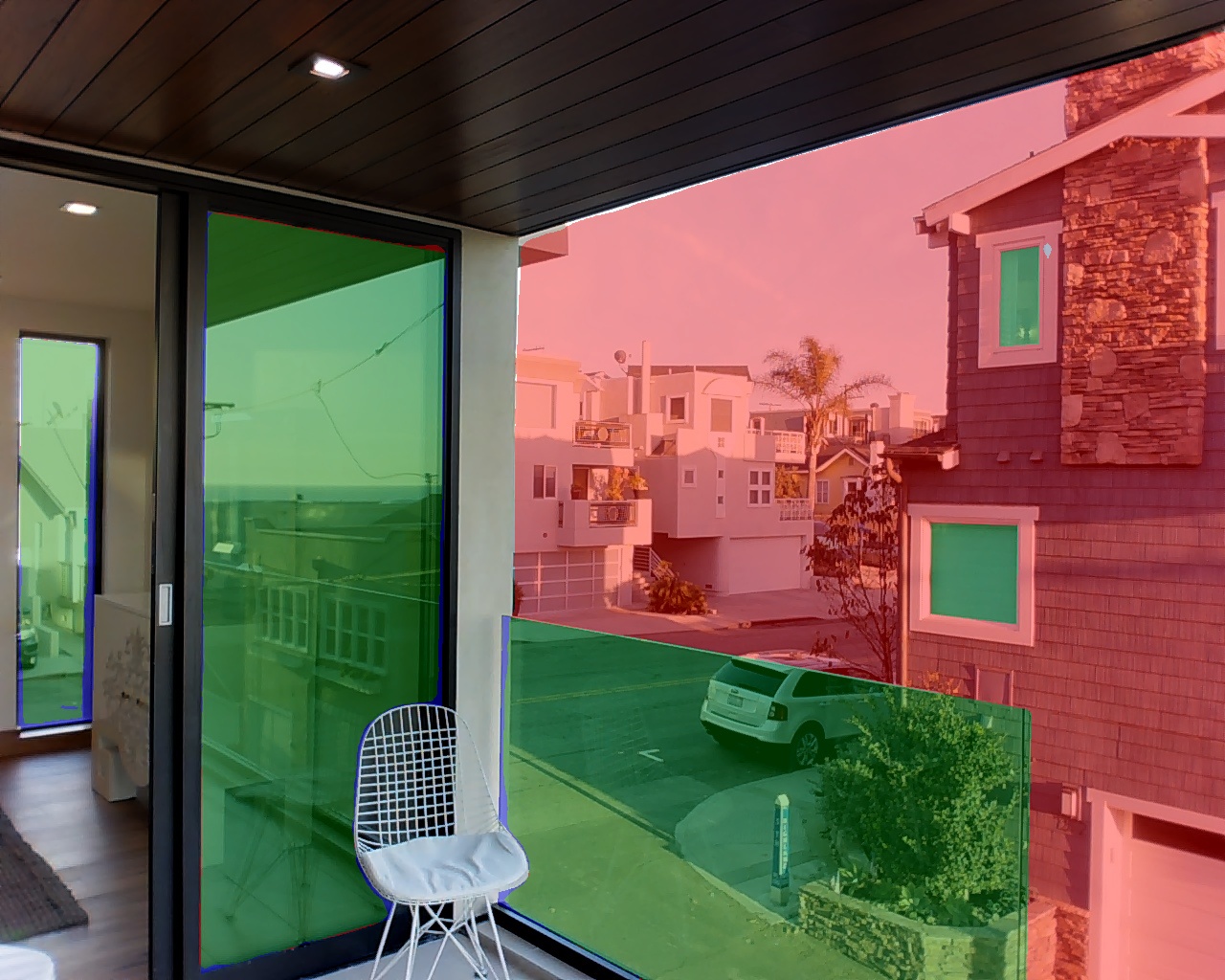} &
\includegraphics[width=\linewidth,height=\linewidth]{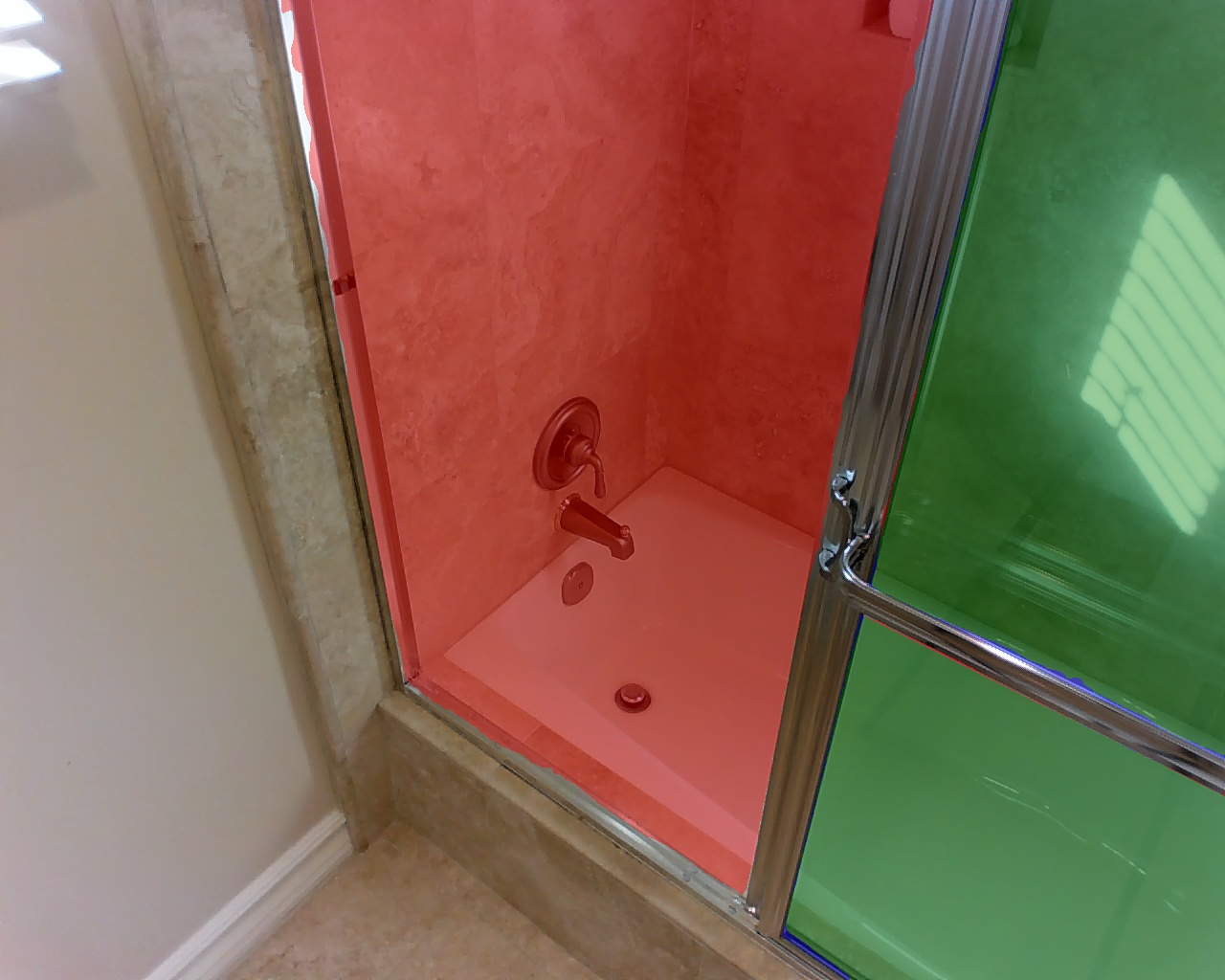} \\
\end{tabular}
\caption{Visualizations of failure cases encountered with the L+GNet model. Model trained with the combined training data. True positives overlaid in green, false positives overlaid in red, and false negatives overlaid in blue.}
\label{fig:failures}
\end{figure*}

\begin{figure*}
\centering
\setlength{\tabcolsep}{3pt}
\scriptsize
\begin{tabular}{>{\centering\arraybackslash}m{0.5cm}|*{7}{>{\centering\arraybackslash}m{0.25\columnwidth}}}

\rotatebox{90}{\makecell{Original\\ image}} &
\includegraphics[width=\linewidth,height=\linewidth]{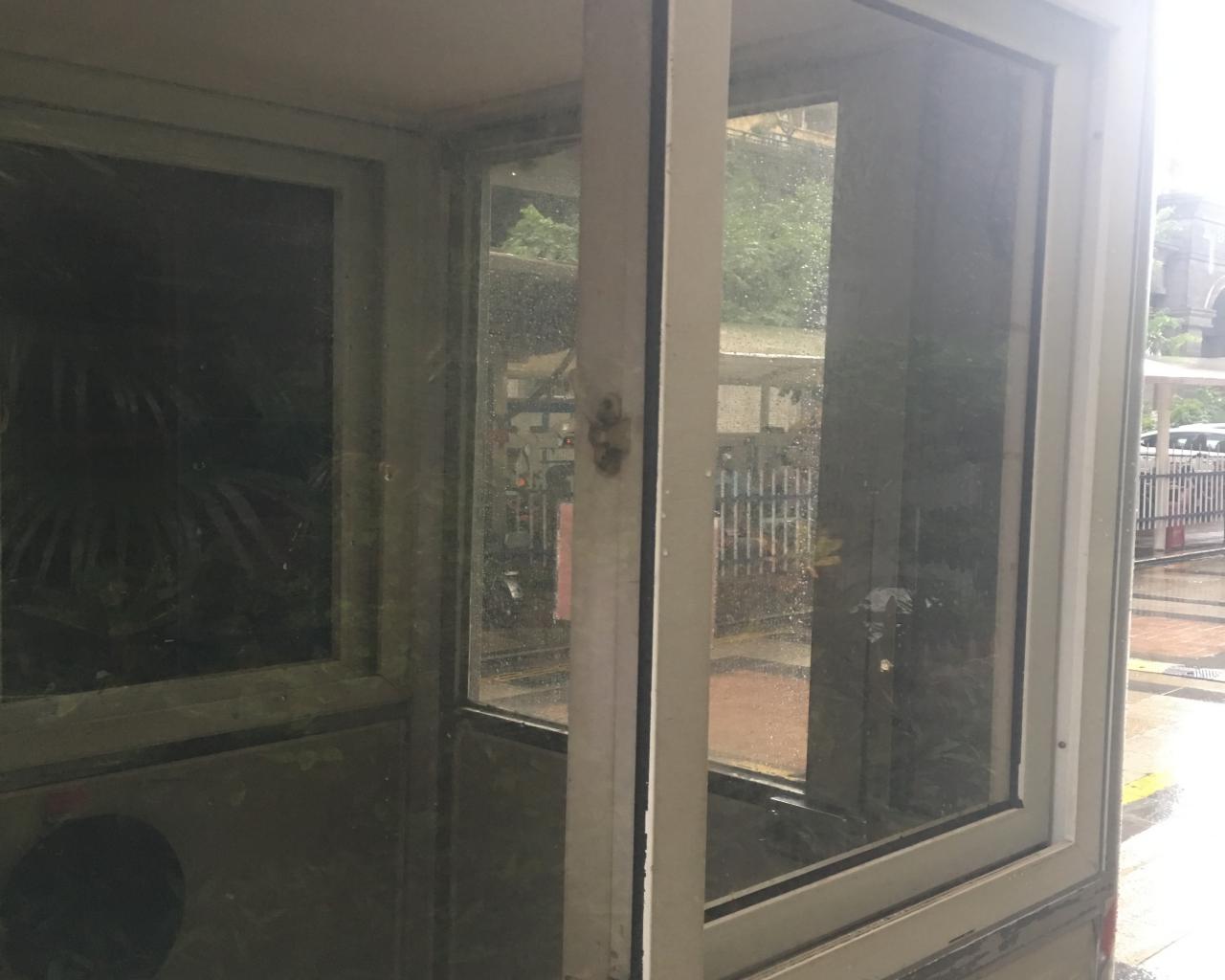} &
\includegraphics[width=\linewidth,height=\linewidth]{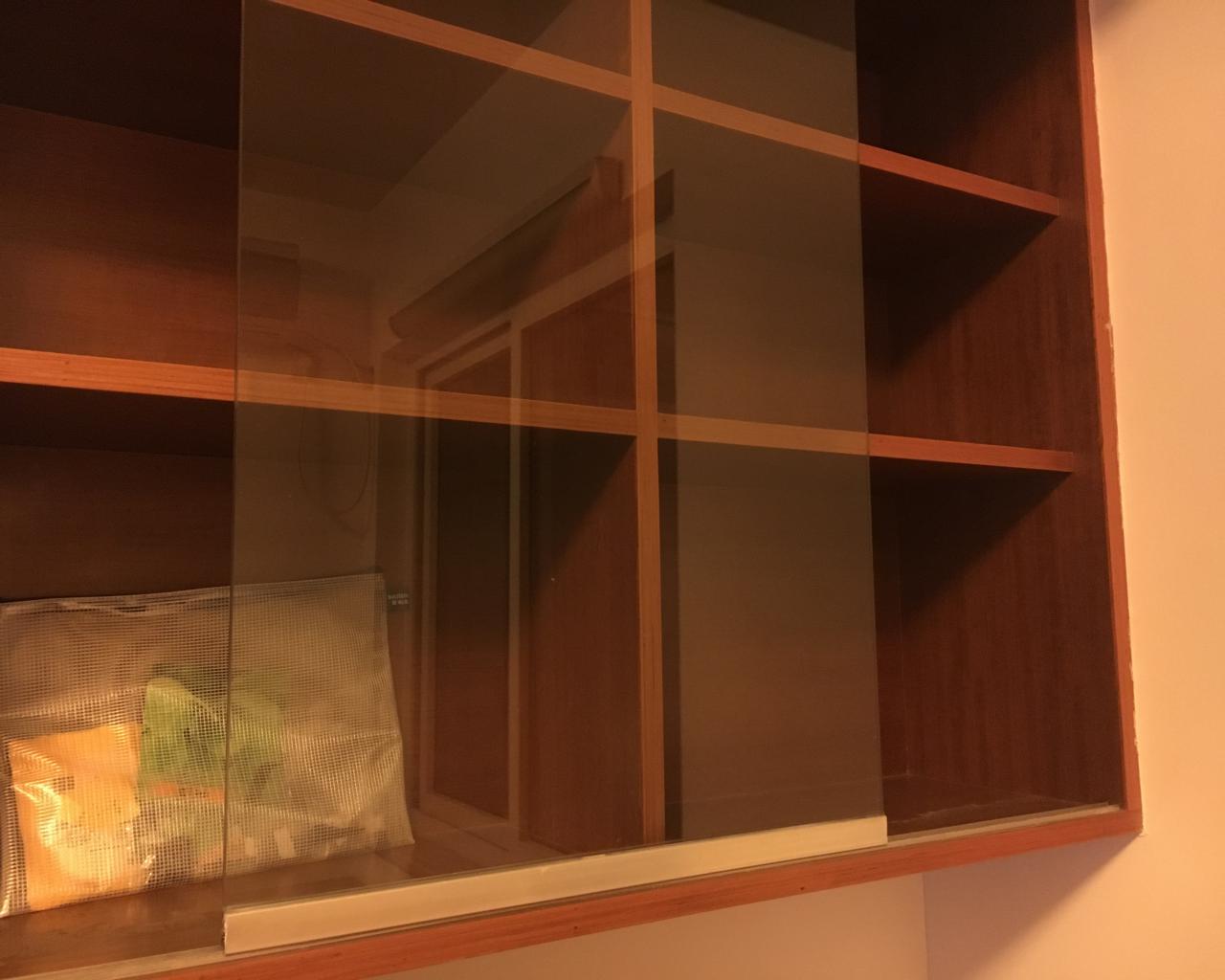} &
\includegraphics[width=\linewidth,height=\linewidth]{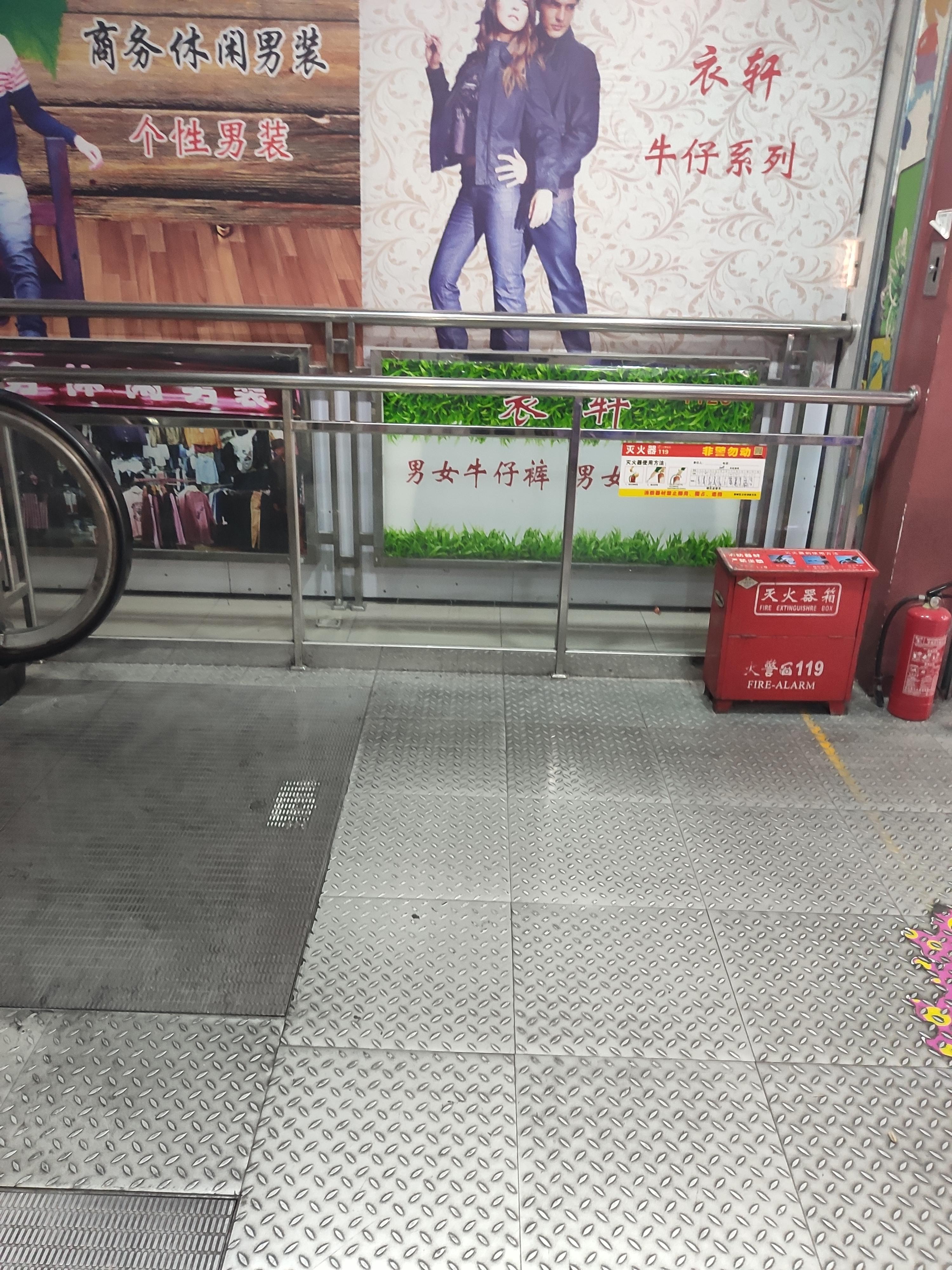} &
\includegraphics[width=\linewidth,height=\linewidth]{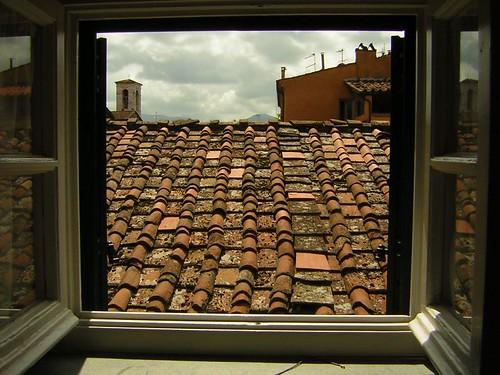} &
\includegraphics[width=\linewidth,height=\linewidth]{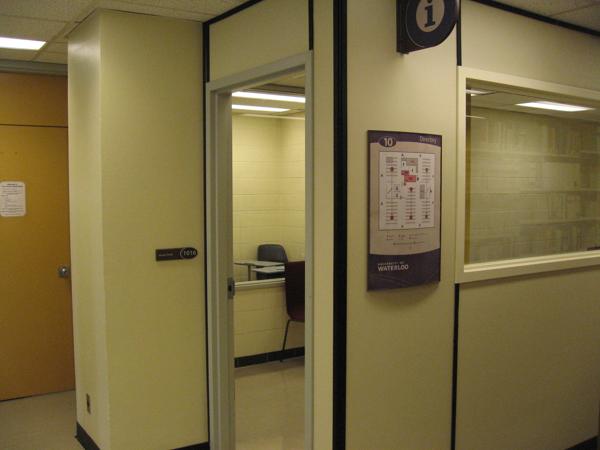} &
\includegraphics[width=\linewidth,height=\linewidth]{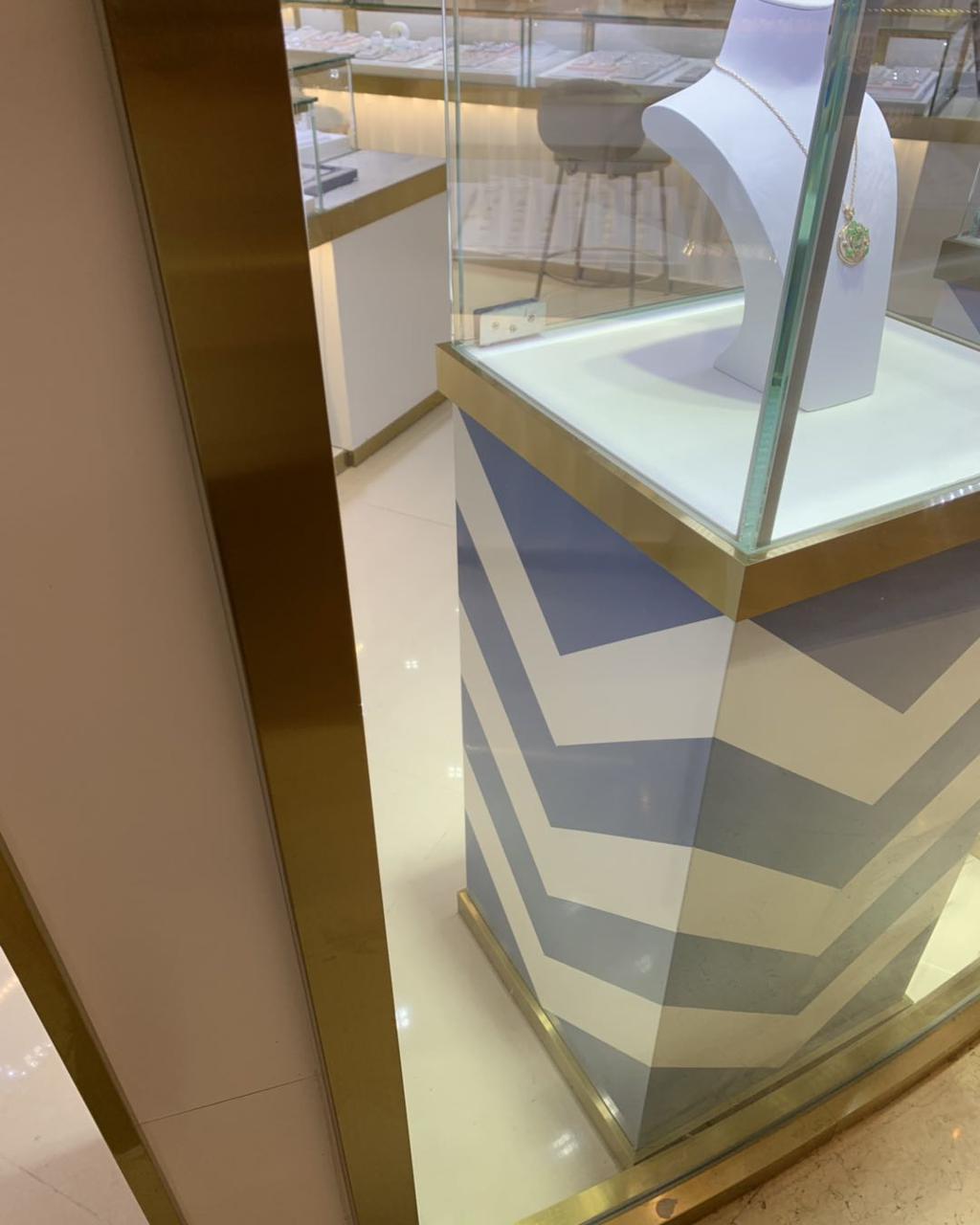} &
\includegraphics[width=\linewidth,height=\linewidth]{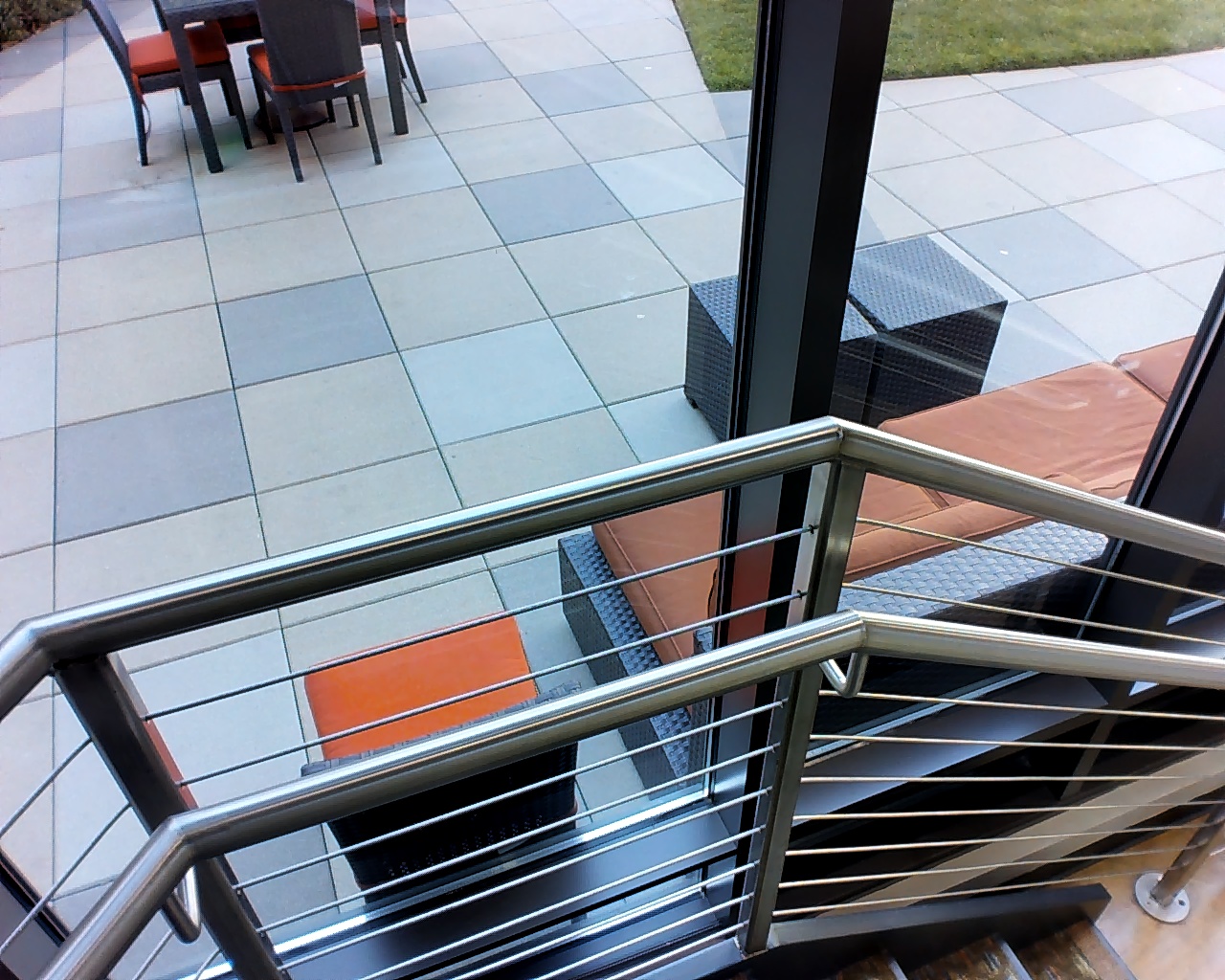} \\

\rotatebox{90}{\makecell{GlassWizard \cite{li2025glasswizard}}} &
\includegraphics[width=\linewidth,height=\linewidth]{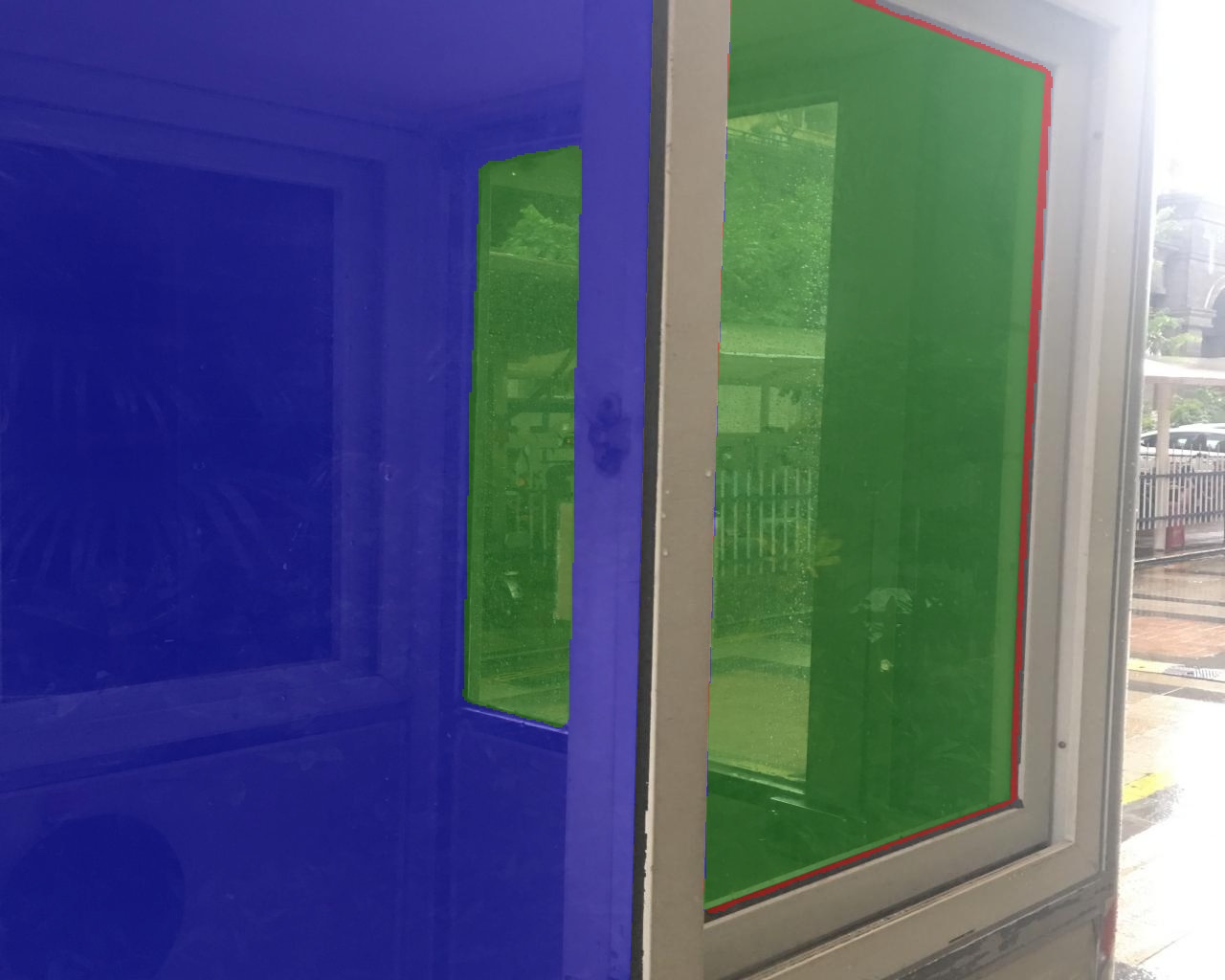} &
\includegraphics[width=\linewidth,height=\linewidth]{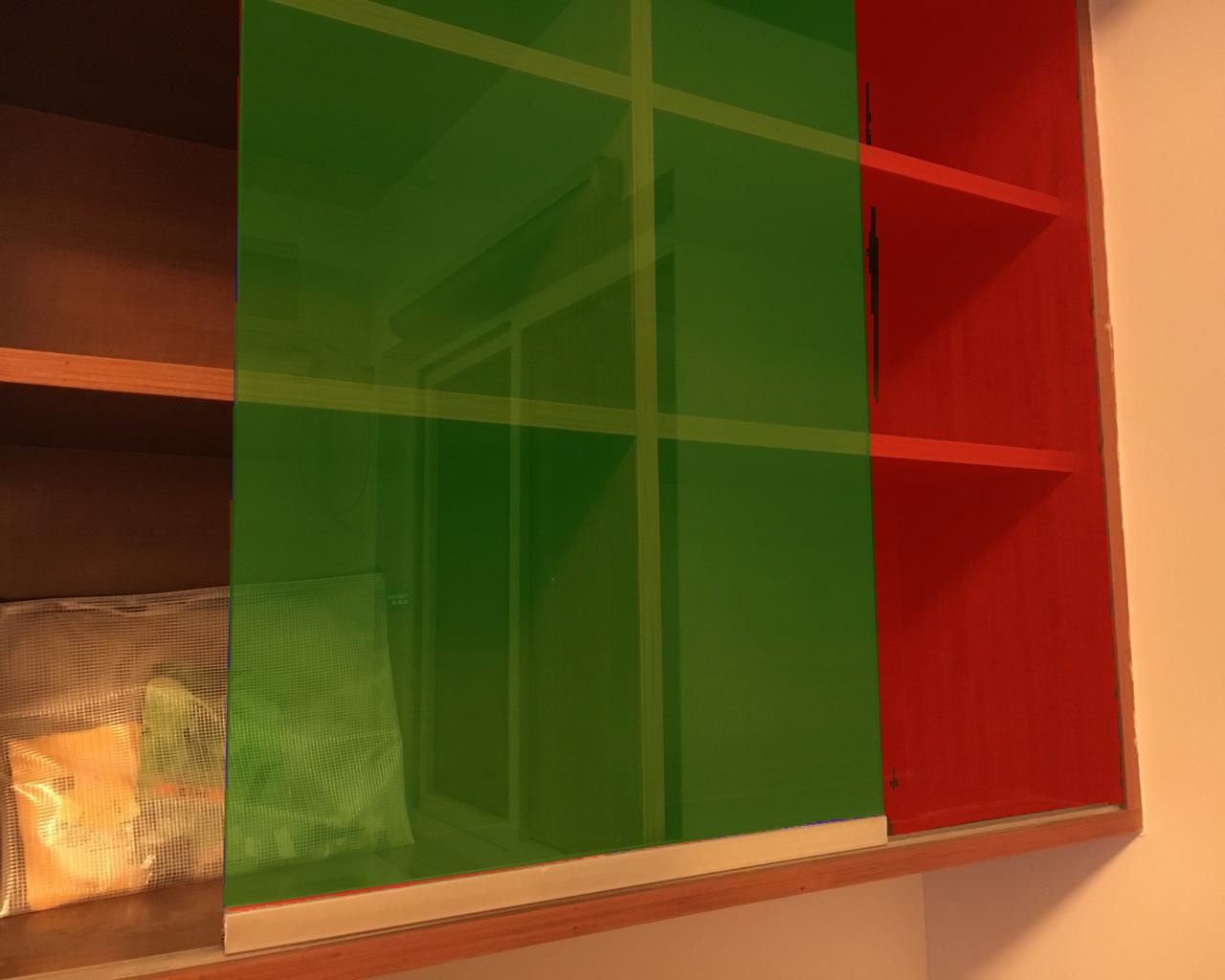} &
\includegraphics[width=\linewidth,height=\linewidth]{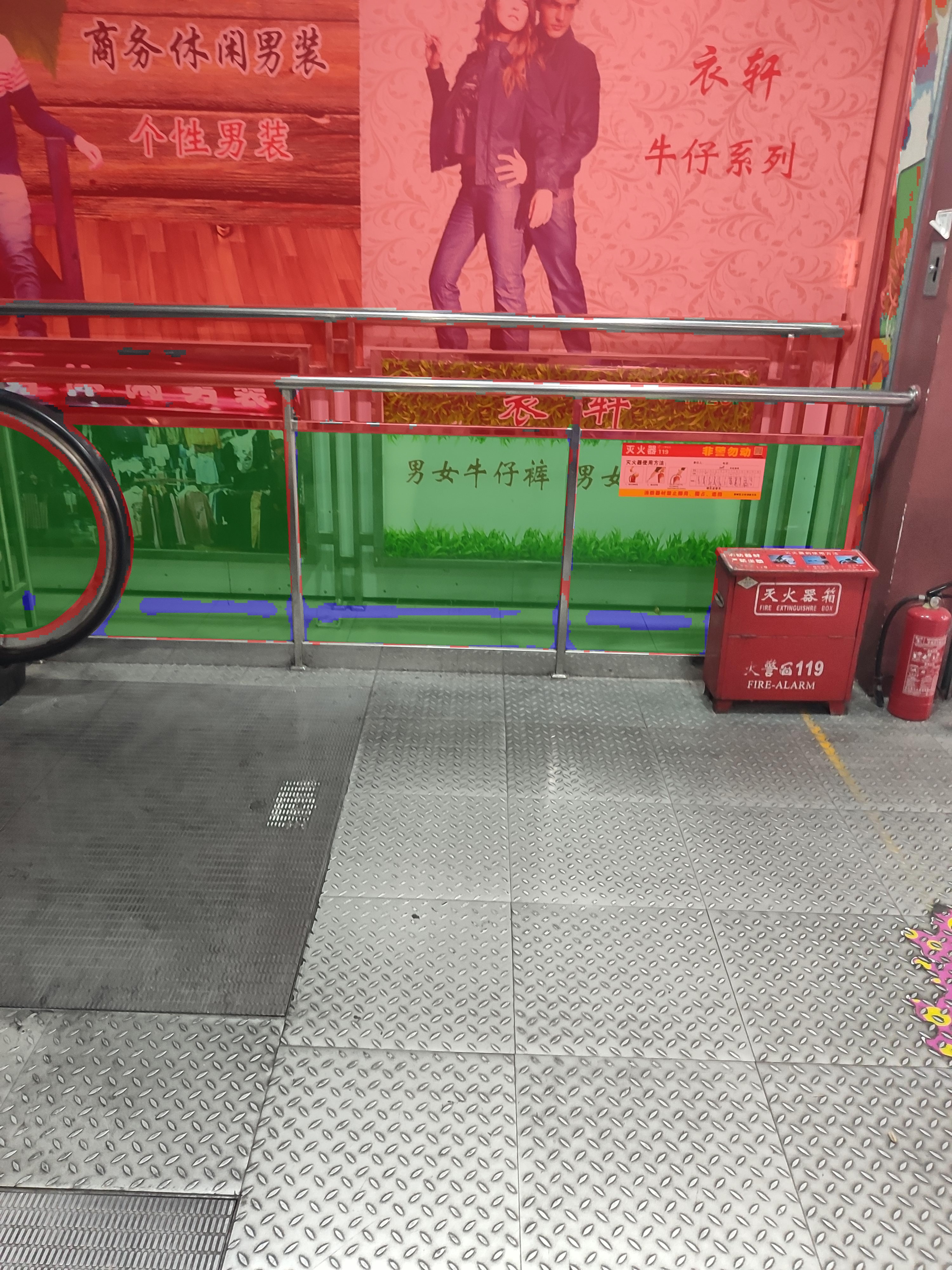} &
\includegraphics[width=\linewidth,height=\linewidth]{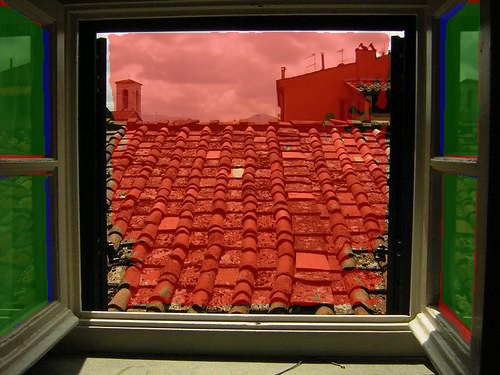} &
\includegraphics[width=\linewidth,height=\linewidth]{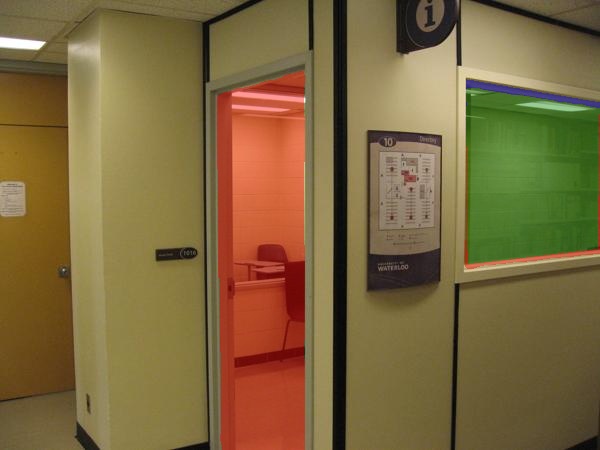} &
\includegraphics[width=\linewidth,height=\linewidth]{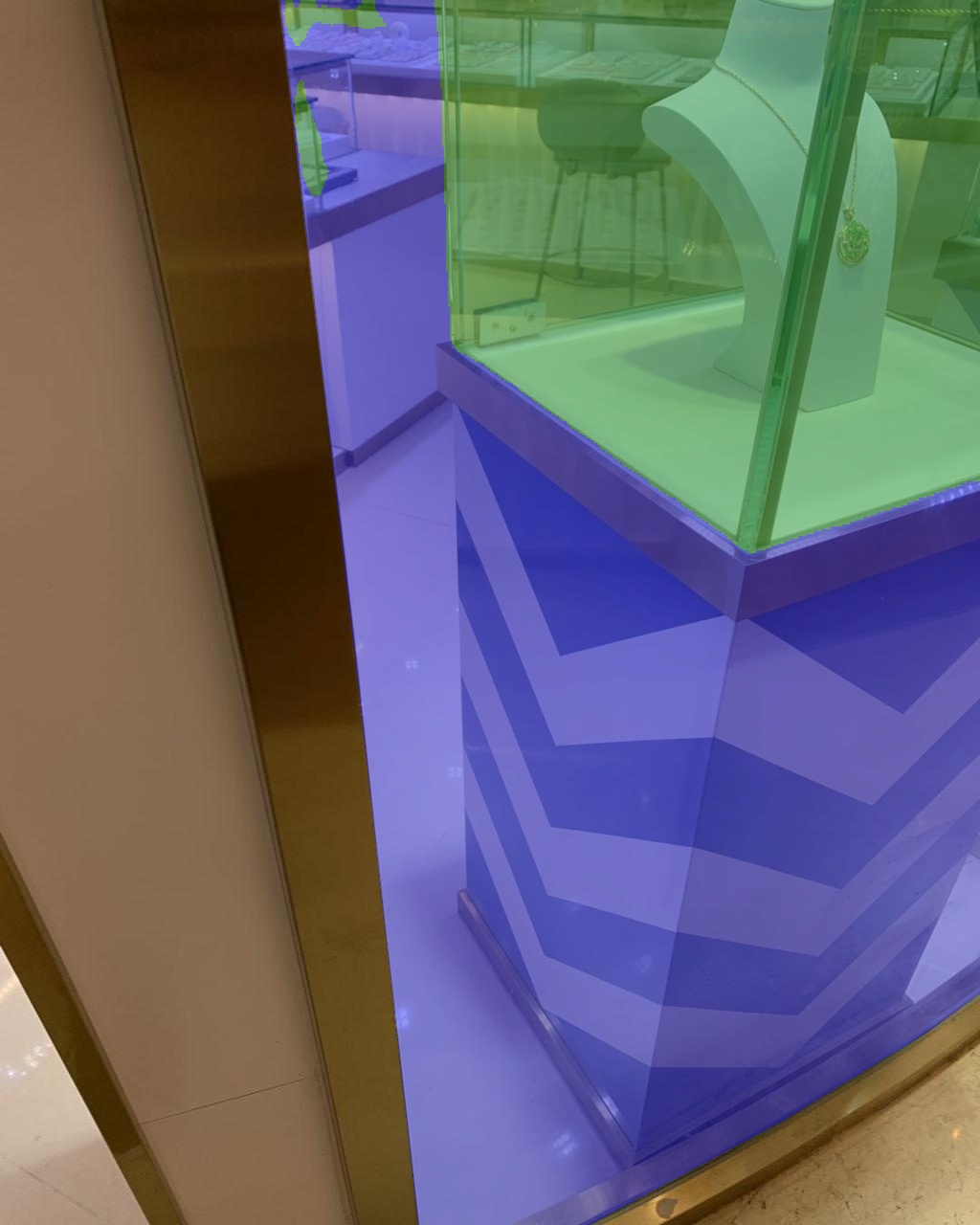} &
\includegraphics[width=\linewidth,height=\linewidth]{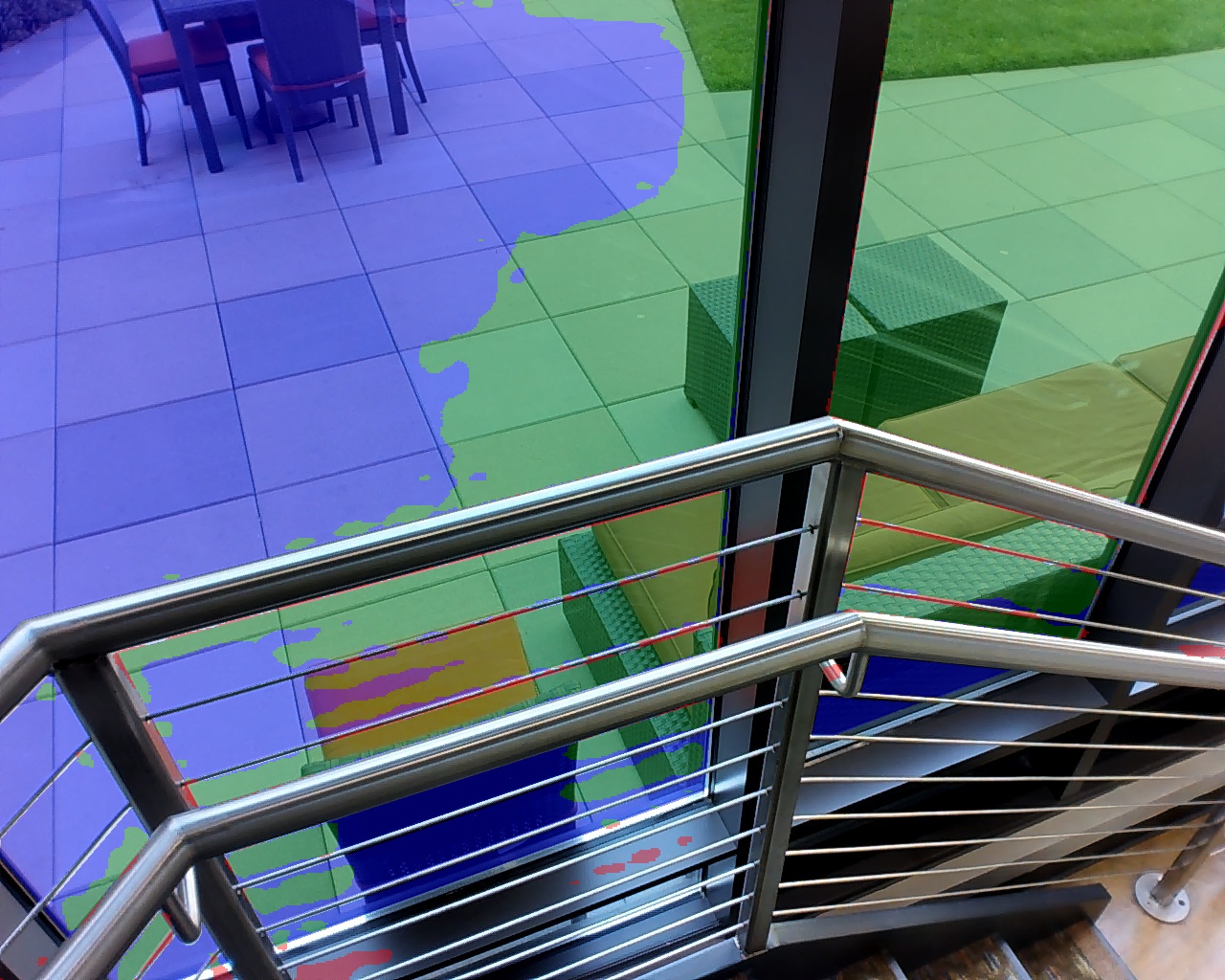} \\

\rotatebox{90}{\makecell{L+GNet}} &
\includegraphics[width=\linewidth,height=\linewidth]{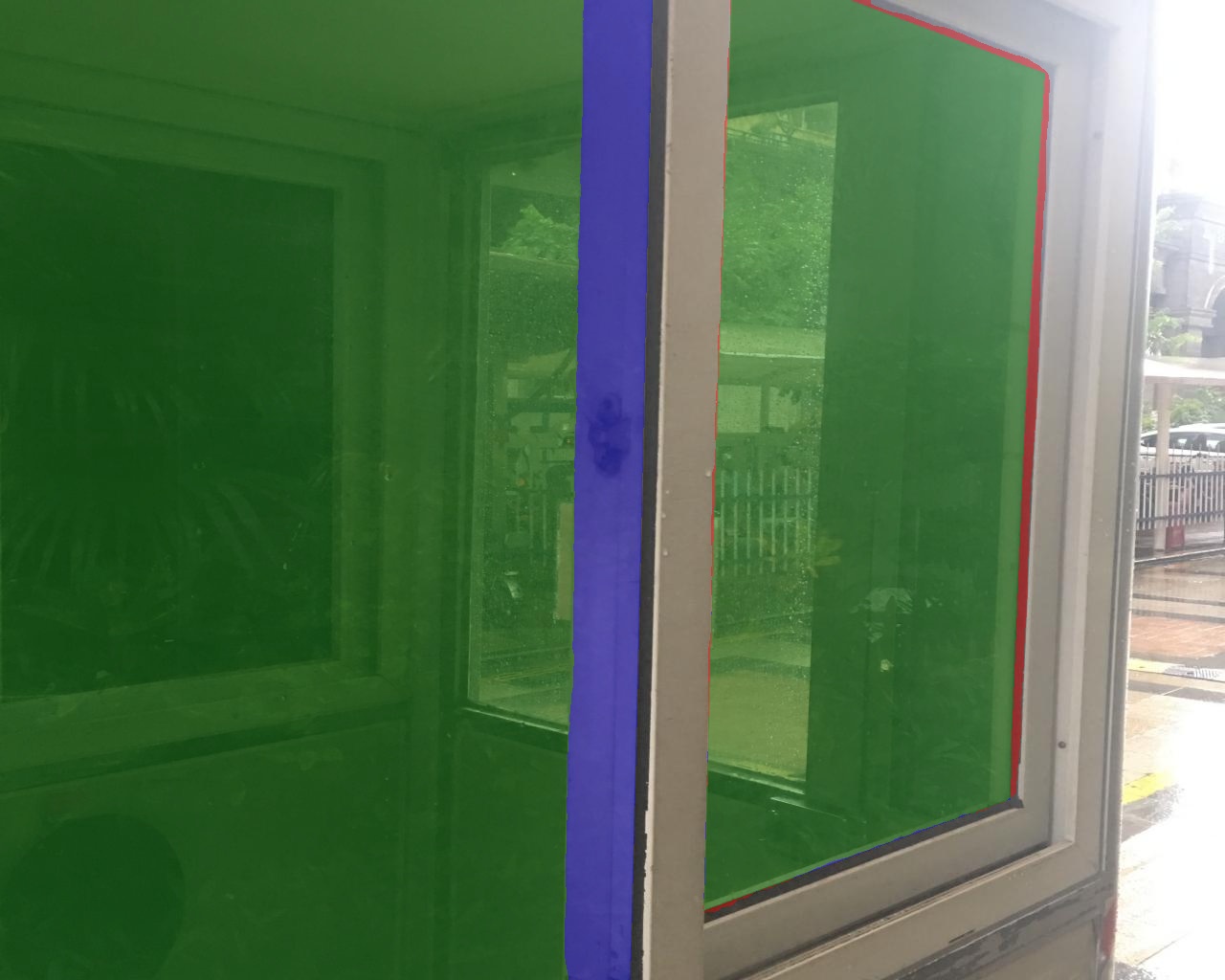} &
\includegraphics[width=\linewidth,height=\linewidth]{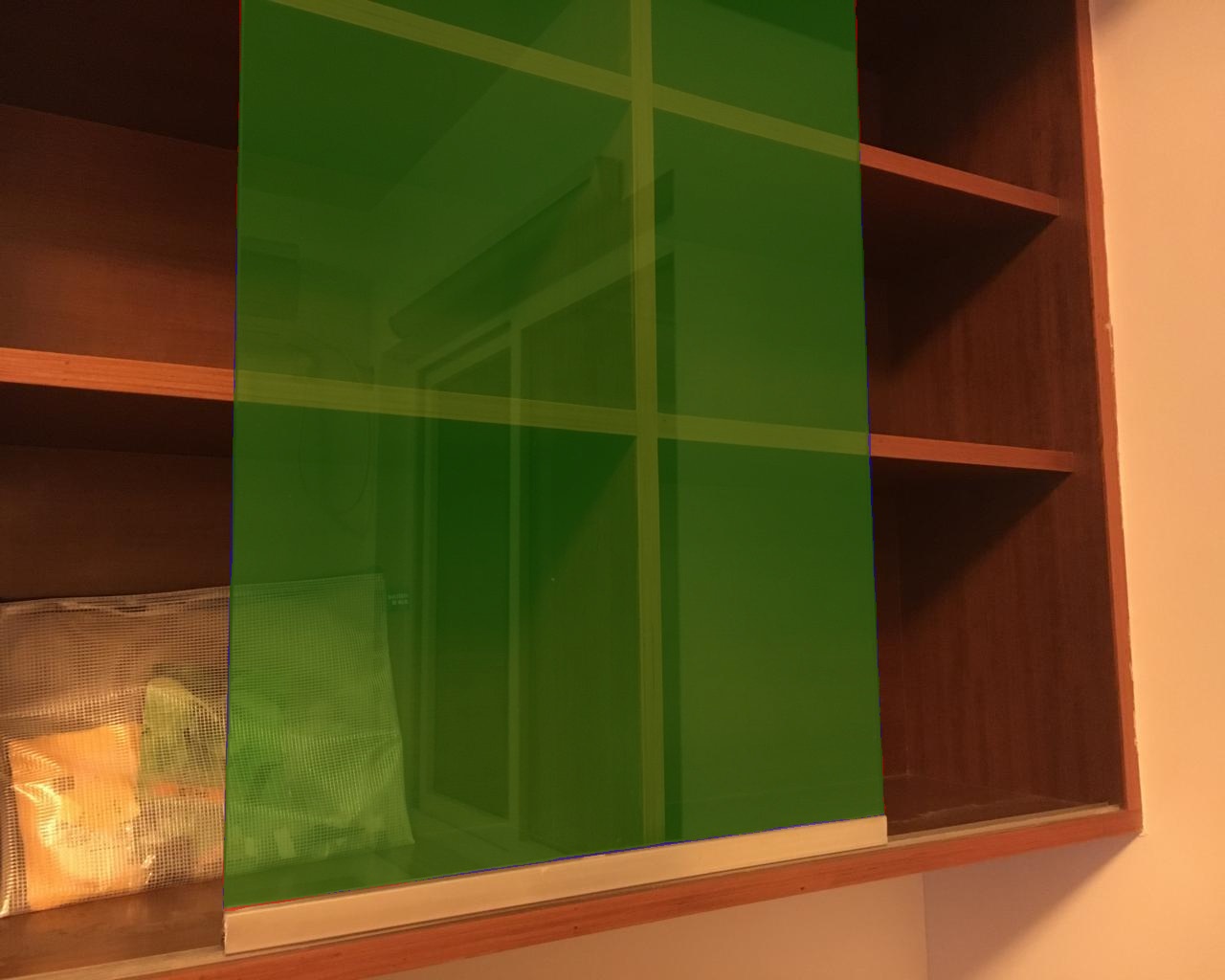} &
\includegraphics[width=\linewidth,height=\linewidth]{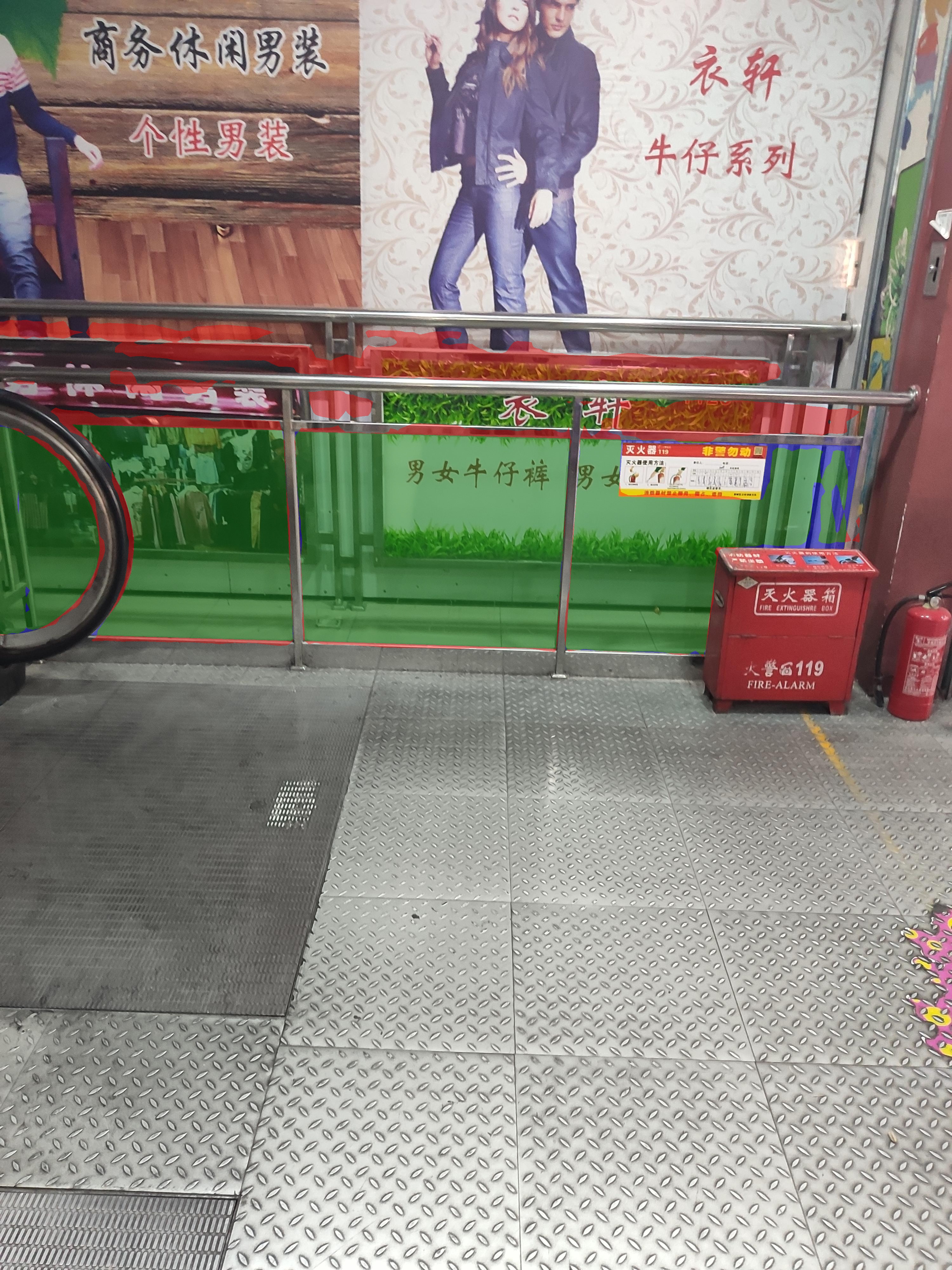} &
\includegraphics[width=\linewidth,height=\linewidth]{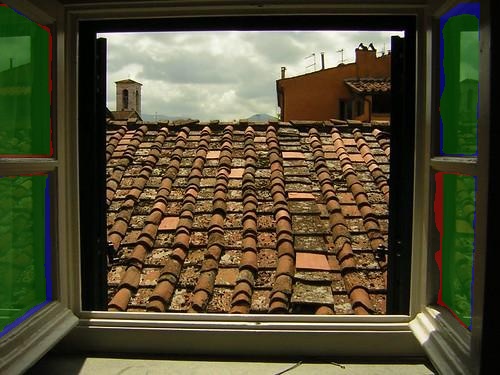} &
\includegraphics[width=\linewidth,height=\linewidth]{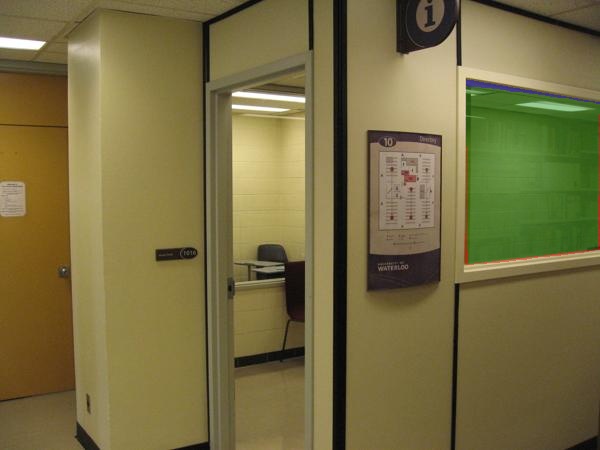} &
\includegraphics[width=\linewidth,height=\linewidth]{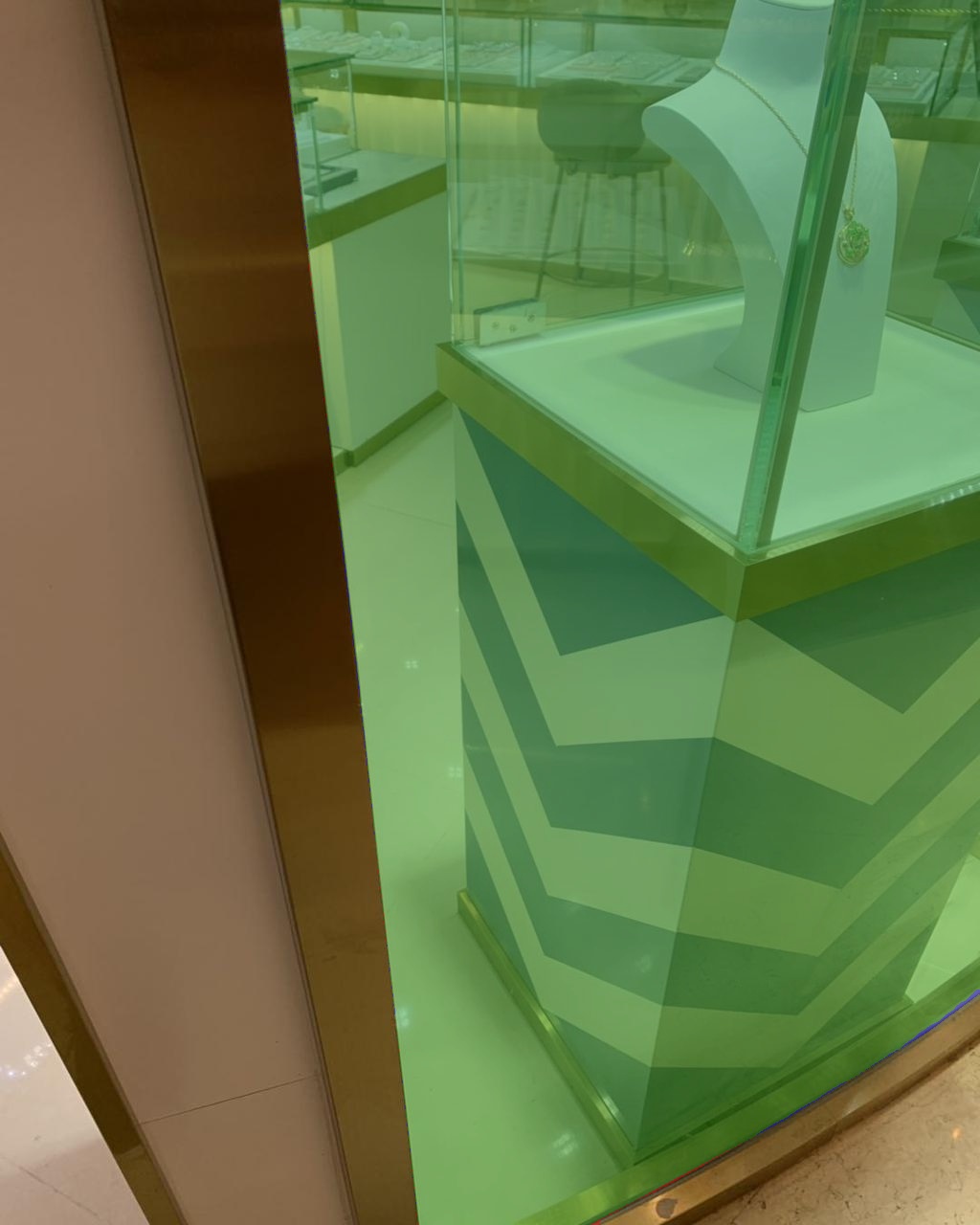} &
\includegraphics[width=\linewidth,height=\linewidth]{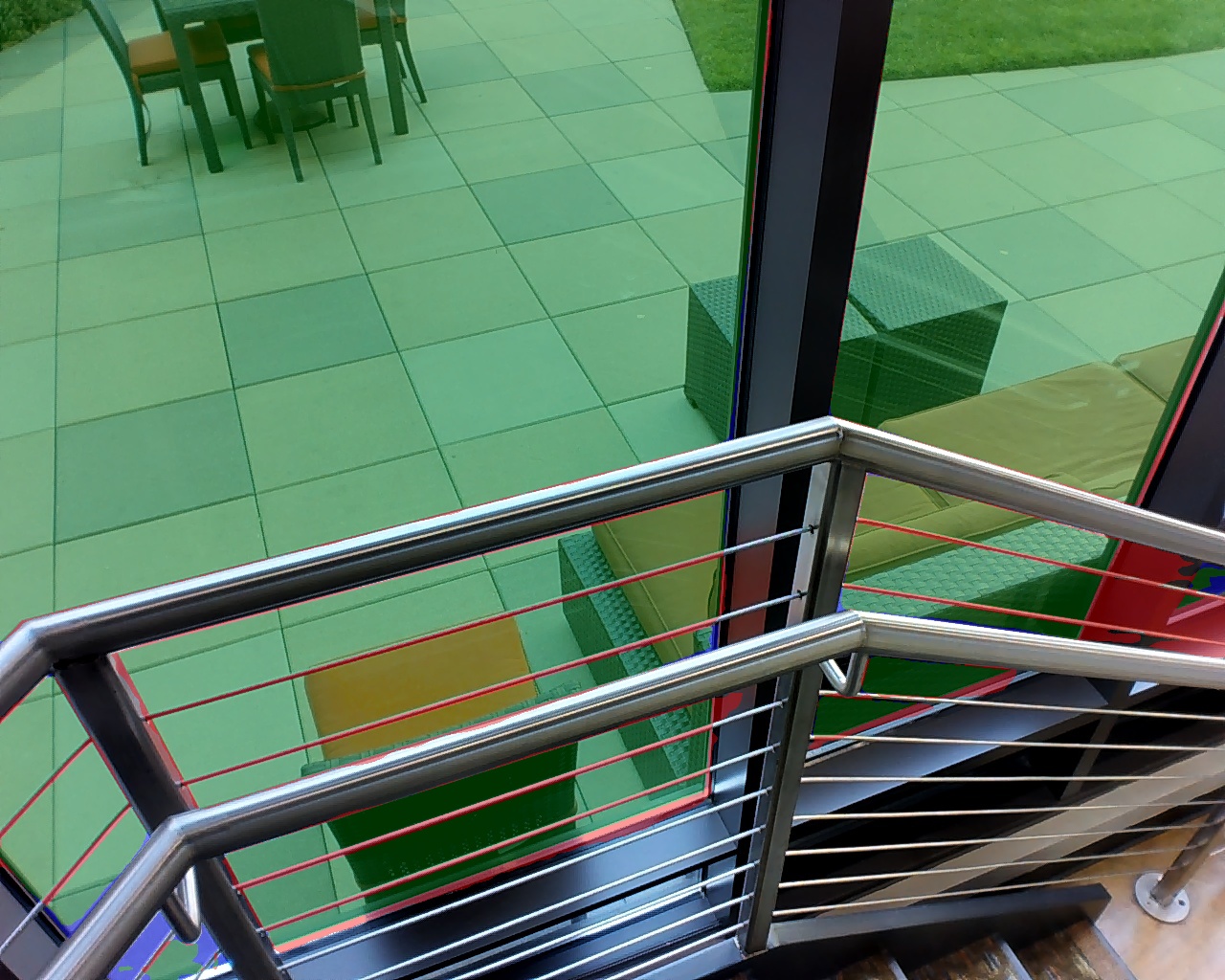} \\
\end{tabular}
\caption{Visualizations of L+GNet performance for images that the previous state of the art fails to correctly segment. L+GNet model trained with combined training data. GlassWizard results provided by the respective authors \cite{li2025glasswizard}. True positives overlaid in green, false positives overlaid in red, and false negatives overlaid in blue.}
\label{fig:comparison}
\end{figure*}

\subsection{Ablation Study}
Ablations of the proposed architecture were tested to quantify the impact of different design choices on the segmentation accuracy of the model.
Additional ablation experiments are presented in the Supplementary Material of the paper.
Table \ref{tab:ablation_components} shows the IoU results of the L+GNet model with different modules of the architecture excluded.
The results highlight the major performance gains contributed by the dual-backbone, as excluding either of the backbones resulted in a notable drop in accuracy.
The impact of SE Channel Reduction was tested by replacing the respective blocks with $1\times 1$ convolutions.
The SE Channel Reduction improved the performance especially on the GSD and HSO datasets.

To validate the generalizability of the L+GNet architecture, different backbone models were tested.
Accuracy results obtained with different backbones are presented in Table \ref{tab:ablation_backbones}.
The results show that the architecture generalizes well to different vision foundation models used in the General Features Backbone, although DINOv3-L yields the highest accuracy.
Even with the smaller DINOv3-B model as the backbone, the architecture delivers strong results, mostly outperforming previous state-of-the-art.
With the smallest DINOv3-S backbone, performance dropped notably.
The choice of Learned Features Backbone had notably less impact on the performance, with MiT-B3 producing similar results as the Swin-S backbone.
Scaling the model size had an adverse impact on the results, as shown by the performance of the larger Swin-B backbone variant.
This was likely due to the limited amount of training data.

\begin{table}
\scriptsize
\setlength{\tabcolsep}{1.5pt}
\centering
\caption{Ablation study of the performance gains of different components in the L+GNet architecture. Models trained and tested respectively on each dataset. Results reported as mean {\tiny$\pm$} standard deviation over 5 independent training runs. Bolded values indicate the best mean in each column.}
\label{tab:ablation_components}
\newcommand{\ck}{\begin{tabular}[c]{@{}c@{}}\checkmark\end{tabular}}
\begin{tabular}{ccc|cccc}
\hline
\makecell{Learned \\ Backbone \\ (Swin-S)} & \makecell{General \\ Backbone \\ (DINOv3-L)} & \makecell{SE Channel \\ Reduction}  & \makecell{GDD \\ \cite{mei2020don} \\ IoU$\uparrow$} & \makecell{GSD \\ \cite{lin2021rich} \\ IoU$\uparrow$} & \makecell{Trans10k- \\Stuff \cite{xie2020segmenting} \\ IoU$\uparrow$} & \makecell{HSO \\ \cite{yu2022progressive}\\ IoU$\uparrow$} \\
\hline
\ck & & &
\begin{tabular}[c]{@{}c@{}}0.910\\{\tiny$\pm$}0.002\end{tabular} &
\begin{tabular}[c]{@{}c@{}}0.889\\{\tiny$\pm$}0.002\end{tabular} &
\begin{tabular}[c]{@{}c@{}}0.933\\{\tiny$\pm$}0.001\end{tabular} &
\begin{tabular}[c]{@{}c@{}}0.830\\{\tiny$\pm$}0.006\end{tabular} \\
\hline
& \ck & &
\begin{tabular}[c]{@{}c@{}}0.930\\{\tiny$\pm$}0.002\end{tabular} &
\begin{tabular}[c]{@{}c@{}}0.907\\{\tiny$\pm$}0.001\end{tabular} &
\begin{tabular}[c]{@{}c@{}}0.928\\{\tiny$\pm$}0.002\end{tabular} &
\begin{tabular}[c]{@{}c@{}}0.864\\{\tiny$\pm$}0.04\end{tabular} \\
\hline
\ck & \ck & &
\begin{tabular}[c]{@{}c@{}}0.941\\{\tiny$\pm$}0.002\end{tabular} &
\begin{tabular}[c]{@{}c@{}}0.923\\{\tiny$\pm$}0.004\end{tabular} &
\begin{tabular}[c]{@{}c@{}}0.946\\{\tiny$\pm$}0.001\end{tabular} &
\begin{tabular}[c]{@{}c@{}}0.872\\{\tiny$\pm$}0.005\end{tabular} \\
\hline
\ck & \ck & \ck &
\begin{tabular}[c]{@{}c@{}}\textbf{0.943}\\{\tiny$\pm$}0.003\end{tabular} &
\begin{tabular}[c]{@{}c@{}}\textbf{0.930}\\{\tiny$\pm$}0.001\end{tabular} &
\begin{tabular}[c]{@{}c@{}}\textbf{0.948}\\{\tiny$\pm$}0.001\end{tabular} &
\begin{tabular}[c]{@{}c@{}}\textbf{0.882}\\{\tiny$\pm$}0.002\end{tabular} \\
\hline
\end{tabular}
\end{table}

\begin{table}
\scriptsize
\setlength{\tabcolsep}{3pt}
\centering
\caption{Ablation study of different backbone models used in the L+GNet architecture. Models trained and tested respectively on each dataset. Bolded values indicate the best result in each column.}
\label{tab:ablation_backbones}
\begin{tabular}{cc|cccc}
\hline
\makecell{Learned \\ Backbone} & \makecell{General \\ Backbone} & \makecell{GDD \\ \cite{mei2020don} \\ IoU$\uparrow$} & \makecell{GSD \\ \cite{lin2021rich} \\ IoU$\uparrow$} & \makecell{Trans10k- \\Stuff \cite{xie2020segmenting} \\ IoU$\uparrow$} & \makecell{HSO \\ \cite{yu2022progressive}\\ IoU$\uparrow$} \\
\hline
Swin-S \cite{liu2021swin} & SigLIP 2-L \cite{tschannen2025siglip} & 0.927 & 0.917 & 0.940 & 0.861 \\
Swin-S \cite{liu2021swin} & PE-Core-L \cite{bolya2026perception} & 0.939 & 0.919 & 0.942 & 0.868 \\
Swin-B \cite{liu2021swin} & DINOv3-L \cite{simeoni2025dinov3} & 0.939 & 0.927 & \textbf{0.947} & 0.871 \\
MiT-B3 \cite{xie2021segformer} & SigLIP 2-L \cite{tschannen2025siglip} & 0.916 & 0.911 & 0.928 & 0.844 \\
MiT-B3 \cite{xie2021segformer} & PE-Core-L \cite{bolya2026perception} & 0.932 & 0.917 & 0.932 & 0.853 \\
MiT-B3 \cite{xie2021segformer} & DINOv3-L \cite{simeoni2025dinov3} & 0.944 & \textbf{0.931} & 0.945 & \textbf{0.885} \\
\hline
Swin-S \cite{liu2021swin} & DINOv3-S \cite{simeoni2025dinov3} & 0.918 & 0.896 & 0.934 & 0.841 \\
Swin-S \cite{liu2021swin} & DINOv3-B \cite{simeoni2025dinov3} & 0.933 & 0.922 & 0.940 & 0.860 \\
\hline
Swin-S \cite{liu2021swin} & DINOv3-L \cite{simeoni2025dinov3} & \textbf{0.948} & \textbf{0.931} & \textbf{0.947} & 0.881 \\
\hline
\end{tabular}
\end{table}

\subsection{Efficiency Analysis}
As many applications often depend on online usage of the models with limited computational power, efficiency of L+GNet and its variants was tested and compared to previous literature.
Available open-source codes of the methods from the literature were used for benchmarking.
Inference speeds, model sizes, GFLOPs, and maximum VRAM used are reported in Table \ref{tab:sizes} for $512\times 512$ inputs.
Reported values were measured as an average of one thousand forward passes for each model.
GFLOPs and maximum VRAM used were measured with built-in PyTorch tools.

The inference speed test results highlight that the L+GNet model can be run on an RTX 3090 GPU at 14.2 fps (frames per second) on FP16 precision and at 8.0 fps on FP32 precision. 
Previous state of the art model GlassWizard \cite{li2025glasswizard} has a slightly faster inference speed than the default L+GNet model.
However, the smaller L+GNet variant with a DINOv3-B backbone achieves a faster inference speed than GlassWizard, while still providing more accurate segmentation results on most datasets.

\begin{table}
\centering
\begin{threeparttable}
\caption{Comparison of model sizes and inference efficiencies. Benchmark carried out on an RTX 3090 GPU.}
\label{tab:sizes}
\setlength{\tabcolsep}{1.6pt}
\scriptsize
\begin{tabular}{p{2.3cm}|cccccc}
\hline
Model & \makecell{Params \\ ($10^6$)} & \makecell{Trainable \\ params \\ ($10^6$)} & \makecell{Inference \\ FP16 \\ (fps)} & \makecell{Inference \\ FP32 \\ (fps)} & \makecell{VRAM \\ FP32 \\ (GB)} & \makecell{GFLOPs}  \\ 
\hline
GlassSemNet \cite{lin2022exploiting} & 240 & 167 & 16.3 & 12.1 & 1.79 & 1400\\
C-LPMoE-u \cite{sun2025controllable} & 327 & 23.4 & -\tnote{2} & 0.69 & 1.67 & -\tnote{3}\\
GlassWizard\tnote{1} \cite{li2025glasswizard} & 950 & 866 & 16.9 & 8.2 & 5.45 & 1780\\
\hline
L+GNet w/ DINOv3-S  & 117 & 94.7 & 21.2 & 18.6 & 0.76 & 380\\
L+GNet w/ DINOv3-B  & 206 & 119 & 18.5 & 13.4 & 1.09 & 820\\
\hline
L+GNet & 450 & 145 & 14.2 & 8.0 & 2.06 & 1610\\
\hline
\end{tabular}
\begin{tablenotes}
\scriptsize
\item[1] Inference model setup. Text encoder excluded.
\item[2] FP16 not supported by the implementation.
\item[3] Implementation not supported by PyTorch FlopCounterMode.
\end{tablenotes}
\end{threeparttable}
\end{table}


%% file: sections/5_discussion.tex
As shown by the accuracy benchmarks in Table \ref{tab:results}, the proposed L+GNet model achieved the highest accuracy on all of the metrics used, setting new state of the art records on the datasets.
Still, some specific cases remain difficult for RGB-based glass segmentation, as demonstrated in Fig. \ref{fig:failures}.
When there are minimal direct visual clues of a glass surface, yet a suitable surrounding is visible, the model is essentially forced to guess whether a glass surface is present.
The model sometimes also interprets the whole scene to be observed through a glass surface.
Additionally, the model may struggle to determine whether objects are positioned in front of a glass surface or behind it.
Tape, posters, and other flat materials directly on glass surfaces may also mislead the model.

The main accuracy benchmarks of this paper are presented with the IoU and BER metrics, although previous literature has commonly additionally applied mean absolute error (MAE) and F-measure ($F_\beta$) as well.
However, MAE values are generally computed based on prediction confidence values, which the L+GNet architecture is not well suited to produce (Fig. \ref{fig:calibration}).
For this reason, a comparison to the previous literature on this metric is not presented.
Analyzing the previous literature, it was noted that papers have mixed the weighted F-measure $F^w_\beta$ proposed by Margolin et al. \cite{margolin2014evaluate} and the traditional F-measure $F_\beta$ \cite{yu2022progressive, qi2024glass} when reporting results.
Additionally, not all works clearly report the $\beta$-value that they are using when computing the results.
Due to this uncertainty, the $F_\beta$ metric was also excluded from the comparison to previous models.
However, results on binary MAE and $F_\beta$ are included in the Supplementary Material.

Although L+GNet with DINOv3-L backbone scored highest results, practical online application of the model could be more convenient with the DINOv3-B backbone. 
Using this backbone likely provides a more feasible accuracy-speed trade-off for many applications, as evident from the results reported in Tables \ref{tab:ablation_backbones} and \ref{tab:sizes}.
However, the computational overhead and size of the General Features Backbone might not limit the applicability of the L+GNet model in specific online perception pipelines.
As the model uses frozen foundation model features, the model can be applied with minimal added computational cost in systems that utilize foundation model features for other tasks as well.

%% file: sections/6_conclusion.tex
This work proposed a novel deep learning architecture, L+GNet, for glass segmentation from RGB images.
The key novelty in the architecture is a foundation model based dual-backbone, combining findings from previous works \cite{sun2025controllable, lin2022exploiting} into a new state-of-the-art model.
The network fuses features produced by a task-specific Learned Features Backbone, and a foundation model-based General Features Backbone.
Features acquired from the backbones are processed with SE Channel Reduction, and fed into a Mask2Former Decoder which produces the final segmentation masks.
This proposed architecture was proven to produce state-of-the-art results by a number of experiments carried out on different datasets widely used in the previous literature.
In addition to accuracy benchmarks, the inference speed of the model was also studied, and found to be similar to previous state-of-the-art.
However, in practical use the computational load could be notably reduced with a more lightweight General Features Backbone variant or with integration to a perception pipeline with shared vision foundation model features.
Future development of the model could focus on improving the calibration of the prediction confidence.
